\documentclass{article}
\usepackage{amssymb}
\usepackage{xcolor}
\usepackage{graphicx}
\usepackage{courier}

\usepackage{hyperref}
\hypersetup{
    colorlinks=true,
    linkcolor=black,
    citecolor=black,
    urlcolor=blue}

\long\def\comment#1{}

\newcommand{\Pa}{{\bf Pa}}
\newcommand{\Pai}{{\bf Pa}_i}
\newcommand{\bX}{{\bf X}}
\newcommand{\bx}{{\bf x}}
\newcommand{\bY}{{\bf Y}}
\newcommand{\by}{{\bf y}}
\newcommand{\node}[1]{\texttt{#1}}
\newcommand{\statefont}[1]{\texttt{#1}}

\title{Heckerthoughts}

\author{David Heckerman\\
  heckerma@hotmail.com}

\date{}

\begin{document}

\maketitle

\comment{add full bib}

\comment{
  The following papers do not have DOIs:
  - Dependency networks
  - Learning mixtures of DAG models -- here using UAI/arXiv version
  - First spam paper
  Redo arXiv bibtex using export feature.
}

\section{Introduction}

\begin{figure}
\begin{center}
\leavevmode
\includegraphics[width=5.0in]{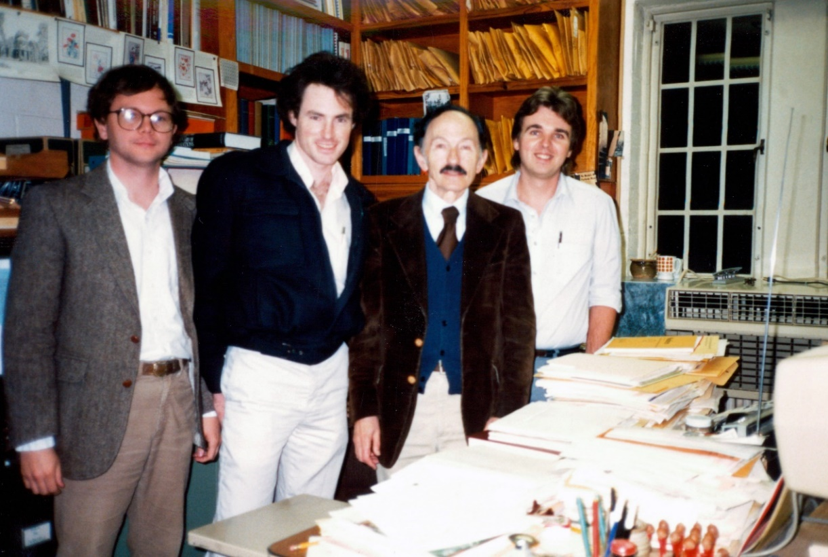}
\end{center}
\caption{A photo of my visit with I.J. Good in 1987.  From left to right: Greg Cooper, Eric Horvitz, I.J. Good, and me.}
\label{fig:ijgood}
\end{figure}

In 1987, Eric Horvitz, Greg Cooper, and I visited I.J. Good at
Virginia Polytechnic and State University. The three of us were at a
conference in Washington DC and made the short drive to see him. The
primary reason we wanted to see him was not because he worked with
Alan Turing to help win WWII by decoding encrypted messages from the
Germans, although that certainly intrigued us. Rather, we wanted to see
him because we had just finished reading his book ``Good Thinking''
\cite{Good83}, which summarized his life's work in Probability and its
Applications. We were delighted that he was willing to have lunch with
us. We were young graduate students at Stanford working in Artificial
Intelligence (AI), and were amazed that his thinking was so similar to
ours, having worked decades before us and coming from such a seemingly
different perspective not involving AI. We had lunch and talked about
many topics of shared interest including Bayesian probability,
graphical models, decision theory, AI (he was interested by then), how
the brain works, and the nature of the universe. Before we knew it, it
was dinner time. We took a brief stroll around the beautiful campus
and sat down again for dinner and more
discussion. Figure~\ref{fig:ijgood} is a photo taken just before we
reluctantly departed.

This story is a fitting introduction this manuscript. Now having
years to look back on my work, to boil it down to its essence, and to
better appreciate its significance (if any) in the evolution of AI and
Machine Learning (ML), I realized it was time to put my work in
perspective, providing a roadmap to any who would like to explore it
in more detail.  After I had this realization, it occurred to me that
this is what I.J. Good did in his book.

This manuscript is for those who want to understand basic concepts
central to ML and AI, and to learn about early applications of these
concepts.  Ironically, after I finished writing this manuscript, I
realized that a lot of the concepts that I included are missing in
modern courses on ML.  I hope this work will help to make up for these
omissions.  The presentation gets somewhat technical in parts, but
I've tried to keep the math to the bare minimum to convey the
fundamental concepts.

In addition to the technical presentations, I include stories about
how the ideas came to be and the effects they have had.  When I was a
student in physics, I was given dry texts to read. In class, however,
several of my physics professors would tell stories around the work,
such as Einstein's thinking that led to his general theory of
relativity.  Those stories fascinated me and really made the theory
stick.  So here, I do my best to present both the fundamental ideas
and the stories behind them.

As for the title, ``Heckerman Thinking'' doesn’t have the same 
ring to it as that of Good's book. I chose ``Heckerthoughts,''
because my rather odd last name has been humorous fodder for friends
and colleagues for naming things related to me such as
``Heckerperson,'' ``Heckertalk,'' and ``Heckerpaper''---you get the
idea. Ironically, a distant relative of mine who connected with me via
23andMe is a genealogist, and discovered that my true last name is
“Eckerman,” with my father's ancestors coming from a region in Germany
near the Ecker River. In any case, ``Heckerthoughts'' it is.

\section{Uncertainty and AI's transition to probability} 
\label{sec:probability}

In the early 1980, I joined the AI program at Stanford while also
getting my M.D. degree. I actually went to medical school to learn how
the brain works and was lucky to go to Stanford where AI---an
arguably better way to study the workings of the brain---was
flourishing. I immediately began taking courses in AI and had one big
surprise.  AI, of course, had to deal with uncertainty.  When robots
move, they are not completely certain about their position. When
suggesting medical diagnoses, an expert system may not be certain
about a diagnosis, even when all the evidence was in.  The surprise
was that I was told, very clearly, that probability was not a good
measure of uncertainty. I was told that measures such as the Certainty
Factor model \cite{S74cf,BS84cf} were better. Just coming off of six
years of training in physics, where fluency with probability was a
must, and where the Bayesian interpretation was often mentioned
(although not by name as I recall), I was taken aback by these
statements.

Over the next half a dozen years or so, probability gradually became
accepted by the AI community, and I've been given some credit for
that.  In this section, I will start with a clean and compelling
argument for probability that has nothing to do with my work. Although
everything in this argument was known by the time John McCarthy coined
the term ``Artificial Intelligence'' in 1955, there was no email,
Facebook, or Twitter to spread the word.  Consequently, AI's actual
transition to probability was much more tortured. At the end of this
next section, I discuss that transition and the role that I think I
played.

\subsection{The inevitability of probability}
\label{sec:cox}

Let’s put aside the notion of probability for the moment. Rather,
let’s focus on the important and frequent need to express one’s
uncertainty about the truth of some proposition or, equivalently, the
occurrence of some event. For example, I’m fairly sure that I don’t
have COVID right now, but I am not certain. How do I tell you just how
certain I am? Also, note that I am talking about {\em my}
uncertainty, not anyone else’s. Uncertainty is {\em subjective}. What
properties do we want this form of communication to have?

\bigskip
\noindent
Property 1: A person’s uncertainty about the truth of some proposition
can be expressed by a single (real) number. Let’s call it a degree of
belief and use $b(x)$ to denote the degree of belief that proposition
$x$ is true. At the end of this section, we will discuss the process
of putting a number on a degree of belief. An important point here is
that a degree of belief is not expressed in a vacuum, but rather in
the context of all sorts of information in the person’s head. We say
that all degrees of belief are ``conditional,’’ and write $b(x|\xi)$
to denote a person’s degree of belief in $x$ given their {\em
  background information} $\xi$.  Another point is that, sometimes, we
want to explicitly call out a specific proposition that a degree of
belief is conditioned on. For example, I may want to express my degree
of belief that I have COVID, given that I have no symptoms. I use
$b(x|y,\xi)$ to denote a degree of belief that proposition $x$ is true
given that $y$ is true and given $\xi$.  In much of my past work, I
explicitly mentioned $\xi$, but this notation can get
cumbersome. Going forward, I will exclude it.

\bigskip
\noindent
Property 2: Suppose a person wants to express their belief about the
proposition $xy$, which denotes the “AND” of proposition $x$ and
proposition $y$. One way to do this is to express their degree of
belief $b(xy)$ directly. Another way to express their degrees of
belief $b(x|y)$ and $b(y)$, and then somehow combine them to yield
$b(xy)$. Property 2 says there is a function $g$ that
combines $b(x)$ and $b(y|x)$ to yield $b(xy)$. We write
\begin{displaymath}
b(xy) = g(b(x|y),b(y)).  
\end{displaymath}
A natural, albeit technical, assumption is that the function $g$ is continuous and monotonic increasing in both of its arguments. 

\bigskip
Now let’s think about $b(xyz)$, the degree of belief that three
propositions $x$, $y$, and $z$ are true.  One way to do this is to
decompose $b(xyz)$ in terms of $b(x|yz)$ and $b(yz)$, and then
decompose the latter belief in terms of $b(y|z)$ and $b(z)$:
\begin{displaymath}
  b(xyz) = g(b(x|yz), b(yz)) = g(b(x|yz), g(b(y|z),b(z))).
\end{displaymath}
Another way to do this is to decompose $b(xyz)$ in terms of $b(xy|z)$ and $b(z)$, and then decompose $b(xy|z)$ in terms of $b(x|yz)$ and $b(y|z)$:
\begin{displaymath}
  b(xyz) = g(b(xy|z), b(z)) = g(g(b(x|yz), b(y|z)),b(z)).
\end{displaymath}
It follows that $g$ is associative:
\begin{equation} \label{eq:associative}
g(b(x|yz), g(b(y|z),b(z))) = g(g(b(x|yz), b(y|z)),b(z)).
\end{equation}

\bigskip
\noindent
Property 3: It’s often useful to relate the degree of belief in $x$ to
the degree of belief about the negation of proposition $x$, which I
denote $\bar{x}$. So, another desirable property is that there is a
function $h$ such that
\begin{displaymath}
b(\bar{x}) = h(b(x)).
\end{displaymath}
That is, a degree of belief in $x$ and its negation move in (opposite)
lock step. Another natural, albeit technical, assumption is that $h$
is continuous and monotonic.

\bigskip
Remarkably, it turns out that, if you accept Properties 1 through 3, a
degree of belief must satisfy the following rules:
\begin{displaymath}
  b(xy)=b(x|y) \cdot b(y),
\end{displaymath}
\begin{displaymath}
  b(\bar{x}) = 1-b(x).
\end{displaymath}
where $b({\rm true}) = 1$ and $b({\rm false}) = 0$.
These rules are precisely those of probability! Namely,
\begin{equation} \label{eq:prod}
  p(xy)=p(x|y) \cdot p(y), \ \ {\rm product \ rule},
\end{equation}
\begin{equation} \label{eq:sum}
  p(\bar{x}) = 1-p(x), \ \ {\rm sum \ rule}.
\end{equation}

Cox \cite{Cox46} was the first to prove that properties 1 through 3
imply that a degree of belief must satisfy the rules of probability,
although he used a slightly stronger technical assumptions for the
functions $g$ and $h$. The results here, which assume only continuity
and monotonicity, are based on work by Aczel \cite{Aczel66}, who uses a
set of mathematical tools known as functional equations to show that
functions satisfying simple properties must have particular forms.
Notable is his ``Associativity Equation,’’ where he shows that any
continuous, monotonic increasing, and associative function is
isomorphic to the addition of numbers (see pages 256-267). This is a
remarkable result, as it says that any such function can be
implemented on a slide rule.  It is a key step in proving the
inevitability of probability from Equation~\ref{eq:associative}
and the other properties.

Given these results, the term ``probability’’ gets confusing
quickly. Namely, a probability can be a long-run fraction in repeated
experiments or a degree of belief, but obey the same {\em syntactic}
rules.  It’s tempting to call them both probabilities, but doing so
obscures their very different {\em semantics}. To avoid this
confusion, when the semantics are unclear from the text, I will refer
to degrees of belief as {\em Bayesian} or {\em subjective
  probabilities}, and will refer to long-run fractions in repeated
experiments as {\em frequentist probabilities}. I’ll use $b(\cdot)$,
$f(\cdot)$, and $p(\cdot)$, to refer to a degree of belief,
frequentist probability, or one of the two, respectively.

One pet peeve of mine is that Bayesian thinking---using the rules of
probability to govern degrees of belief---is often confused with Bayes
rule, which follows directly from the syntactic product rule of
probability (Equation~\ref{eq:prod}). That is, Bayes rule can be
applied equally well to Bayesian or frequentist probabilities. In this
manuscript, I’ll try to eliminate this confusion by avoiding use of
the phrase “Bayes rule” and instead refer to “the rules of
probability” or sometimes more precisely ``the product rule of
probability.''

Thomas Bayes was possibly the first person known to apply the
syntactic rules of probability to the semantics of degree of belief
\cite{Bayes}. (I say ``possibly,'' because historical scholars
disagree on this point.) Laplace definitively made this observation
\cite{LaPlace}. Note that Laplace and perhaps Bayes said only that the
rules of probability {\em could} be applied to degrees of belief. They
did not realize that, given Properties 1 through 3, this application
was necessary.

\begin{figure}
\begin{center}
\leavevmode
\includegraphics[width=1.0in]{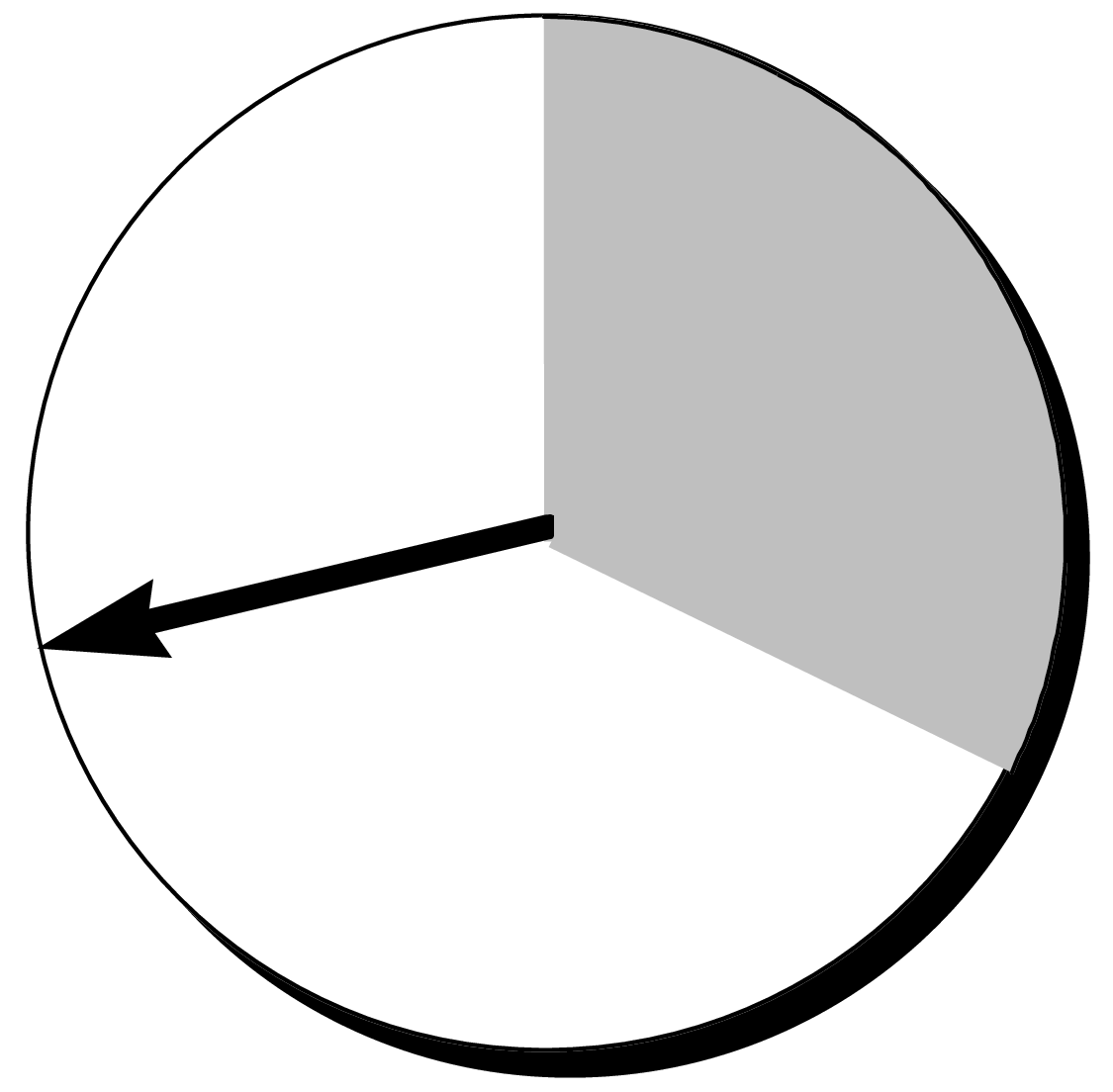}
\end{center}
\caption{The wheel of fortune: a tool for assessing degrees of belief.}
\label{fig:wheel}
\end{figure}

One very practical problem is how to put a number on a degree of
belief. A simple approach is based on the observation that people find
it fairly easy to say that two propositions are equally likely.  The
approach is as follows.  Imagine a simplified wheel of fortune having
only two regions (shaded and not shaded), such as the one illustrated
in Figure~\ref{fig:wheel}.  Assuming everything about the wheel is
symmetric (except for shading), the degree of belief that the wheel
will stop in the shaded region is the fractional area of the wheel
that is shaded (in this case, 0.3). (Of course, given that degrees of
belief are subjective, a person is not forced to make this judgment
about symmetry.)  This wheel now provides a reference for measuring
your degrees of belief for other events.  For example, what is my
probability that I currently have COVID?  First, I ask myself the
question: Is it more likely that I have COVID or that the wheel when
spun will stop in the shaded region?  If I think that it is more
likely that I have COVID, then I imagine another wheel where the
shaded region is larger.  If I think that it is more likely that the
wheel will stop in the shaded region, then I imagine another wheel
where the shaded region is smaller.  Now I repeat this process until
I think that having COVID and the wheel stopping in the shaded region
are about equally likely.  At this point, my degree of belief that you
have COVID is the percent area of the shaded area on the
wheel.

In general, the process of measuring a degree of belief is commonly
referred to as a {\em belief assessment}.  The technique for
assessment that I have just described is one of many available
techniques discussed in the Management Science, Operations Research,
and Psychology literature.  One problem with belief assessment that is
addressed in this literature is that of precision.  Can someone really
say that their degree of belief is $0.601$ and not $0.599$?  In most
cases, no.  Nonetheless, in most cases, degrees of belief are used to
make decisions (I’ll say a lot more about this in
Section~\ref{sec:dm}) and these decisions are typically not sensitive
to small variations in assessments.  Well-established practices of
{\em sensitivity analysis} help one to know when additional precision
is unnecessary ({\em e.g.}, \cite{BlueBook}).  Alternatively, a person can
reason with upper and lower bounds on degrees of belief
\cite{Good83}. Another problem with probability assessment is that of
accuracy.  For example, recent experiences or the way a question is
phrased can lead to assessments that do not reflect a person's true
beliefs \cite{Tversky74}. Methods for improving accuracy can be found
in the decision-analysis literature ({\em e.g.}, \cite{Spetzler75}).

\begin{figure}
\begin{center}
\leavevmode
\includegraphics[width=1.5in]{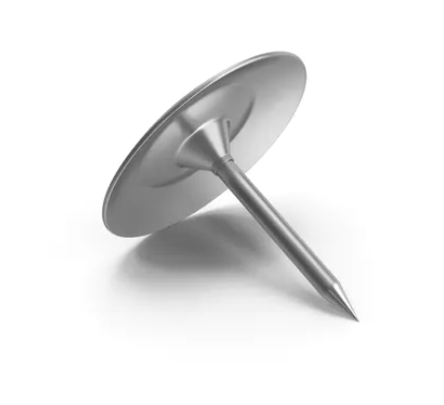}
\end{center}
\caption{A thumbtack having landed heads.}

\label{fig:thumbtack}
\end{figure}

Finally, let’s consider frequentist probabilities and how they are
related to degrees of belief. As we’ve just seen, a similarity is that
both semantics are governed by the rules of probability. And, of
course, there are differences.  For example, assessing a degree of
belief does not require repeated trials, whereas estimating a
frequentist probability does. Consider an old-fashioned
thumbtack---one with a round, flat head (Figure~\ref{fig:thumbtack}).
If the thumbtack is thrown up in the air, it will come to rest either
on its point and an edge of the head (heads) or lying on its head with
point up (tails). In the Bayesian framework, a person can look at the
thumbtack and assess their degree of belief that it will land heads on
a toss. In the frequentist approach, we assert that there is some
unknown frequentist probability of heads, $f$(heads), such that the
thumbtack will land heads on any toss with the same degree of belief.
(Getting heads is said to be {\em independent and identically
  distributed.})  The connection?  This degree of belief must be equal
to $f$(heads).

It turns out that the Bayesian and frequentist frameworks are fully
compatible.  If a person’s beliefs about a series of flips of
arbitrary length has the property that any two sequences of the same
length and the same number of heads and tails are judged to be equally
likely, then the degrees of belief assigned to those outcomes are the
same as the degrees of belief assigned to the outcomes in a situation
where there exists an unknown frequentist probability that is equal to
the degree of belief of heads on every flip. This result was first
proven by de Finetti and generalizations of it followed
\cite{HS55deFinetti}. In short, the frequentist framework can exist
within the Bayesian framework. Ironically, despite this compatibility,
Bayesians and frequentists---people who work exclusively within the
Bayesian and frequentist frameworks, respectively---have been at odds
with each other for many decades.  The good news is that there is no
need to choose---we can embrace both.  And the battles seem to be
subsiding.

\subsection{Independence and graphical models}

In the AI community, before 1985, a common argument made against the
use of probability to reason about uncertainty was that general
probabilistic inference was computationally intractable
\cite{S74cf,BS84cf}. For example, the number of combinations of $N$
binary variables is $2^N$, and no human could assess that many degrees
of belief or reason about them when $N$ is more than a dozen or so.
The only alternative, it was argued, was to make incorrect assumptions
of conditional independence.  For example, when building an expert
system to diagnose medical diseases, the builder could make the
incorrect assumption that all symptoms are mutually independent given
the true disease---the so called, ``naive Bayes assumption.’’

What the community missed was the fact that there were representations
of conditional independence that were ideal for the task of
representing uncertainty.  They were graphical and hence easy to read
and write and would accommodate assertions of conditional independence
that were tailored to the task at hand.

In this manuscript, I will focus on the most used probabilistic
graphical model, the directed acyclic graphical model, or DAG
model. The model was first described by Sewell Wright in 1921
\cite{Wright21} and advanced by many researchers including I.J. Good
\cite{Good83}.

To better understand DAG models, we should now move beyond the simple
notion of a proposition to talk about a variable---a collection of
mutually exclusive and exhaustive values (sometimes called
``states''). I will typically denote a variable by an upper-case,
potentially indexed letter ({\em e.g.}, $X, Y, X_i$, and a value of a
corresponding variable by that same letter in lower case ({\em e.g.},
$x, y, x_i$).  To denote a set of variables, I will use a bold-face
upper-case letter ({\em e.g.}, $\bX, \bY, \bX_i$), and will use a
corresponding bold-face lower-case letter ({\em e.g.}, $\bx, \by, \bx_i$) to
denote a value for each variable in a given set. When showing specific
examples, I will sometimes give variables understandable names and
denote them in typewriter font---for example, \node{Age}.  To denote a
probability distribution over $X$, where $X$ has a finite number of
values, I will use $p(x)$.  In addition, I will use this same term to
denote a probability for a value of $X$.  Whether the mention is a
distribution or a single probability should be clear from context.
Similarly, if $X$ is a continuous variable, then I will use $p(x)$ to
refer to $X$'s probability density function or the density at a given
point.

DAG models can represent either Bayesian or frequentist
probabilities. So, to start, let’s talk about them generically.
Suppose we have a problem {\em domain} consisting of a set of
variables $\bX=\{X_1,\ldots,X_n\}$.  A DAG model for $\bX$ consists of
(1) a DAG structure that encodes a set of conditional independence
assertions about variables in $\bX$, and (2) a set of {\em local
probability distributions}, one associated with each variable.
Together, these components define a probability distribution for
$\bX$, sometimes called a joint probability distribution.  The
nodes in the graph are in one-to-one correspondence with the variables
$\bX$.  I use $X_i$ to denote both the variable and its corresponding
node, and $\Pai$ to denote the parents of node $X_i$ in the graph as
well as the variables corresponding to those parents.  The {\em lack}
of possible arcs in the graph encode conditional independencies.  To
understand these independencies, let’s fix the ordering of nodes in the
graph to be $X_1,\ldots,X_n$. This ordering is not necessary, but it
keeps things simple. Because the graph is acyclic, it is possible to
describe the parents of $X_i$ as a subset of the nodes
$X_1,\ldots,X_{i-1}$. Given this description, the graph states the
following conditional independencies:
\begin{equation} \label{eq:bn-indep}
p(X_i|X_1,\ldots,X_{i-1})=p(X_i|\Pai).
\end{equation}
Note that other conditional independencies can be derived from these
assertions. Now, from (a generalization of) the product rule of
probability, we can write
\begin{equation} \label{eq:gen-prod}
p(\bX) = \prod_{i=1}^n p(X_i|X_1,\ldots,X_{i-1}).
\end{equation}
Combining Equations~\ref{eq:bn-indep} and \ref{eq:gen-prod}, we obtain
\begin{equation} \label{eq:bn-def}
p(\bX) = \prod_{i=1}^n p(X_i|\Pai).
\end{equation}
Consequently, the conditional independencies implied by the graph
combined with the local probability distributions $p(X_i|\Pai),
i=1,\ldots,N$, determine the joint probability distribution for $\bX$.

One other note about the acyclicity of these graphs: Such acyclicity
has been criticized due to its lack of applicability in many
situations, such as when there is bi-directional cause and
effect. That said, if a model is built that includes the
representation of time, where the observation of a quantity over time
is represented with multiple variables, then most criticism disappears.
Also, notably, Sewall Wright and others have considered atemporal
graph models with cycles.

In Section~\ref{sec:lgm}, we will see how DAG models can be
constructed for frequentist probabilities. Here, let’s return to the
state of AI in the 1980s and consider how to create a DAG model for
degrees of belief. Such models became known as {\em Bayesian networks}.
After that, we’ll look at how these models can make
reasoning with uncertainty computationally efficient.

\begin{figure}
\begin{center}
\leavevmode
\includegraphics[width=5.0in]{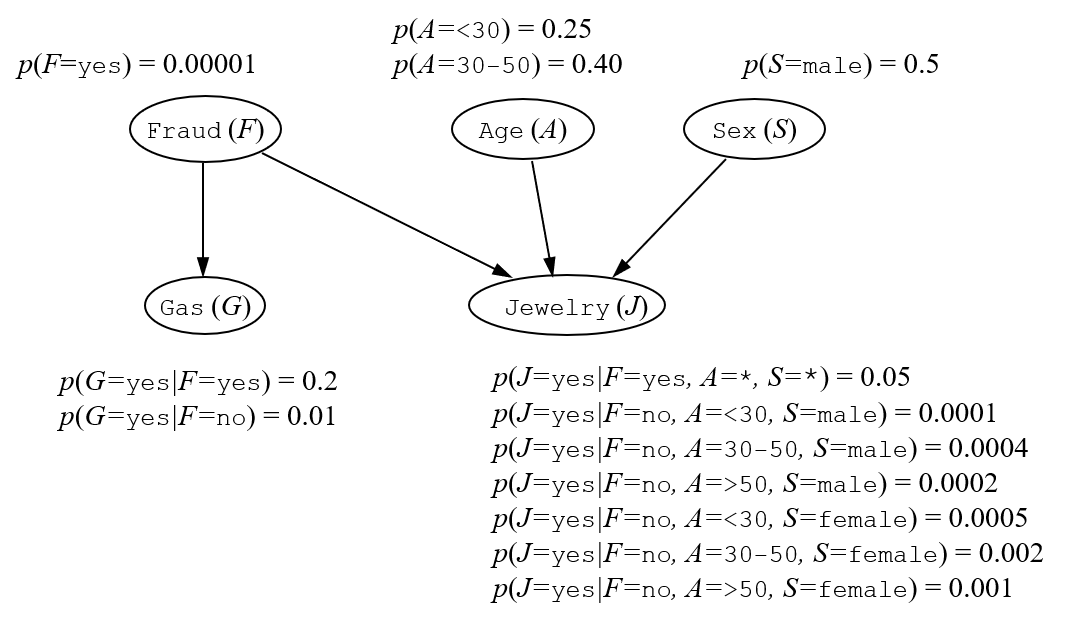}
\end{center}
\caption{A Bayesian network for detecting credit-card fraud.
Arcs are drawn from cause to effect.  The local probability
distribution(s) associated with a node are shown adjacent to the
node.  Asterisks are wild cards---a shorthand for any value.}
\label{fig:fraud}
\end{figure}

To be concrete, let us build a simple Bayesian network for the task of
detecting credit-card fraud.  (Of course, in practice, there is an
individual, typically an expert, who provides the knowledge that goes
into the Bayesian network. Here, I'll simply use ``us'' and ``we''.)
We begin by determining the variables to include in the model.  One
possible choice of variables for our problem is whether or not there
is fraud (\node{Fraud} or $F$), whether or not there was a gas purchase
in the last 24 hours (\node{Gas} or $G$), whether or not there was
a jewelry purchase in the last hour (\node{Jewelry} or $J$), the age of
the cardholder (\node{Age} or $A$), and the sex (at birth) of the
cardholder (\node{Sex} or $S$).  The values of these variables are shown
in Figure~\ref{fig:fraud}.

This initial task is not always straightforward.  As part of this task
we should (1) correctly identify the goals of modeling ({\em e.g.},
prediction versus decision making), (2) identify many possible
findings that may be relevant to the problem, (3) determine what
subset of those findings is worthwhile to model, and (4) organize
the findings into variables having mutually exclusive and
collectively exhaustive values.  Difficulties here are not unique to
modeling with Bayesian networks, but rather are common to most tasks.
Although solutions are not always simple, some guidance is offered by
decision analysts ({\em e.g.}, cite{BlueBook}).

In the next phase of Bayesian-network construction, we build a
directed acyclic graph that encodes our assertions of conditional
independence.  One approach for doing so is to choose an ordering on
the variables, $X_1,\ldots,X_n$, and, for each $X_i$, determine the
subset of $X_1,\dots,X_{i-1}$ such that Equation~\ref{eq:bn-indep}
holds.  In our example, using the ordering $(F,A,S,G,J)$, let’s assert
the following:
\begin{eqnarray} \label{eq:car-ci}
p(a|f) & = & p(a),    \\
p(s|f,a) & = & p(s),  \\
p(g|f,a,s) & = & p(g|f), \\
p(j|f,a,s,g) & = & p(j|f,a,s).
\end{eqnarray}
These assertions yield the structure shown in Figure~\ref{fig:fraud}.

This approach has a serious drawback.  If we choose the variable order
carelessly, the resulting network structure may fail to reveal many
conditional independencies among the variables.  For example, if we
construct a Bayesian network for the fraud problem using the ordering
$(J,G,S,A,F)$, we obtain a fully connected network structure.  In the
worst case, we must explore $n!$ variable orderings to find the
best one.

Fortunately, there is another technique for constructing Bayesian
networks that does not require an ordering.  The approach is based on
two observations: (1) people can often readily assert causal
relationships among variables, and (2) causal relationships typically
reveal a full set of conditional independence assertions.  So, to
construct the graph structure of Bayesian network for a given set of
variables, we can simply draw arcs from cause variables to their
immediate effects.  In almost all cases, doing so results in a network
structure that satisfies the definition of Equation~\ref{eq:bn-def}.  For
example, given the assertion that \node{Fraud} is a direct cause of
\node{Gas}, the assertion that \node{Fraud}, \node{Age}, and
\node{Sex} are direct causes of \node{Jewelry}, and the assertion that
there are no other causes among these variables, we obtain the network
structure in Figure \ref{fig:fraud}.

In the next section, we will explore causality in detail. Here, it
suffices to say that the human ability to naturally think in terms
of cause and effect is in large part responsible for the success of
the DAG model as a representation in AI systems (see
Section~\ref{sec:dm}).

In the final step of constructing a Bayesian network, we assess the
local probability distributions $p(X_i|\Pai)$.  In our fraud example,
where all variables are discrete, we assess one distribution for $X_i$
for every set of values for $\Pai$.  Example distributions are shown
in Figure~\ref{fig:fraud}. Although we have described these
construction steps as a simple sequence, they are often intermingled
in practice.  For example, judgments of conditional independence
and/or cause and effect can influence problem formulation.  Also,
assessments of probability can lead to changes in the network
structure.

One final note: In some situations, a child can be a deterministic
function of its parents.  We can still be uncertain about the value of
such a node, because we can be uncertain about the values of its
parents. Given the values of its parents, however, the value of the
child is determined with certainty. In those situations, I will draw
the child node with a double oval. We will see an example of this
situation when we consider influence diagrams.

Let’s now turn to how we can use this model to reason under
uncertainty.  In this case, we may want to infer the Bayesian
probability of fraud, given observations of the other variables. This
probability is not stored directly in the model and hence needs to be
computed using the rules of probability.  In general, the computation
of a probability of interest given a model is known as {\em
  probabilistic inference}. This discussion on inference applies
equally well to frequentist probabilities, so we can talk about both
at the same time.

Because a DAG model for $\bX$ determines a joint probability
distribution for $\bX$, we can---in principle---use the model to
compute any probability over the its variables.  As an example, let’s
compute the probability of fraud given observations of the other
variables.  Using the rules of probability, we have
\begin{equation} \label{eq:dumb}
p(f|a,s,g,j) = 
  \frac{p(f,a,s,g,j)}{p(a,s,g,j)} =
  \frac{p(f,a,s,g,j)}{\sum_{f'} p(f',a,s,g,j)}.
\end{equation}
For problems with many variables, however, this direct approach is 
computationally slow.  Fortunately, we can use the conditional
independencies encoded in the DAG, Equation~\ref{eq:car-ci}, to make
this computation more efficient.  Equation~\ref{eq:dumb} becomes
\begin{eqnarray*} \label{eq:smart} 
p(f|a,s,g,j) & = & \frac{p(f)\ p(a)\ p(s)\ p(g|f)\ p(j|f,a,s)}{\sum_{f'}
  p(f')\ p(a)\ p(s)\ p(g|f')\ p(j|f',a,s)} \\
& = &
\frac{p(f)\ p(g|f)\ p(j|f,a,s)}{\sum_{f'} p(f')\ p(g|f')\ p(j|f',a,s)}.
\end{eqnarray*}
Many algorithms that exploit conditional independence in DAG models to
speed up inference have been developed (see the annual proceedings of
The Conference on Uncertainty in Artificial Intelligence---UAI).
In some
cases, however, inference in a domain remains intractable.  In such
cases, techniques have been developed that approximate results
efficiently---for example, \cite{GTS94,JGJS99}.

\subsection{AI’s transition to probability}

The representation and use of conditional independence makes reasoning
under uncertainty practical for a wide variety of tasks and, in
hindsight, is largely responsible for AI’s transition to the use of
probability for representing uncertainty. In this section, I highlight
some key steps in the actual transition and the role I think I played
in it.

\begin{figure}
\begin{center}
\leavevmode
\includegraphics[width=3.0in]{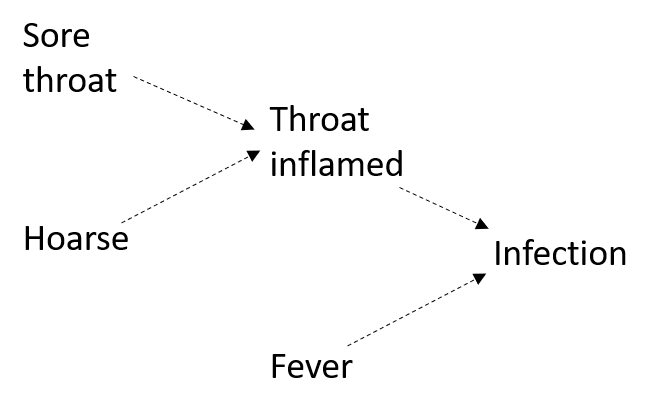}
\end{center}
\caption{A simple Certainty Factor model for diagnosing infection.}
\label{fig:cf-example}
\end{figure}

In the late 1970s and early 1980s, multiple alternatives to
probability were described, including Dempster--Shafer Theory and
Fuzzy Logic. As my PhD advisor was Ted Shortliffe, my greatest
exposure was to the Certainty Factor model, which he had developed for
use in expert systems for medical diagnosis such as MYCIN and EMYCIN
\cite{S74cf,BS84cf}. The expert systems consisted of IF--THEN rules,
which could be combined together in parallel or series, effectively
forming a directed graph. Figure~\ref{fig:cf-example} shows a simple
made-up example for diagnosing infection. An arc from node $e$ to $h$
corresponds to the rule IF $E=e$ THEN $H=h$, where $e$ and $h$ are
propositions that typically correspond to evidence and a hypothesis,
respectively.  An example of series combination is \node{Sore throat}
$\rightarrow$ \node{Throat inflamed} $\rightarrow$
\node{Infection}. An example of parallel combination is \node{Sore
  throat} $\rightarrow$ \node{Throat inflamed} $\leftarrow$
\node{Hoarse}. Importantly, the semantics of the graph associated with
a Certainty Factor model is different from that of the DAG model we
have been considering. I use dashed lines for the former graph to
highlight this difference.  In the Certainty Factor model, arc
direction goes from evidence (typically observed) to hypothesis
(typically unobserved).  In contrast, in a DAG model, arc direction
typically goes from cause to effect.  For some problems, each
hypothesis in the model is a cause of the corresponding evidence, so
that every arc in the Certainty Factor model and DAG model will point
in the opposite direction. There are many instances, however, where
this relationship does not hold.

Associated with each rule $e \rightarrow h$ is a certainty factor
$CF(h,e)$, which was intended to quantify the degree to which $e$ {\em
  changes} the belief in $h$. CFs range from -1 to 1. A CF of 1
corresponds to complete confirmation of $h$; a CF of -1 corresponds to
complete disconfirmation of $H$; and a CF of 0 leaves the degree of
belief of $H$ unchanged. In particular, a CF was not intended to
represent an absolute degree of belief in $H$ given $E$.  The
Certainty Factor model offers functions that combine these CFs when
used in series and parallel. As a result, the model can infer a net
increase or decrease in the degree of belief for a target hypothesis
(\node{Infection} in this example), given multiple pieces of evidence.
A list of properties or ``desiderata'' that these functions should
obey were described.

One day in early 1985, as a teaching assistant for one of Ted’s
medical informatics courses, I was listening to his explanation of the
model. It occurred to me that CFs may simply be a transformation of
probabilistic quantities. A simple candidate in my mind was the
likelihood ratio $\lambda(h,e)=p(e|h)/p(e|\bar{h})$.  From the rules
of probability, we have
\begin{eqnarray} \label{eq:lr}
  \frac{p(h|e)}{p(\bar{h}|e)} & = & \frac{p(e|h)}{p(e|\bar{h})} \ \frac{p(h)}{p(\bar{h})} \\
                              &   & \nonumber \\
                       O(h|e) & = &\lambda(h,e) \ O(h), \nonumber 
\end{eqnarray}
where $O(h|e)$ and $O(h)$ are the prior and posterior odds of
$h$, respectively.  That is, the posterior odds of $h$ are just the
product of the likelihood ratio and the prior odds of $h$.  (I.J. Good
\cite{Good83} wrote extensively about the logarithm of the likelihood
ratio, known as the {\em weight of evidence.}) One catch was that the
likelihood ratio ranges from 0 to infinity, but that was easy to fix:
simply use a monotonic transformation to map the likelihood ratio to
$[-1,1]$.

That night, I went home and gave the mapping $CF(h,e)=
(\lambda(h,e)-1)/ (\lambda(h,e)+1)$ a try. It satisfied the desiderata
and closely matched the series and parallel combination functions in
the Certainty Factor model.  Even more important, the correspondence
revealed that the Certainty Factor model facilitated the expression of
conditional independence assertions that were less extreme than the
naive-Bayes assumption.  (In the example in
Figure~\ref{fig:cf-example}, the resulting conditional independencies
can be identified by reversing the arcs in the figure and interpreting
the result as a DAG model.)  Given my proclivity for probability, I
was excited and wrote up the result \cite{H85cf}
\href{https://arxiv.org/abs/1304.3419}{[arXiv:1304.3419]}.

The work got an interesting mix of reactions.  I was particularly
nervous about what Ted would say about it, but he was very
gracious. When he saw the result, he was very interested and gave me a
set of papers to read to help me structure my thinking and prepare a
publication. When I presented the work at the first UAI meeting that
summer, it got a surprisingly good reception.  At the time, the debate
about probability as a measure of uncertainty was in full swing,
having been ignited by Judea Pearl and Peter Cheeseman at the previous
year’s AAAI Conference.  Those in the probability camp welcomed the
work. Ross Shachter and Jack Breese were perhaps the most surprised
that someone from Stanford, the home of the Certainty Factor model,
would be presenting such a result.  We met for the first time there
and have been friends and collaborators ever since.

I should note that there is an interesting connection between the
Certainty Factor model and the modern reincarnation of the DAG
model. In 1978, Dick Duda and Peter Hart created a variant of the
Certainty Factor model for PROSPECTOR, a computer-based consultation
system for mineral exploration \cite{DGH79prospector}.  Peter Hart went on
to work at Stanford Research Institute, where he met Ron Howard and
Jim Matheson. Ron and Jim saw this model and were inspired to create
influence diagrams, an extension of DAG models that we discuss in the
next section. After that, Judea Pearl heard Ron Howard talk
about influence diagrams and began work his on DAG models
\cite{Pearl88}.

For me, the next question to answer in AI's transition to probability
was the following: Is a monotonic transform of the likelihood ratio
the only measure of a belief update?  Over the next year, I took Ted’s
desiderata, converted them to a series of functional equations (after
learning about what Cox did with degrees of belief), and answered the
question: a monotonic transform of the likelihood ratio is the only
quantity satisfying these desiderata \cite{H86bu}
\href{https://arxiv.org/abs/1304.3091}{[arXiv:1304.3091]}.  The
support for probability was increasing.  That said, critics were still
in the majority.

The key remaining criticism of probability was that no one had built a
successful probabilistic expert system. In 1984, that was about to
change. Bharat Nathwani, a successful surgical pathologist from USC,
visited Ted Shortliffe and several of students with the idea of
building an expert system to aid surgical pathologists in the
diagnosis of disease. The idea was straightforward: a surgical
pathologist would look at one or more slides of tissue from a patient
under the microscope, identify values for features such as the color
or size of cells, report them to an expert system, and get back a list
of possible diseases ordered by how likely they were given what was
seen. Eric Horvitz and I were in that group of students and jumped at
the chance. Over the next year, we built the first version of a
probabilistic expert system for the diagnosis of lymph-node pathology,
called Pathfinder. Even though we knowingly made the inaccurate
assumption that features were mutually independent given disease, the
system worked well---so well, that Bharat, Eric, and I started
Intellipath, a company that would build expert systems for the
diagnosis of all tissue types.

By now it was mid-1985, and fully appreciating Bayesian networks, I
wanted to use the representation to incorporate important conditional
dependencies among features. But there was a big catch: Bharat and I
wanted to extend Pathfinder to include over 60 diseases and over 100
features.  Even if we assumed features were mutually independent given
disease, Bharat would have to assess almost ten thousand 
probabilities.  So, over the next year I developed two techniques
that made the construction of Bayesian networks for medical diagnosis in
large domains easier and more efficient. This work became the focus of
my PhD dissertation \cite{H91book}
\href{https://arxiv.org/abs/1911.06263}{[arXiv:1911.06263]}.

Before describing these techniques, I should note that diagnosis of
disease in surgical pathology is well suited to the assumption that
diseases are mutually exclusive. In particular, it is certainly
possible for a patient to have two or more diseases, but these
diseases are almost always spatially separated and easily identified
as separate diseases.  Both techniques make strong use of this
assumption.  In addition, both techniques make use of the following
point.  Let $d_1$ and $d_2$ denote two diseases, let $X$ denote some
feature ({\em e.g.}, cell size), and suppose that $p(x|d_1)=p(x|d_2)$
for each value of $X$.  From the rules of probability, we see that the
ratio of disease priors and posteriors is the same:
\begin{equation} \label{eq:differentiate}
\frac{p(d_1|x)}{p(d_2|x)} = \frac{p(x|d_1)}{p(x|d_2)} \ \frac{p(d_1)}{p(d_2)} = \frac{p(d_1)}{p(d_2)}.
\end{equation}
In words, $X$ does not {\em differentiate} between the two diseases.
It turns out that surgical pathologists (and experts in other fields)
find it easier to think about whether $X$ differentiates diseases
$d_1$ and $d_2$, than think about whether $p(x|d_1)=p(x|d_2)$.
Comparing Equations~\ref{eq:lr} and \ref{eq:differentiate}, another
  way to think about this situation is that the observation $X=x$, for
  any value $x$, is a null belief update for $d_i$, under the
  assumption that only $d_1$ and $d_2$ are possible.

\begin{figure}
\begin{center}
\leavevmode
\includegraphics[width=4.0in]{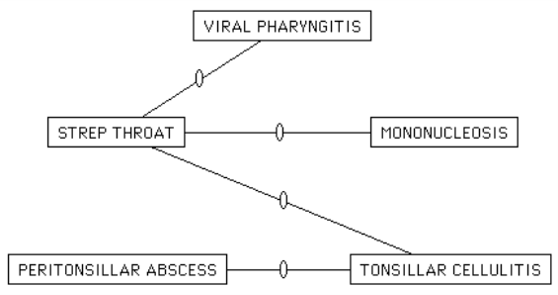}
\end{center}
\caption{A similarity graph for the diagnosis of sore throat.}
\label{fig:sim-graph}
\end{figure}

The first technique was inspired by my discussions with Bharat about
how he related findings to disease. He found it difficult to think
about how a finding differentiated among all diseases. Rather, he
liked to think about pairs of diseases and, for each pair, which of
the many findings differentiated them.  So, I developed the {\em
  similarity graph}, both an abstract concept and a graphical user
interface, to allow him to do just that.  An example for the simpler
task of diagnosing five diseases that cause sore throat is shown in
Figure~\ref{fig:sim-graph}.  Each of the five diseases, which we
assume to be mutually exclusive for purposes of presentation,
correspond to a node in the similarity graph.  An undirected edge
between two diseases denotes that the expert is comfortable thinking
about what features differentiate them.  Note that every disease in
the graph must border at least one edge, because otherwise some
diseases could not be differentiated.  After drawing the similarity
graph, the expert can then click on one of the edges (in the small
oval on the edge) and draw a {\em local Bayesian network structure}, a
Bayesian network structure that includes the findings useful for
differentiating the two diseases and the conditional independencies
among these findings.  Local structures for different disease pairs
need not contain the same findings nor the same assertions of
independence. So, unlike an ordinary Bayesian network structure, the
technique allows for the specification of {\em asymmetric conditional
  independencies}.  In my dissertation, I showed how the collection of
local structures can be combined automatically to form a coherent
(ordinary) Bayesian network structure for the entire domain.  In
addition, I showed how probability assessments for the local networks
can automatically populate the distributions for this whole-domain
structure \cite{H91book}
\href{https://arxiv.org/abs/1911.06263}{[arXiv:1911.06263]}.  The
mathematics is interesting (see Chapter 3 and \cite{GH93}
\href{https://arxiv.org/abs/1611.02126}{[arXiv:1611.02126]}).  I
called the similarity graph combined with the collection of local
Bayesian networks a {\em Probabilistic Similarity Network}.

\begin{figure}
\begin{center}
\leavevmode
\includegraphics[width=5.0in]{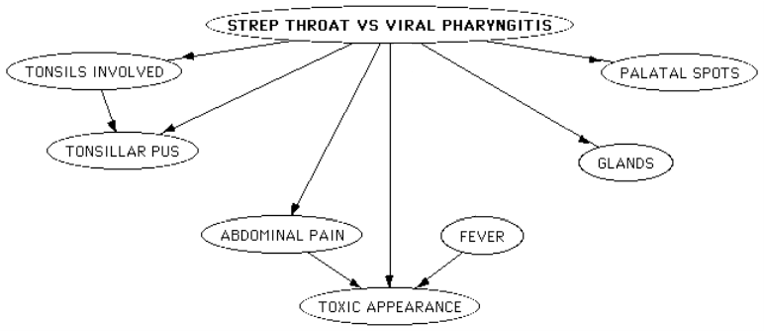}
\end{center}
\caption{A local Bayesian network structure for differentiating strep
  throat from viral pharyngitis.}
\label{fig:sim-local}
\end{figure}

\begin{figure}
\begin{center}
\leavevmode
\includegraphics[width=5.0in]{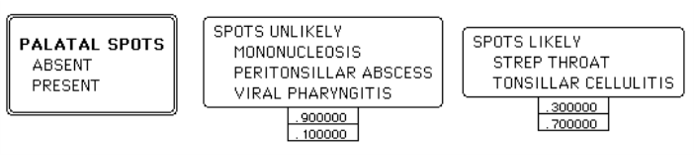}
\end{center}
\caption{A partition for assessing the probability distribution over
  palatal spots given disease.}
\label{fig:partition}
\end{figure}

The second technique I developed made probability assessment even
easier and more efficient than assessment within local structures.
The technique uses a visual structure known as a {\em partition}.  An
example, again for the task of diagnosing sore throat, is shown in
Figure~\ref{fig:partition}.  The box on the left shows a finding to
be assessed (Palatal spots) along with the values for that finding.
Each of the other boxes (in this example, there are two of them)
contain diseases that are not differentiated by the finding.  To
construct this partition, the expert first groups diseases into boxes,
and then, for each box, assesses a single probability distribution for
the finding given any disease in the box.  By
Equation~\ref{eq:differentiate}, there is only one distribution for
each box. This technique generalizes easily to the case where the
probabilities depend on other findings.

Using these techniques, Bharat and I built the extended version of
Pathfinder in a relatively short amount of time and showed that its
probabilistic diagnoses were accurate.  Bharat, Eric, and I then used
these same techniques to build many more expert systems for
Intellipath. Then, on one sunny day in 1986 at Stanford, while Eric
and I were riding our bikes from one class to the next, he said,
``Let’s start a company to build expert systems for any diagnostic
problem.’’ We agreed that Jack Breese should join the venture, and the
company Knowledge Industries was born. Over the next several years,
both Intellipath and Knowledge Industries continued to produce
successful probabilistic expert systems based on Bayesian networks.
With these results and the successful work of many others mostly in
the UAI community, AI was fully embracing probability.

As luck would have it, these efforts brought Eric, Jack, and me to
Microsoft. While finishing my Ph.D. and M.D., I got accepted as an
assistant professor at UCLA---my dream job.  Two weeks after
finishing, I got a call from Nathan Myhrvold, a friend of mine from
high school. He asked that we meet, ostensibly to catch up, and
invited me to join him at Spago, a fancy restaurant in Beverly Hills
(which seemed a bit odd).  When I arrived, he told me about his work
at Yale with Steven Hawking and his rise to vice-president at
Microsoft having overseen version 2 of Windows. He then popped the
question: ``I read your dissertation and want you to come to
Microsoft to apply your work to problems there.’’ Having just landed
my dream job, I said, ``No way, but you can buy Knowledge
Industries.’’  Several months later, he did just that---an acqui-hire
of Jack, Eric, and me.

\section{Decision making and causal reasoning}
\label{sec:dm}

Much of the work I did at Microsoft involved methods for decision
making and its close cousin, causal reasoning. In this section, I
introduce some basic concepts in these areas. In the following
section, I cover some of my work in detail.

\subsection{Decision theory, influence diagrams, and decision trees}

Using probabilities to be precise about one’s uncertainties is
important. It allows us to be precise in communicating our
uncertainties to others. It also helps us make good decisions. To the
second point, decision theory, developed in the middle of the
twentieth century, provides a formal recipe for how to use subjective
probability to make decisions.  By the way, please don’t let the
phrase ``decision {\em theory}’’ scare you. As you will see, it is an
extremely simple and intuitive theory---in essence, it is just common
sense made precise.

\begin{figure}
\begin{center}
\leavevmode
\includegraphics[width=3.0in]{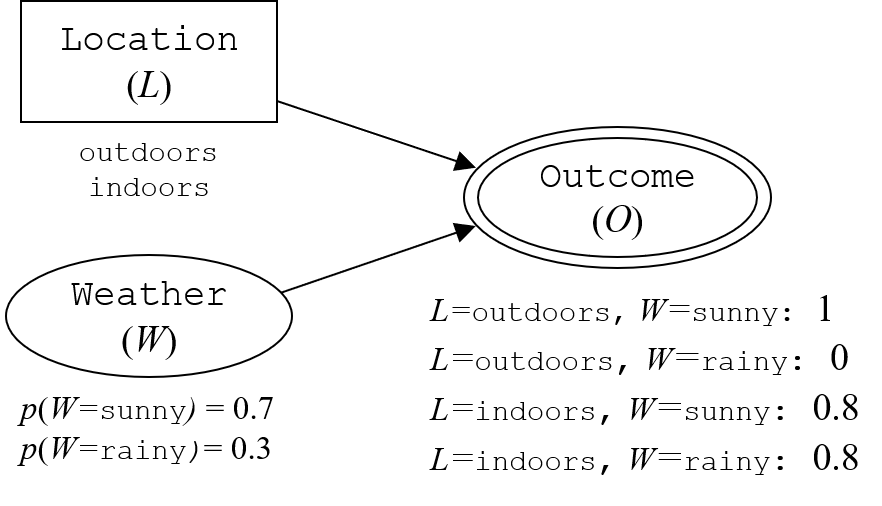}
\end{center}
\caption{An influence diagram for the party problem.}
\label{fig:id-party}
\end{figure}

Let’s start with an example to introduce the basics. Suppose we are
the decision maker (so I don't have to keep saying ``decision maker'')
and are deciding whether to have a party outdoors or indoors on a
given day where the weather can be sunny or rainy.  We can depict this
decision problem using a graphical model known as an influence
diagram, introduced by Ron Howard, my PhD advisor, and Jim Matheson
\cite{HM81id}. The influence diagram for this party problem is shown
in Figure~\ref{fig:id-party}.

In general, an influence diagram always contains three types of nodes,
corresponding to three types of variables. Each type of variable, like
any other variable, has a set of mutually exclusive values.  In our
example, there is one node/variable of each type.  The square node
\node{Location} corresponds to a decision variable. Each value of a
decision variable corresponds to a possible action or alternative---in
this example, \statefont{indoors} and \statefont{outdoors}.  Other
names that have been used for the value of a decision variable include
``act’’ and ``intervention.’’ The oval node \node{Weather} corresponds
to an uncertainty variable (the type of variable seen in DAG
models). In this example, let's suppose this variable has only two
relevant values: \statefont{sunny} or \statefont{rainy}.  Finally,
every influence diagram contains exactly one deterministic outcome
node corresponding to the outcome variable. The values of the outcome
variable correspond to all possible outcomes---that is, all
combinations of alternatives and possible values of the uncertainty
variables.  In this example, the possible outcomes are (1) an outdoor
sunny party, (2) an outdoor rainy party, (3) an indoor sunny party,
and (4) an indoor rainy party.

An influence diagram can include multiple decision nodes, and can
represent observations of uncertainty nodes interleaved with the
making of individual decisions. The latter capability is important to
the process of making real-world decisions, but is not central to the
connection between decision making and causal reasoning.
Consequently, to keep this presentation simple, I will assume that all
decisions are made up front, before any uncertainty nodes are
observed. For details on interleaving observations with decisions, see
the discussion of ``informational arcs'' in \cite{HM81id}.

As in a DAG model, the lack of arcs pointing to uncertainty variables
encode assertions of conditional independence.  In particular, given
an ordering over decision and uncertainty variables,
Equation~\ref{eq:bn-indep} holds for each uncertainty variable, but
now parent variables can include decision variables. Also, based on
our assumption that decisions are made up front, the decision
variables appear first in the variable ordering.  In our party
problem, the lack of an arc from \node{Location} to \node{Weather}
corresponds to the assertion that the weather does not depend on our
decision about the location of the party.  Also as in a DAG model, an
uncertainty variable in an influence diagram is associated with a
local (subjective) probability distribution for each set of values of
the variable’s parents. In our example, \node{Weather} has no parents,
so this variable has only one distribution that encodes our prior
beliefs that it will be rainy or sunny. Together, the assertions of
independence and the local probability distributions define a joint
distribution for the uncertainty variables conditioned on all possible
alternatives.

The single outcome variable in an influence diagram encodes our
preferences for the various outcomes. In particular, associated with
each possible outcome is a number known as a {\em utility}. The larger
the number, the more desirable the outcome. Utilities range between 0
and 1 (we will see why in a moment). That said, if the utilities are
transformed linearly, the choice of the best alternative remains the
same.

Decision theory says the decision maker should choose the alternative
with the highest expected utility, where expectation is taken with
respect to the decision maker's joint distribution over uncertainty
variables given each of the possible alternatives. In the party
problem, expectation is taken with respect to $W$ (weather), which we
assert does not depend on party location. Based on the utilities in
Figure~\ref{fig:id-party}, the expected utility of the outdoor
location is 0.7; the expected utility of the indoor location is 0.8;
so, we should choose the indoor alternative.  This principle is
generally known as the {\em principle of maximum expected utility} or
{\em MEU principle}.

\begin{figure}
\begin{center}
\leavevmode
\includegraphics[width=6.0in]{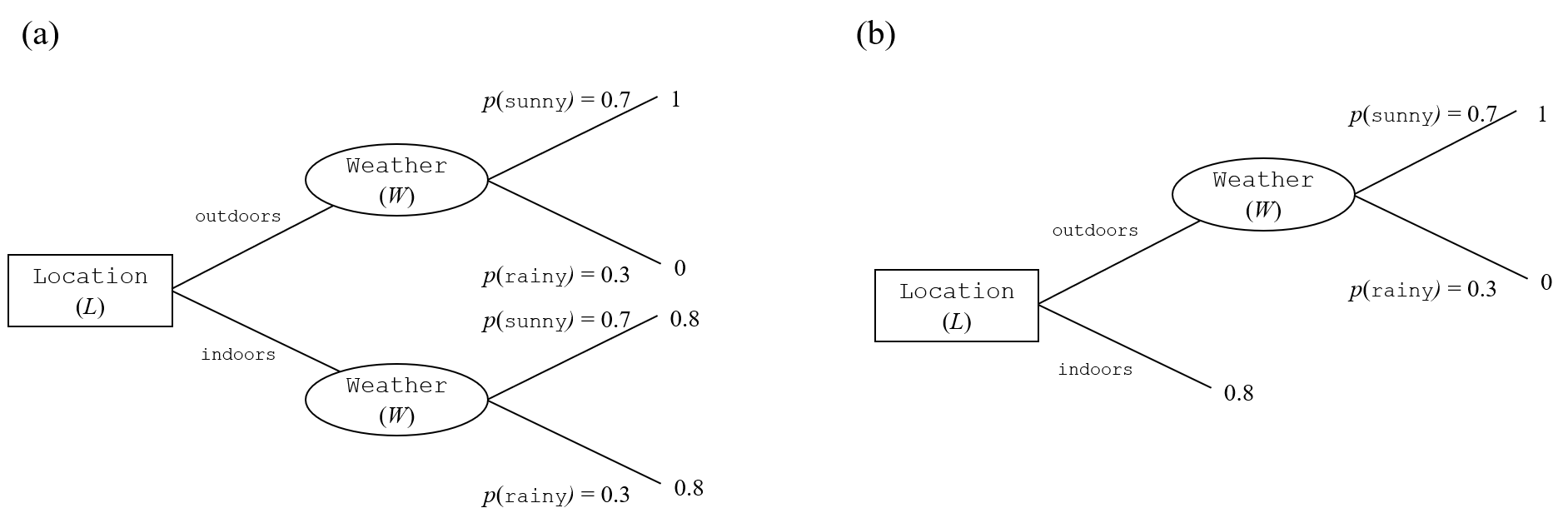}
\end{center}
\caption{(a) A decision tree for the party problem. (b) A simplified version.}
\label{fig:tree-party}
\end{figure}

Decision problems can also be represented with another type of
graphical model known as a {\em decision tree}.  This graphical model
should not be confused with the ``decision tree'' used in the
machine-learning community. Unfortunately, the two communities chose
the same name for these rather different representations. When I use the
term ``decision tree,'' I will make it clear which one I am talking
about.  Figure~\ref{fig:tree-party}a shows the party problem depicted as a
decision tree.

As in the influence diagram, a square node represents a decision
variable, but now the variable’s values are drawn as edges emerging
from the node.  To the right of the decision variable come the
uncertainty variables and possibly additional decision variables.  In
this case, there is just one uncertainty variable and no additional
decision variables. Finally, utilities of the various outcomes are
shown at the far right of the decision tree.  Note that, in this case,
the decision tree can be simplified, because the utility of the indoor
location does not depend on the weather. The simplified decision tree
is shown in Figure~\ref{fig:tree-party}b.

When I first learned of influence diagrams (and decision trees) and
the MEU principle, two questions immediately came up in my mind: What
are these mysterious utilities? And what justifies the use of the MEU
principle to choose the best alternative? In 1948, von Neumann and
Morgenstern provided a very elegant answer to both questions.
In a manner similar to Cox, von Neumann and Morgenstern introduced
several properties that, if followed by a decision maker, define
utilities and imply the MEU principle.  Here, I'll present a slightly
different version of their argument given by my PhD advisor, Ron
Howard \cite{DTproof83}.  I'll use decision trees to present his
argument and assume the rules of probability are in force.

\bigskip
\noindent
Property 1 (orderability): Given any two outcomes, $o_1$ and $o_2$, a
decision maker can say whether they prefer $o_1$ to $o_2$, prefer
$o_2$ to $o_1$, or are indifferent between the two---which I will
write $o_1>o_2$, $o_2>o_1$, and $o_1 \sim o_2$, respectively. In
addition, given any three outcomes $o_1$, $o_2$, and $o_3$, if
$o_1>o_2$ and $o_2>o_3$, then $o_1>o_3$.  That is, preferences are
transitive.  If preferences were not transitive, a decision maker
could be taken advantage of.  For example, if they have the
non-transitive preferences $o_3>o_1$, $o_2>o_3$, and $o_1>o_2$, then
they would pay money to get $o_3$ rather than $o_1$, and again pay
money to get $o_2$ rather than $o_3$, and again to get $o_1$ rather
than $o_2$.  So, they started and ended up with $o_1$, but with less
money. This observation illustrates an important point about this
property and the other properties that we will consider.  Namely, the
properties are ``normative'' or ``prescriptive''---that is, properties
that a decision maker should follow.  It is well known, however, that
people often do not follow these properties when making real decisions
\cite{Tversky74}.  Such decision making is sometimes referred to as
``descriptive.''  Unfortunately, people's descriptive behavior have
taken advantage of---for example, with unscrupulous forms of
advertising.

\begin{figure}
\begin{center}
\leavevmode
\includegraphics[width=3.0in]{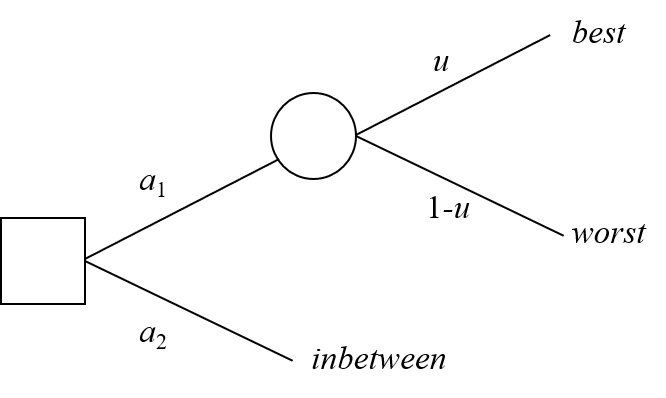}
\end{center}
\caption{A decision problem illustrating the continuity property.  The
  decision maker should have some probability $u$ such that they are
  indifferent between the two alternatives shown.}
\label{fig:continuity}
\end{figure}

\bigskip
\noindent
Property 2 (continuity): Suppose there are three outcomes $best$,
$worst$, and $inbetween$, such that the decision maker has the
preferences $best > inbetween > worst$.  Then the decision maker has
some probability $u$ (with $0<u<1$) such that they are indifferent
between getting $inbetween$ for certain and getting a chance at the
better versus worse outcome with probability $u$.  The situation is
shown in the decision problem of Figure~\ref{fig:continuity}. As is
the case with probability assessment in general, it is impossible in
practice to assess $u$ with infinite precision, and such imprecision
can be managed with sensitivity analysis.

\begin{figure}
\begin{center}
\leavevmode
\includegraphics[width=5.7in]{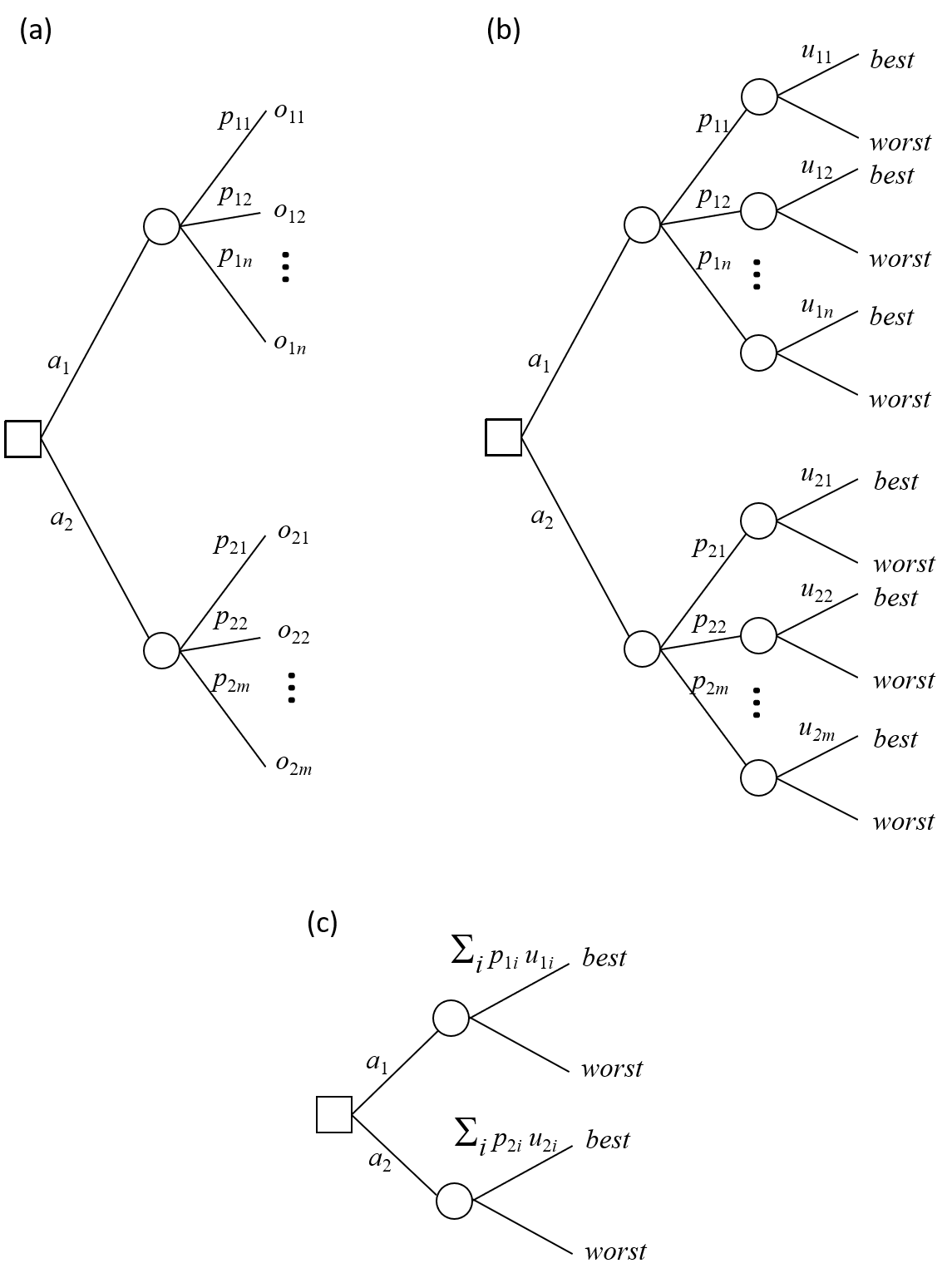}
\end{center}
\caption{(a) A general decision problem with two alternatives. (b) An
  equivalent decision problem based on the properties of continuity
  and substitutability. (c) Another equivalent decision problem based
  on the property of ``no fun in gambling.''}
\label{fig:dt-proof}
\end{figure}

\bigskip
Now consider the rather general decision problem shown in
Figure~\ref{fig:dt-proof}a.  Here, there are two alternatives and, for
each alternative, a set of possible outcomes that will occur with the
probabilities shown.  As we will see, this decision problem is
sufficiently general to make the argument.  Given this decision
problem, the first task of the decision maker is to identify the best
and worst outcome, again denoted $best$ and $worst$,
respectively.  Then, for each possible outcome, the decision maker
uses the continuity property to construct an equally preferred chance
at getting the best versus worst outcome.  The following property says
that the decision maker can substitute this chance for the original
outcome in the original decision problem, yielding the equivalent
decision problem in Figure~\ref{fig:dt-proof}b.

\bigskip
\noindent
Property 3 (substitutability): If a decision maker is indifferent
between two alternatives in one decision problem, then they can
substitute one for the other in another decision problem.

\bigskip
The resulting decision problem is simply a set of chances at the best
versus worst outcome.  The next property says that the decision maker
can collapse these chances into one chance at the best versus worst
outcome using the rules or probability.

\bigskip
\noindent
Property 4 (decomposability): Given multiple chances at outcome $o_1$
versus $o_2$ with probabilities $p_1,\ldots,p_k$, the decision maker
should equally prefer this situation to a single chance at outcome
$o_1$ versus $o_2$ with probability $\sum_i p_i$.  Ron calls this
property the ``no fun in gambling'' property.

\bigskip
Given this property, the decision maker now has the equivalent
decision problem shown in Figure~\ref{fig:dt-proof}c.  The final
property describes how the decision maker should make this decision.

\bigskip
\noindent
Property 5 (monotonicity): Given two alternatives $a_1$ and $a_2$
between a chance at two outcomes $o_1>o_2$, the decision maker should
prefer the alternative $a_1$ if and only if the probability of getting
$o_1$ in $a_1$ is higher than the probability of getting $o_1$ in
$a_2$.

\bigskip
Applying this property to the decision problem in
Figure~\ref{fig:dt-proof}c yields the MEU principle, where the utility
of an outcome is the probability of the chance at outcome $best$
versus $worst$.  In addition, given the substitutability and
decomposability properties, the decision problem in
Figure~\ref{fig:dt-proof}a is sufficiently general to address any
decision problem with two alternatives.  Finally, there is a
straightforward generalization of this argument to a decision problem
with any number of alternatives.

In essence, these properties allow a decision maker to boil down any
decision problem into a series of simple decision problems like the
one in Figure~\ref{fig:continuity}.  Note that this simple decision
problem has the same form as the party problem.  In the party problem,
the decision maker is asked to assess utilities and apply the MEU
principle to choose.  The clever trick from von Neumann and
Morgenstern is to recast this problem into one of choosing a
probability of \node{sunny} that makes the alternatives
\node{outdoors} and \node{indoors} equally preferable.

When I first saw this argument, I thought to myself, “This is the most
elegant and practical application of math I have ever seen.”  Both the
assumptions and the proof of the MEU principle are simple, and the
result is a tool that just about anyone can use to improve their
decision making.  This realization came to mind because my father was
a high school math teacher and was always looking for practical
applications of math to show his students.  By the time I saw this
proof, he had already retired, but nevertheless, we were both excited
to discuss it.

To close this section, it’s worthwhile to comment on the two rather
different graphical models for decision problems. The decision tree
more naturally depicts the flow of time and the display of
probabilities and utilities. Also, it allows the explicit depiction of
any asymmetries in a decision problem, as we see in our example.  In
contrast, an influence diagram is typically more compact and, as
mentioned, enables the explicit representation of probabilistic
independencies.  We will see examples of this benefit later.  Given
these tradeoffs, both representations are still in use.

\subsection{Proper scoring rules}

There is a special class of utility that is worth mentioning---namely,
proper scoring rules. To motivate this class of utility, imagine you
operate a company that invests is new and upcoming businesses. You are
constantly making high-stakes decisions and sometimes need
probabilities (or probability densities) from experts that various
possible outcomes will occur, such as whether a particular piece of
legislation in Washington is going to pass.  You have access to
experts, and want to be sure that the probabilities you get from them
are ``honest.''  One way to do this is to frame the expert as a
decision maker who is deciding what distribution to report over some
variable $X$, and then receiving a utility for that report based on
the value of $X$ that is eventually observed.  The utility they
receive should be a function of (1) every possible observed value of
$X=x$ and (2) the expert's reported distribution $q(x')$ for every
possible value of $X=x'$. Let's denote this utility function as
$u(x,q(x'))$.  For the reported probability distribution $q$ to be
honest, we want the expert's expected utility to be a maximum when $q$
is equal to the expert's actual probability distribution $p$.  That
is, we want $\int_x p(x) \ u(x,q(x'))$ to be a maximum when $q=p$.
(Here, I am using $\int_x$ generally---it can be a sum when $X$ is
discrete.)  When this condition holds, $u$ is said to be a {\em proper
  scoring rule.}

One proper scoring rule is the {\em log scoring rule}:
$u(x,q(x'))=\log q(x)$---that is, $u(x,q(x'))$ is the logarithm of
$q(x)$ where $x$ is the value of $X$ that is actually observed. It is
easy to check that utility is maximized when $p=q$.  What is rather
remarkable, however, is that the log scoring rule is the only proper
scoring rule (up to linear transformation) when (1) $X$ takes on three
or more states, (2) $u(x,q(x'))$ is {\em local} in the sense that it
only depends on $q$ at the observed value of $X$, and (3) certain
regularity conditions hold \cite{Bernardo79,BS94}. The concept of a
proper scoring rule will come in handy in
Section~\ref{sec:learn-one-var}, where we consider learning graphical
models from data.

\subsection{Correlation versus causation and a definition of causality}

It’s tempting to sum up the difference between a purely probabilistic
DAG model and an influence diagram by viewing the latter as merely the
former with decisions and utilities “tacked on.”  Nonetheless, there
is something fundamentally different about them. In particular, an
influence diagram can express {\em causal} relationships among its
variables, whereas a probabilistic graphical model strictly represents
{\em correlational} or {\em acausal} relationships among its
variables.

Ironically, Howard and Matheson, have been reluctant to associate
their elegant representation for decision problems with any notion of
causality. Note the label ``influence diagrams’’ rather than
``causal diagrams.’’ This reluctance had also been prominent in the
statistics community until pioneers including Peter Spirtes, Clark
Glymour, Richard Scheines, Jamie Robbins, and Judea Pearl began their
relentless work to change that.

Perhaps one source of reluctance was the lack of a clear definition of
cause and effect. Even in his recent work \cite{Pearl19why}, Judea
describes the notion of cause and effect as a {\em primitive}, like
points and lines in geometry. Personally, I think causality can and
should not be left as a primitive, but should be defined in more basic
terms. Before I go any further, allow me to take a crack at this.

Let’s start with a simple example: colliding billiard balls.  When a
moving ball strikes another, we naturally say that the collision
``caused’’ the subsequent motion. (At a party in the mid-1980s for Ted
Shortliffe's research group, I was playing Pictionary.  Eric Horvitz,
my game partner, drew a picture of a billiard table with a pool cue
about to strike a ball. I immediately guessed the answer:
``causality.’’ The other players were astounded. Eric and I think a
lot alike.) This easily generalizes.  When we say, ``event x causes
event y,’’ we mean that there are physical interactions, via the
forces of nature, that relate event x to event y.  \comment{new} Note
that this definition {\em may} extend to quantum interactions.
Current experimental results involving quantum entanglement, perhaps
the oddest of quantum interactions, suggest two possibilities. One is
that special relativity holds and thus there is correlation without
causation, which is extremely strange
\cite{Glymour06,Gisin14}.  The other is that special
relativity does not hold and a quantum signal that governs the
entanglement is propagating faster than the speed of light. Under this
second possibility, my definition still applies. Unfortunately, it is
quite possible that we may never know which possibility holds. That
said, to me, the concepts in Newtonian physics and classical chemistry
are adequate to describe almost all causal relationships that have
practical relevance.

With this definition, it becomes clear that reasoning about actions
and their consequences is tantamount to causal reasoning. For example,
my decision to hold the party outdoors leads to me writing
invitations, which in turn are read by my guests, who then follow
those instructions and show up at my house---all happening via forces
of nature.  Note that, although there are many definitions of cause
and effect, there is general agreement on how to test for the
existence of a causal effect and to quantify its strength: the
randomized trial. We examine this test in Section~\ref{sec:confounders}.

\subsection{Fully causal models and the force/set/do decision} \label{sec:fullycausalmodels}

\begin{figure}
\begin{center}
\leavevmode
\includegraphics[width=3.0in]{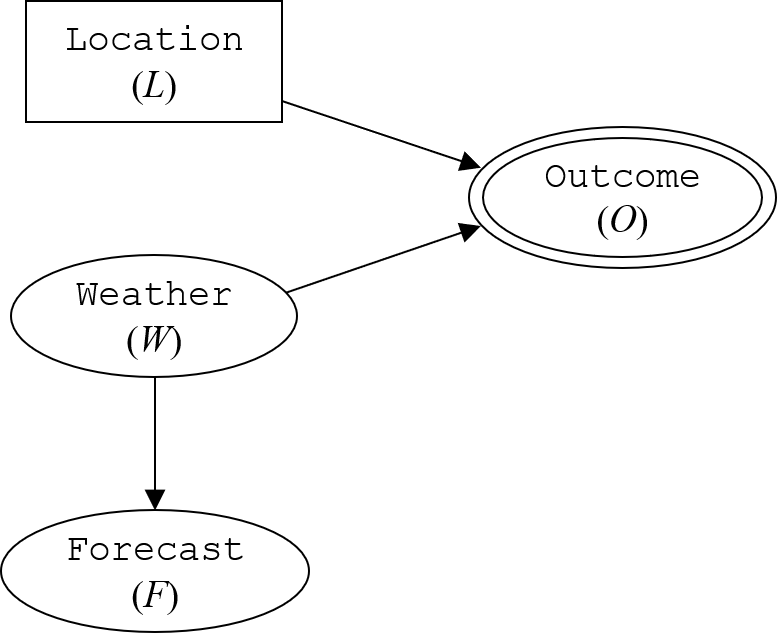}
\end{center}
\caption{An influence-diagram structure for the party problem where
  the decision maker can check a weather forecast before deciding on
  the location.}
\label{fig:id-forecast}
\end{figure}

An influence diagram can represent both causal and acasual
relationships.  In the party problem, suppose that the decision maker
can check a weather forecast before making the location decision. The
updated influence diagram (structure only) is shown in
Figure~\ref{fig:id-forecast}.  Here, the relationship from
\node{Location} and \node{Weather} to \node{Outcome} is causal,
whereas the relationship between \node{Weather} and \node{Forecast}
need only be considered in terms of probabilities.  Of course, we
could model a causal relationship between \node{Weather} and
\node{Forecast} by introducing other variables such as the current
weather and its time derivatives at nearby locations, but doing so is
not needed to decide on the location of the party.  To make the
decision, only the probabilistic relationship between \node{Weather}
and \node{Weather forecast} matters.  For a detailed discussion of the
flexibility of influence diagrams to represent both causal and acausal
relationships, see \cite{HS95cause}
\href{https://doi.org/10.1613/jair.202}{[10.1613/jair.202]}.

\comment{new}

In contrast, many researchers working on the theory of causal
inference focus on models where every arc is causal.  In addition,
they don't represent explicit decisions, but rather reason about
decisions that are implicitly defined by the domain variables. In
particular, for every domain variable, they imagine a decision with
alternatives that either force the variable to take on a particular
value, or take no action and simply observe the variable. I call this
framework a {\em fully causal framework} and a model therein a {\em
  fully causal model}.

\begin{figure}
\begin{center}
\leavevmode
\includegraphics[width=5.0in]{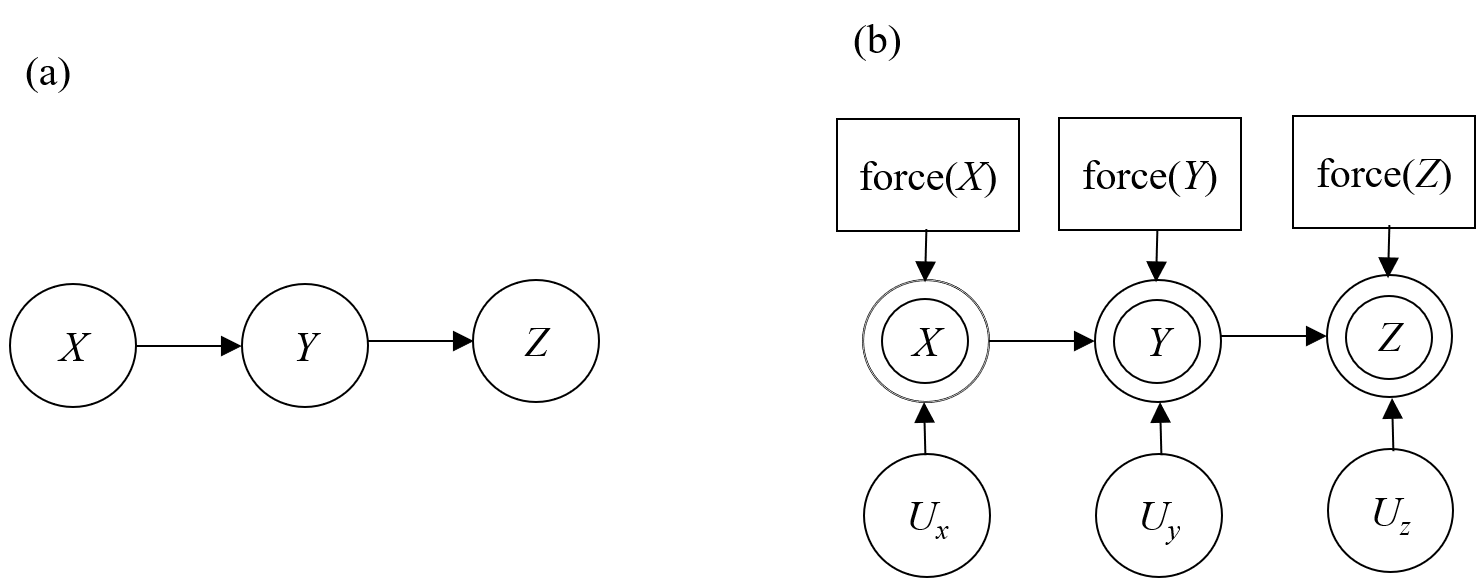}
\end{center}
\caption{A fully causal model structure asserting that $X$ causes $Y$
  and $Y$ causes $Z$ shown (a) in simple form and (b) as an
  influence-diagram structure.}
\label{fig:xyz}
\end{figure}

The fully causal framework lies strictly within the decision-theoretic
framework.  In particular, a fully causal model within this framework
can be represented by an influence diagram in a straightforward
manner.  As an example, consider the fully causal model for the
assertions ``$X$ causes $Y$’’ and ``$Y$ causes $Z$’’. The model
structure can be drawn in {\em simple form} as a DAG on the domain
variables as shown in Figure~\ref{fig:xyz}a. It can also be
represented by the influence-diagram structure shown in
Figure~\ref{fig:xyz}b.

First, let's look at the decision variables in this influence diagram.
To be concrete, consider the decision variable force$(Z)$; similar
remarks apply to force$(X)$ and force$(Y)$.  One value of the decision
variable force$(Z)$ is ``do nothing’’, whereby $Z$ is observed rather
than manipulated. Other values of force$(Z)$ corresponding to $Z$
taking on one of its possible values, regardless of the values of
$Z$’s parents in the graph.  Pratt and Schlaifer \cite{PS88} describe
this action as ``forcing'' $X$ to take on some value.  Note the
connection between the verb ``force'' and the noun ``force'' in my
definition of causality. Spirtes, Glymour, Scheines also describe this
type of action, calling it a ``manipulation'' in \cite{SGS93} and
``set$(X)$'' in various other writings.  Pearl calls it a ``do
operator'' and uses the notation ``do$(X)$.''  In addition, he
sometimes talks about $Z$ ignoring its parents as {\em graph surgery}
\cite{Pearl00,Pearl19why}.

Now let's look at the remaining variables. Each observed variable
($X$, $Y$, and $Z$) is a deterministic function of its corresponding
force/set/do decision and its corresponding uncertainty variables
($U_x$, $U_y$, and $U_z$) that encode the uncertainties in the
problem.  When force$(Z)$ takes on the value ``do nothing,'' then
$Z=U_z$.  When, for example, force$(Z)$ takes on the value ``force $Z$
to value $z$,'' then $Z=z$, as described in the previous paragraph.
Typically, as depicted in the figure, we assume these uncertainty
variables are mutually independent, reflecting the modularity of
causal interactions (which I will discuss shortly). As for the outcome
variable, I've omitted it, because here we are not concerned with
making the actual decisions, but rather representing the relationships
among the variables.

The influence diagram for a fully causal model not only represents
causal interactions, it also represents assertions of conditional
independence. For example, given the fully causal model in
Figure~\ref{fig:xyz}b, when we set all force/set/do decisions to ``do
nothing'' and absorb the uncertainties into the observable variables,
we end up with a probabilistic DAG model with the same independence
semantics as the model in simple form (Figure~\ref{fig:xyz}a) 
interpreted as a probabilistic model. In general, a fully causal model
implies the same assertions of conditional independence as those
implied by its simple form interpreted as an acausal model. This
statement is sometimes called the {\em causal Markov assumption}
\cite{SGS93} and, as we shall see, is important for learning causal
models from observational data.

The fully causal framework makes certain forms of causal reasoning
accessible to simple mathematical treatment (see, {\em e.g.,} the
section in \cite{Pearl19why} starting on p.231). In addition, when
creating a model within this framework, a person need not specify up
front whether a variable is manipulated or observed. That said, it’s
important to note that force/set/do decisions are not the norm in real
life. A classic example where a decision is intended to be a
force/set/do decision, but isn't, is making a wish to a magic genie:
\begin{quotation}
\noindent Me: ``Genie, I wish to be rich.’’ \newline
Genie: ``Your wish is granted. I killed your aunt, and you will now inherit her fortune.’’ \newline
Me: ``No, that’s not what I meant!’’
\end{quotation}
The genie acts to increase the value of my bank account, but his
actions cause other changes that I did not intend. (When I was a very
young, I watched The Twilight Zone television series, which left a
deep impression on me. Season 2, episode 2 has a variation on the
magic-genie theme. The protagonist wishes to be all-powerful, and the
genie turns him into Hitler in his final moments in the bunker---very
disturbing.  When I first heard of the force/set/do decision, I
couldn’t help but think of this episode.)  Nonetheless, there are
occasional situations where force/set/do decisions can be realized in
practice (see Sections~\ref{sec:lid} and \ref{sec:confounders}).

\subsection{The modularity of cause--effect relationships}

A useful property of cause--effect relationships is that they tend to
be modular.  To illustrate this point, consider a clinical trial to
test the efficacy of drugs in lowering cholesterol. A reasonable
causal model structure is as follows:

\begin{displaymath}
  \node{Drug offered} \rightarrow \node{Drug taken} \rightarrow \node{Cholesterol level}
\end{displaymath}

\noindent
Now, suppose there are two components or ``arms'' of the trial, each
testing a different drug.  In this situation, it is quite possible
that the relationship between drug taken and cholesterol level will be
different in the two arms.  In contrast, assuming the side
effects of the two drugs are the same, it is reasonable to assume that
the relationship between drug offered and drug taken is the same in
both arms.  In this case, the causal interactions involving a person's
proclivity to take a drug when asked to do so are the same for the two
arms of the trial.

It is interesting to speculate that the modularity of causal reasoning
has a potential survival advantage for humans.  Humans excel above all
other creatures in large part because they can make good decisions.
As we've just discussed, causal reasoning is a key component of
decision making, and the modularity of causal relationships allows
this reasoning to be more easily executed in the human brain.
Furthermore, the high-level process of decision making is itself
modular---we decompose the process into selecting alternatives,
specifying uncertainty in the face of interventions, and specifying
preferences for the possible outcomes.  This additional modularity
further facilitates the execution of decision making in the brain.
Furthermore, the ability to imagine possible alternatives
necessarily requires at least the perception of free will. Whether we
actually have free will is a matter of debate. Sadly, experimental
evidence suggests it is an illusion \cite{Hallett16}.  Ross Shachter
and I discuss these points in more detail in \cite{SH22why}
\href{https://doi.org/10.1145/3501714.3501756}{[10.1145/3501714.3501756]}.

One final note: individual decision making is better than group
decision making. Specifically, a single decision maker has {\em a} set
of beliefs and {\em a} set preferences that prescribe an optimal
decision. This is not the case for groups. One merely needs to
experience a group trying to decide where to go out to eat or witness
the decision paralysis of political bodies to appreciate this point.
(Also, see Arrow's impossibility theorem \cite{Arrow50}.)  Going down
the rabbit hole one step further, scientific investigations suggest
that the human thought proceeds as a collection of subconscious
processes that feed into a sense of a single, conscious “I” ({\em
  e.g.,} \cite{Hallett16}). If these processes did not feed into a
single ``I,'' then decision making would suffer in a manner similar to
that of group decision making, leading to a survival
disadvantage. Thus, it is not too surprising that the sense of a
single, conscious “I” would emerge through evolution. I’m not
suggesting that a single “I” must be conscious to have this advantage,
but it seems clear that if evolution were to produce such a conscious
``I,'' then it would have a survival advantage.

\section{Learning graphical models from data}
\label{sec:lgm}

Around 1988, as work and applications in probabilistic expert systems
was in full swing, Ross Shachter and I were sitting in his office one
afternoon, brainstorming about what should come next. We came up with
the idea of starting with a Bayesian network for some task and using
data to improve it. Neither of us ran with the idea, but several years
later, Cooper and Herskovits wrote a seminal paper on the topic
\cite{CH91}.  The work was very promising. When I got to Microsoft
Research in 1992, I set out to see how I could contribute. My
colleagues and I, especially Dan Geiger, Max Chickering, and Chris
Meek, managed to make a lot of progress over the next decade ({\em
  e.g.}, \cite{HGC95ml}
\href{https://doi.org/10.1023/A:1022623210503}{[10.1023/A:1022623210503]}
and \cite{CHM04np-hard}
\href{https://dl.acm.org/doi/10.5555/1005332.1044703}{[10.5555/1005332.1044703]}.) \comment{for some reason, the standard doi link does not work
  for this paper} In this section, I highlight many of the fundamental
concepts that emerged from this work and, as usual, provide an index
to the key papers.

Imagine we have data for a set of variables (the domain) and
would like to learn one or more graphical models from that data---that
is, identify one or more graphical models that are implicated by the
data.  The data we have may be purely observational or a mix of
interventions and observations. There are two rather different
approaches that have been developed over the last four decades to do
so. In the first approach, the frequentist methods are used to test for
the existence of various conditional independencies among the
variables. Then, graphs consistent with those independencies are
identified. In the second approach, a Bayesian approach, we build a
prior distribution over possible graph structures and prior
distributions over the parameters of each of those structures. We then
use the rules of probability to update these priors, yielding a
posterior distribution over the possible graph structures and their
parameters. We can then average over these Bayesian networks, using
their posterior probabilities, to make predictions or decisions.

Here, I will focus on the Bayesian approach, frankly, because it is
what I have worked on for many years. That said, the reason I have
worked on the approach for so many years is that it is principled.  In
contrast, the frequentist approach requires a rather arbitrary
threshold to test for independence. I will start with the task of
learning acausal DAG models (where we just care about the expression
of independence assertions) and then move to learning fully causal
models and influence diagrams.

\subsection{Learning an acausal DAG model for one finite variable}
\label{sec:learn-one-var}

To begin a discussion of learning acausal DAG models, let’s look at
the simplest possible problem: learning a DAG model for a single
binary variable $X$ with observational data only. To be concrete,
let’s return to the problem of flipping a thumbtack, which can come up
heads (denoted $x$) or tails (denoted $\bar{x}$). We flip the
thumbtack $N$ times, under the assumption that the physical properties
of the thumbtack and the conditions under which it is flipped remain
stable over time, yielding data $D={X_1,\ldots,X_N}$. From these $N$
observations, we want to determine the probability of (say) heads on
the $N+1$th toss. As we have discussed, there is some unknown
frequency of heads.  We could use $f$(heads) to denote this frequency,
but to avoid the accumulation of parentheses, let’s use $\theta_x$
instead (and similarly $\theta_{\bar{x}}$).  In the context of
learning graphical models from data, $\theta_x$ and $\theta_{\bar{x}}$
are often referred to as {\em parameters} for the variable $X$. In
addition, I will use $\theta_X$ to apply to the case where $X$ is
either $x$ or $\bar{x}$.

\begin{figure}
\begin{center}
\leavevmode
\includegraphics[width=4.0in]{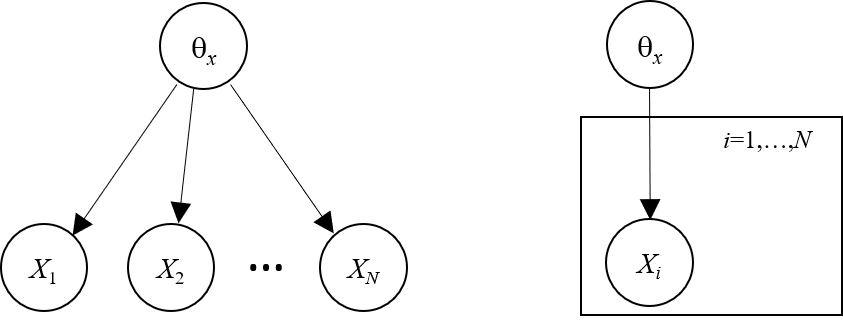}
\end{center}
\caption{The figure on the left-hand side is a DAG model stating that
  the observations of $X$ are mutually independent given
  $\theta_X$. The figure on the right-hand side is same model depicted
  in plate notation.}
\label{fig:binomial}
\end{figure}

As we are uncertain about $\theta_x$, we can express this uncertainty
with a (Bayesian) probability distribution (in this case, a
probability density function) $p(\theta_x)$. The independence
relationships among the variables $\theta_x$ and $X_1,\ldots,X_N$ can
be described by the Bayesian network, shown on the left-hand side of
Figure~\ref{fig:binomial}.  Namely, as we discussed at the end of
Section~\ref{sec:cox}, given $\theta_x$, the values for the
observations $X_1,\ldots,X_N$ are mutually independent. The right-hand
side of Figure~\ref{fig:binomial} is the same model depicted in {\em
  plate notation}, first described by \cite{Buntine94}. Plates depict
repetition in shorthand form. Anything inside the plate ({\em i.e.},
the box) is repeated over the index noted inside the plate.

Learning the parameter for the one-node Bayesian network now boils
down to determining the posterior distribution $p(\theta_x|D)$ from
$D$ and the prior distribution $p(\theta_x)$.  From the rules of
probability, we have
\begin{equation} \label{eq:bt}
p(\theta_x|D) = \frac {p(\theta_x) \ p(D|\theta_x)}{ p(D) }.
\end{equation}
where
\begin{displaymath} \label{eq:ml}
p(D) = \int p(D|\theta_x) \ p(\theta_x) \ d\theta_x.
\end{displaymath}
The quantity $p(D)$ is sometimes called the {\em marginal likelihood}
of the data. As we will see, this quantity is important for inferring
posterior probabilities of model structures.  From the independencies in
Figure~\ref{fig:binomial}, we have
\begin{displaymath}
p(D|\theta_x) = \prod_{i=1}^N p(x_i|\theta_x),
\end{displaymath}
where each term in the product is $\theta_x$ when $X_i$ is heads and
($1-\theta_x$) when $X_i$ is tails.  Both Bayesians and frequentists
agree on this term: it is the likelihood function for Bernoulli
sampling.  Equation~\ref{eq:bt} becomes
\begin{equation} \label{eq:bt2}
p(\theta_x|D) = \frac{ p(\theta_x) \ \theta_x^{N_x} \ (1-\theta_x)^{N_{\bar x}} }{p(D)},
\end{equation}
where $N_x$ and $N_{\bar x}$ are the number of heads and tails
observed in $D$, respectively.  The quantities $N_x$ and $N_{\bar x}$
are sometimes called {\em sufficient statistics}, because they provide
a summarization of the data that is sufficient to describe the
data under the assumptions of their generation.

Now that we have the posterior distribution for $\theta_x$, we can
determine the probability that the $N+1$th toss of the thumbtack will
come up heads (denoted $x_{N+1}$) heads:
\begin{eqnarray*} \label{eq:bx}
p(x_{N+1}|D) & = & \int p(x_{N+1}|\theta_x) \
          p(\theta_x|D) \ d\theta_x \\ 
 & = & \int \theta_x \ p(\theta_x|D) \ d\theta _x
       \equiv {\rm E}_{p(\theta_x|D)}(\theta_x). 
\end{eqnarray*}
where ${\rm E}_{p(\theta_x|D)}(\theta_x)$ denotes the expectation of
$\theta_x$ with respect to the distribution $p(\theta_x|D)$. The
probability that the $N+1$th toss comes up tails is simply
\begin{displaymath}
p(\bar{x}_{N+1}|D)=1-p(x_{N+1}|D).
\end{displaymath}

\begin{figure}
\begin{center}
\leavevmode
\includegraphics[width=4.0in]{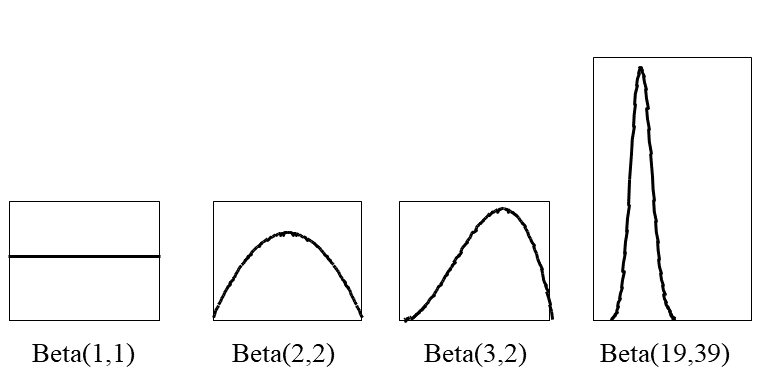}
\end{center}
\caption{Several beta distributions.}
\label{fig:beta}
\end{figure}

To complete the Bayesian story for this example, we need a method to
assess the prior distribution over $\theta_x$.  A common approach,
often adopted for convenience, is to assume that this distribution is
a {\em beta} distribution:
\begin{displaymath} \label{eq:beta}
p(\theta_x) = {\rm Beta}(\theta_x|\alpha_x,\alpha_{\bar{x}}) \equiv
  \frac{\Gamma(\alpha_x)}{\Gamma(\alpha_x) \Gamma(\alpha_{\bar{x}})} \theta_x^{\alpha_x - 1}
  (1-\theta_x)^{\alpha_{\bar{x}}-1},
\end{displaymath}
where $\alpha_x>0$ and $\alpha_{\bar{x}}>0$ are the parameters of the
beta distribution, $\alpha_X=\alpha_x+\alpha_{\bar{x}}$, and
$\Gamma(\cdot)$ is the {\em Gamma} function which satisfies
$\Gamma(x+1)= x \Gamma(x)$ and $\Gamma(1)=1$.  The quantities
$\alpha_x$ and $\alpha_{\bar{x}}$ are often referred to as {\em
  hyperparameters} to distinguish them from the parameter $\theta_x$.
For the beta distribution, the hyperparameters $\alpha_x$ and
$\alpha_{\bar{x}}$ must be greater than zero so that the distribution
can be normalized.  Examples of beta distributions are shown in
Figure~\ref{fig:beta}.

The beta prior is convenient for several reasons.  By
Equation~\ref{eq:bt2}, the posterior distribution will also be a beta
distribution:
\begin{equation} \label{eq:beta-p}
p(\theta_x|D) =  {\rm Beta}(\theta_x|\alpha_x+N_x,\alpha_{\bar{x}}+N_{\bar{x}}).
\end{equation}
We say that the set of beta distributions is a {\em conjugate} family
of distributions for binomial sampling. In general, when both the
prior and posterior live in the same distribution family for a
particular type of sampling, we say that distribution family is
conjugate for that type of sampling.  Also, the expectation of
$\theta_x$ with respect to this distribution has a simple form:
\begin{displaymath} \label{eq:beta-mean}
\int \theta_x \ {\rm Beta}(\theta_x|\alpha_x,\alpha_{\bar{x}}) \ d\theta_x
   = \frac{\alpha_x}{\alpha_X}.
\end{displaymath}
So, given a beta prior, we have a simple expression for the
probability of heads in the $N+1$th toss:
\begin{equation} \label{eq:pxb}
p(x_{N+1}|D) = \frac{\alpha_x+N_x}{\alpha_X+N}.
\end{equation}
Note that, for very large $N$ with $\theta_x > 0$, the posterior
distribution over $\theta_x$ is sharply peaked around $N_x/N$, the
fraction of heads observed.  Finally, the marginal likelihood becomes
\begin{displaymath} \label{eq:ml-beta}
p(D) = \frac{\Gamma(\alpha_X)}{\Gamma(\alpha_{X}+N)} \ 
       \frac{\Gamma(\alpha_x)}{\Gamma(\alpha_{x}+N_x)} \ 
       \frac{\Gamma(\alpha_{\bar{x}})}{\Gamma(\alpha_{\bar{x}}+N_{\bar{x}})}.
\end{displaymath}

Assuming $p(\theta_x)$ is a beta distribution, it can be assessed in a
number of ways.  For example, we can assess our probability for heads
in the first toss of the thumbtack ({\em e.g.}, using a probability wheel).
Next, we can imagine having seen the outcomes of $k$ flips and
reassess our probability for heads in the next toss.  From
Equation~\ref{eq:pxb}, we have (for $k=1$):
\begin{displaymath}
p(x_1) = \frac{\alpha_x}{\alpha_x+\alpha_{\bar{x}}},
\ \ \ \ \ 
p(x_2|x_1) =
  \frac{\alpha_x+1}{\alpha_x+\alpha_{\bar{x}}+1}.
\end{displaymath}
Given these probabilities, we can solve for $\alpha_x$ and
$\alpha_{\bar{x}}$.  This assessment technique is known as the method
of {\em imagined future data}.

Another assessment method is based on Equation~\ref{eq:beta-p}.  This
equation says that, if we start with a Beta$(0,0)$ prior and observe
$\alpha_x$ heads and $\alpha_{\bar{x}}$ tails, then our posterior
({\em i.e.}, new prior) will be a Beta$(\alpha_x,\alpha_{\bar{x}})$
distribution. (Technically, the hyperparameters of this prior should
be small positive numbers so that $p(\theta)$ can be normalized.)  If
we suppose that a Beta$(0,0)$ prior encodes a state of minimum
information (a supposition that remains debated), we can assess
$\alpha_x$ and $\alpha_{\bar{x}}$ by determining the (possibly
fractional) number of observations of heads and tails that is
equivalent to our actual knowledge about flipping thumbtacks.
Alternatively, we can assess $p(x_1)$, the probability of heads on the
first toss, and $\alpha_X$, the latter of which can be regarded as an
{\em effective sample size} for our current knowledge.  This technique
is known as the method of {\em effective samples}.  Other techniques
for assessing beta distributions are discussed in \cite{Winkler67} and
\cite{CD83}.

Although the beta prior is convenient, it is not accurate for some
problems.  For example, suppose we think that the thumbtack may have
been purchased at a magic shop.  In this case, a more appropriate
prior may be a mixture of beta distributions---for example,
\[
p(\theta_x) = 
  0.4 \ {\rm Beta}(20,1) +
  0.4 \ {\rm Beta}(1,20) +
  0.2 \ {\rm Beta}(2,2),
\]
where 0.4 is our probability that the thumbtack is heavily weighted
toward heads (tails).  In effect, we have introduced an additional
unobserved or ``hidden'' variable $H$, whose values correspond
to the three possibilities: (1) thumbtack is biased toward heads, (2)
thumbtack is biased toward tails, and (3) thumbtack is normal; and we
have asserted that $\theta$ conditioned on each value of $H$ is a beta
distribution.  In general, there are simple methods ({\em e.g.}, the method
of imagined future data) for determining whether or not a beta prior
is an accurate reflection of one's beliefs.  In those cases where the
beta prior is inaccurate, an accurate prior can often be assessed by
introducing additional hidden variables, as in this example. 

Finally, I note that this approach generalizes to the case where $X$
has $k>2$ values with parameters $\theta_1,\ldots,\theta_k$. In this
case, the conjugate distribution for these parameters is the Dirichlet
distribution, the generalization of the Beta distribution, given by
\begin{displaymath}
p(\theta_1,\ldots,\theta_k) = c \ \prod_{i=1}^k \theta_i^{\alpha_i-1},
\end{displaymath}
where hyperparameters $\alpha_1,\ldots,\alpha_k > 0$, and $c$ is the
normalization constant.

\subsection{Learning acausal DAG models for two or more finite variables}
\label{sec:ladm}

Next, let’s consider the problem of learning DAG models having only
two binary variables, which I will denote $X$ and $Y$. In this case,
there are only two possible model structures to consider---the one
where $X$ and $Y$ are independent, and one where there is no claim of
independence.  Note that there are two graph structures that express
no claim of independence: $X \rightarrow Y$ and $X \leftarrow Y$.
(When depicting DAG models inline, I will draw them without ovals
around node names.)  As acausal networks, they are equivalent, so we
only need consider one of them. To be concrete, let’s use the graph
structure $X \rightarrow Y$.

I denote the model structures representing independence and no claim
of independence $s_{XY}$ and $s_{X \rightarrow  Y}$, respectively.
Using a Bayesian approach, we can reason about the posterior
probability of either of the graph structures (generically denoted
$s$) given their prior
probabilities and marginal likelihoods for data
$D=X_1,Y_1,\ldots,X_N,Y_N$:
\begin{equation} \label{eq:structure-posteriors}
  p(s|D) = \frac{p(s) \ p(D|s)}{p(s_{XY}) \ p(D|s_{XY}) +
    p(s_{X \rightarrow Y}) \ p(D|s_{X \rightarrow Y})}.
\end{equation}
For a historical perspective on this sort of reasoning, see
\cite{KR95}.

Interestingly, there is a relationship between the marginal likelihood
and cross validation. Using the product rule of probability, the
logarithm of the marginal likelihood given a model $s$ can be written
as follows:
\begin{displaymath} \label{eq:seq-val-marg-like}
\log p(D|s) = \log p(X_1)+\log p(X_2|X_1)+\ldots +\log p(X_N|X_1,\ldots,X_{N-1}).
\end{displaymath}
Each term in the sum is a proper score ({\em i.e.}, utility) for the
prediction of one data sample.  In words, first we see how well we predict
the first observation.  Then we train our model with the first
observation and see how well we predict the second observation.  We
continue in this fashion until we predict and score each observation
in sequence. By the rules of probability, the total score will be the
same, regardless of how the data samples are ordered.  This perspective
was first noticed by Phil Dawid \cite{Dawid84}. In this form, the
marginal likelihood resembles cross validation, except that there is
no crossing---that is, there is never a case where $X_i$ is used in
the training data to predict $X_j$ and vice versa.  As crossing can
lead to over fitting \cite{Dawid84}, the marginal likelihood appears
to be an ideal validation approach. In practice, however, I have seen
many instances where cross validation yields a result that seems to be
more reasonable.

Now let’s examine the Bayesian approach to learning models in detail.
First, let’s consider the simpler graph structure $s_{XY}$.  As both
$X$ and $Y$ are binary, let’s use $\theta_x$ and $\theta_y$ to denote
the frequency of $X$ and $Y$ coming up heads, respectively.  When
determining the marginal likelihood for the data given this graph
structure, the assumption that these two parameters are independent
greatly simplifies the computation. In particular,

\begin{eqnarray*}
p(D|s_{xy}) & = & \int p(\theta_x, \theta_y|s_{xy}) \ p(D|\theta_x, \theta_y, s_{xy}) \ d\theta_x d\theta_y \\
& = & \int p(\theta_x|s_{xy}) \ p(\theta_y|s_{xy}) \ p(D|\theta_x, \theta_y, s_{xy}) \ d\theta_x d\theta_y \\
& = & \int p(\theta_x|s_{xy}) \ p(X_1,\ldots,X_n|\theta_x, s_{xy}) \ d\theta_x \ 
\int p(\theta_y|x_{xy}) \ p(X_1,\ldots,X_n|\theta_y, s_{xy}) \ d\theta_y
\end{eqnarray*}

In effect, with this assumption of {\em parameter independence}, the
determination of the marginal likelihood decomposes into the
determination of the marginal likelihood for flips of two separate
thumbtacks. The marginal likelihood becomes
\begin{eqnarray*} \label{eq:ml-two-var-independent}
p(D|s_{xy}) & = & \frac{\Gamma(\alpha_X)}{\Gamma(\alpha_X+N)} \ 
       \frac{\Gamma(\alpha_x)}{\Gamma(\alpha_x+N_X)} \ 
       \frac{\Gamma(\alpha_{\bar{x}})}{\Gamma(\alpha_{\bar{x}}+N_{\bar{x}})} \times\\ 
   & & \frac{\Gamma(\alpha_Y)}{\Gamma(\alpha_Y+N)} \ 
       \frac{\Gamma(\alpha_y)}{\Gamma(\alpha_y+N_y)} \ 
       \frac{\Gamma(\alpha_{\bar{y}})}{\Gamma(\alpha_{\bar{y}}+N_{\bar{y}})},
\end{eqnarray*}
where the hyperparameters and sufficient statistics for $Y$ have
definitions analogous to those for $X$ described in the previous
section.

\begin{figure}
\begin{center}
\leavevmode
\includegraphics[width=2.0in]{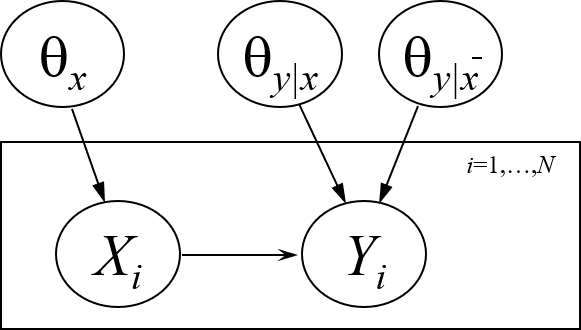}
\end{center}
\caption{Assertions of parameter independence for learning a two-variable DAG model.}
\label{fig:parami}
\end{figure}

Now consider $s_{X \rightarrow Y}$, the graph structure that does not
assume independence. We now have parameter $\theta_x$ as before, and
two parameters: $\theta_{y|x}$, which denotes the frequency of $Y$
when $X$ comes up heads, and $\theta_{y|\bar{x}}$, which denotes the
frequency of $Y$ when $X$ comes up tails.  If we make two assumptions,
the marginal likelihood decomposes into three terms, and we can think
of there being flips for three separate thumbtacks: one for $X$ and
two for $Y$. When $X$ comes up heads on a toss, we flip one of the
thumbtacks for $Y$. When $X$ comes up tails on a toss, we flip the
other thumbtack for $Y$.  The first assumption is that the three
parameters are mutually independent.  The second assumption is that
for every observation of $Y$, we see an observation for $X$. If this
assumption did not hold, the parameters would not remain mutually
independent after observing the data. The independence assumptions for
this case are depicted in Figure~\ref{fig:parami}.

Tracing through the argument for the single-thumbtack case for these
three thumbtacks, we obtain the marginal likelihood for the graph
structure $s_{x \rightarrow y}$:
\begin{eqnarray*} \label{eq:ml-two-var-dependent}
p(D|s_{x \rightarrow y}) & = & \frac{\Gamma(\alpha_X)}{\Gamma(\alpha_X+N)} \ 
       \frac{\Gamma(\alpha_x)}{\Gamma(\alpha_x+N_x)} \ 
       \frac{\Gamma(\alpha_{\bar{x}})}{\Gamma(\alpha_{\bar{x}}+N_{\bar{x}})} \times \\
   &&  \frac{\Gamma(\alpha_{x\cdot})}{\Gamma(\alpha_{x\cdot}+N_x)} \ 
       \frac{\Gamma(\alpha_{xy})}{\Gamma(\alpha_{xy}+N_{xy})}
       \frac{\Gamma(\alpha_{x\bar{y}})}{\Gamma(\alpha_{x\bar{y}}+N_{x\bar{y}})}\times\\
   &&  \frac{\Gamma(\alpha_{\bar{x}\cdot})}{\Gamma(\alpha_{\bar{x}\cdot}+N_{\bar{x}})} \ 
       \frac{\Gamma(\alpha_{\bar{x}y})}{\Gamma(\alpha_{\bar{x}y}+N_{\bar{x}y)}}
       \frac{\Gamma(\alpha_{\bar{x}\bar{y}})}{\Gamma(\alpha_{\bar{x}\bar{y}}+N_{\bar{x}\bar{y}})},
\end{eqnarray*}
where $\alpha_{x\cdot}=\alpha_{xy}+\alpha_{x\bar{y}}$ are the
hyperparameters for the Beta distribution for $Y$ when $X$ is heads,
and $N_{xy}$ and $N_{x\bar{y}}$ are the corresponding sufficient
statistics, and similarly for the terms for $Y$ when $X$ is tails.

Technically, these arguments complete the story.  We start with priors
on the two graph structures and their parameters, observe data, and
compute posteriors on the graph structures using
Equation~\ref{eq:structure-posteriors}. The catch, however, is that
for a domain with $n$ variables, the number of possible graph
structures---assuming none are eliminated by prior knowledge---is
super exponential in $n$. So, it would be extremely desirable to
identify additional techniques, which are reasonable
in many situations, that lead to a reduction in the number of priors
that need to be assessed.

One technique follows from the semantics of the acausal DAG
model---namely, that such models encode only assumptions of
conditional independence. Given these semantics, if two models encode
the same independencies, their parameters should be equivalent. For
example, the parameters of the two-variable model $s_{X \rightarrow
  Y}$ and those of the model $s_{X \leftarrow Y}$, which are
frequencies, show be related according to the rules of probability as
follows:
\begin{displaymath}
\theta_{X} \ \theta_{Y|X} = \theta_{Y} \ \theta_{X|Y}.
\end{displaymath}
(Transforming a distribution for one set of parameters to an
equivalent set is sometimes call a {\em change of variable}.)
Unfortunately, however, unless the hyperparameters for the parameters
of $s_{X \rightarrow Y}$ are specified carefully, the computed
parameters for $s_{X \leftarrow Y}$ will be neither independent nor
distributed according to the Beta distribution, which, as we will see
shortly, leads to a complication. One day in 1993, I was sitting in my
office in Building 9, one of the original Microsoft buildings, and
started to fiddle with the hyperparameters to try to create
independence for both graphs. I discovered that, if I (1) defined
parameters $\theta_{xy}$, $\theta_{x\bar{y}}$, $\theta_{\bar{x}y}$,
and $\theta_{\bar{x}\bar{y}}$ to be the parameters of a single,
four-value variable $XY$, (2) assumed these parameters were
distributed according to the Dirichlet distribution (with arbitrary
hyperparameters), and (3) computed the parameters for the two graph
structures using
\begin{displaymath}
\theta_{XY} = \theta_{X} \ \theta_{Y|X} = \theta_{Y} \ \theta_{X|Y},
\end{displaymath}
then the parameters for both graph structures were mutually
independent and distributed according to the Beta distribution.  At
this point, I tried to find generalizations of the Dirichlet
distribution that satisfied this property.  I couldn’t and wondered
whether the only distribution on $XY$ that yielded parameter
independence for both graph structures was the Dirichlet.  After about
a year of hard work, Dan Geiger and I showed that my guess was
correct. Our theorem states that, if $X$ are finite distributions, and
if the parameters for the two graph structures
$s_{X \rightarrow   Y}$ and $s_{X \leftarrow Y}$ are mutually independent, then the
parameters for $XY$ have a Dirichlet distribution, and each of the
parameters for the two structures are also Dirichlet
\cite{GH97dirichlet}.
\href{https://doi.org/10.1214/aos/1069362752}{[10.1214/aos/1069362752]}.
The theorem is rather remarkable, as qualitative
assertions about independence provide a characterization of the
Dirichlet distribution. To prove the theorem, we used the tools
of functional equations discussed previously.

Based on this theorem, in order to have the mutually independent
parameters for the equivalent graph structures $s_{X \rightarrow Y}$
and $s_{X \leftarrow Y}$, there must exist a probability distribution
for the variable $XY$ and an effective sample size $\alpha > 0$ such
that the hyperparameter $\alpha_{XY}=\alpha \ p(XY)$ for each of the
four values of $XY$.  Under these conditions, our prior knowledge
about the parameters for both graph structures is, in effect, the
situation where we started from complete ignorance and saw complete
data on both variables. Furthermore, this theorem and its consequences
for two variables generalize to domains with more than two
variables. Namely, in a domain with $n$ variables, there are $n!$
graph structures (one for each ordering of the variables) that have no
missing arcs and thus equivalently avoid the assumption of conditional
independence among any of the variables. From the theorem, it follows
that all Dirichlet hyperparameters for the parameters of all complete
graph structures can be derived from a single joint distribution over
the $n$ variables and a single effective sample size $\alpha >0$. And,
to assess this joint distribution, we can assess a single Bayesian
network for it. Thus, all parameter priors are defined from a single
{\em prior Bayesian network} for the domain and the effective sample
size $\alpha$.

\begin{figure}
\begin{center}
\leavevmode
\includegraphics[width=4.0in]{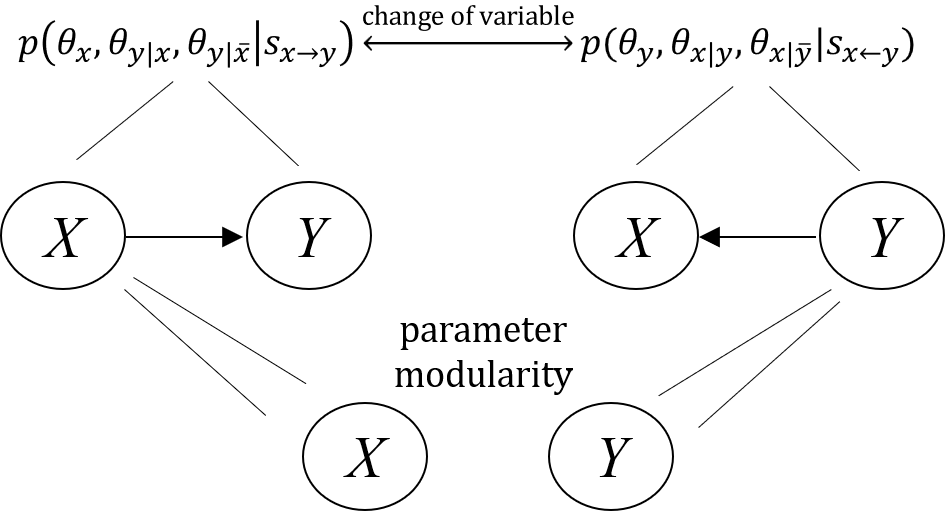}
\end{center}
\caption{The use of parameter modularity to derive the distribution for $s_{xy}$ from the distributions for complete graph structures.}
\label{fig:paramm}
\end{figure}

A second technique that reduces the number of assessments needed has
to do with copying parameter priors from one graph structure to
another. The assumption, called {\em parameter modularity} says that,
if variable $X$ in two different graph structures have the same
parents, then the prior distribution over the parameters for $X$ are
equal.  Note that missing data can destroy parameter modularity as
well as parameter independence \cite{H95lcn}
\href{https://arxiv.org/abs/1302.4958}{[arXiv:1302.4958]}.
Thus, to determine the prior for any parameter $X$ in a graph $s$ with
parents $\Pa(X)$, we first find a complete graph that has an ordering
where $X$ also has parents $\Pa(X)$ and then copy the prior for the
parameters of $X$ from this complete graph structure to the prior for
the parameters of $X$ in $s$. This procedure is illustrated for $n=2$
in Figure~\ref{fig:paramm}. The procedure only makes sense when the
parameters for both complete graph structures satisfy parameter
independence. For the case where $n=2$ and for the case with
arbitrary $n$, the assumption of parameter independence for all
complete graphs is essential for tractable assessment.

Finally, we need to assign a prior, $p(s)$, to every possible DAG
model structure $s$.  A uniform distribution is one possibility, and
is often used. Another approach is to assess a prior Bayesian network
as we have just discussed, and specify a distance metric between an
arbitrary $s$ and the structure of this prior network to define the
prior for $s$ \cite{HGC95ml}.

Putting it altogether, the above description realizes the vision that
Ross and I had years before: We can start with a prior Bayesian
network and improve it with data. Dan and I extended this
approach to include continuous variables with parameters described by a
multivariate-Normal distribution.  The conjugate distribution for
these parameters is the Normal-Wishart and, like the finite-variable
case, Dan and I developed a characterization of this distribution from
an analogous assumption of parameter independence \cite{GH02nw}
\href{https://arxiv.org/pdf/2105.03248.pdf}{[arXiv:2105.03248]}.  In
addition, we showed that it is possible to derive parameter
priors for all possible graph structures from a single Bayesian
network and the assessment of two sample sizes, one for the Normal
distribution and one for the Wishart distribution.

\subsection{Learning fully causal models from observational data only}

Now let's consider learning fully causal models. To be concrete, I
will describe learning fully causal models in simple form (where a
model structure is depicted as a DAG over the domain variables). That
said, the model in simple form can always be converted to an influence
diagram as described in Section~\ref{sec:fullycausalmodels}.

In \cite{SGS93} (second edition)
\href{https://doi.org/10.1007/978-1-4612-2748-9}{[https://10.1007/978-1-4612-2748-9]},
Chris Meek, Greg Cooper, and I show how fully causal models can be
learned from observational data only using essentially the same
approach as that used to learn acausl models.  The key takeaway is
that learning takes place via the probabilistic semantics of the
causal graph, not the causal semantics.  In particular, the learning
derives from the causal Markov assumption
(Section~\ref{sec:fullycausalmodels}), which states that arcs missing
in the graph imply specific conditional independencies. Technically,
learning also derives from the converse assumption that arcs present
in the graph imply specific conditional dependencies.  That said, this
assumption, sometimes called the {\em faithfulness assumption}, is
built into the parameter priors typically used for learning.

When learning causal and acausal DAG models, are the assumptions of
parameter independence, parameter modularity, and parameter
equivalence among graph structures with the same independence
assertions reasonable? When learning causal DAG models, the
assumptions of parameter independence and parameter modularity are
often reasonable, due to the modularity often seen among causal
relationships.  Nonetheless, the assumption of parameter equivalence
among graph structures that encode equivalent conditional
independencies is not mandated. For example, the causal graphs $X
\rightarrow Y$ and $X \leftarrow Y$ are not equivalent.  In contrast,
when learning acausal DAG models, the assumption of parameter
equivalence follows from the semantics of acausal graph structures,
whereas the assumptions of parameter independence and parameter
modularity are less likely to apply than in the case of learning
causal models.  In general, applications of the assumptions need to be
considered carefully.

Another key issue to consider when learning causal DAG models is the
possible presence of confounders. We look at this issue in detail in
Section~\ref{sec:confounders}.

\subsection{Learning influence diagrams from a mix of observations and
  decisions}
\label{sec:lid}

Next, let's consider the case of learning influence diagrams from data
containing a mix of observations and interventions. The approach is
similar to what we have just described for acausal DAG models with two
exceptions.  One, because utilities represent preferences, they are
not learned---they are specified by the decision maker.  Two, decision
variables, by their nature, have no parameters to update and are root
nodes in any possible influence-diagram structure.  So, the task of
learning influence diagrams from data boils down to learning causal
structures for the domain's uncertainty variables conditioned on its
decision variables.  If we assume that all decisions are made in each
data sample and all uncertainty variables are observed in each data
sample, make the assumptions of parameter independence and parameter
modularity, and use conjugate priors, then we can learn influence
diagrams using the the same techniques we have discussed for learning
acausal DAG models.

The task of learning a fully causal model from a mix of observations
and interventions is a simple special case.  Here, only those
observations of uncertainty variables whose corresponding force/set/do
decisions have the value ``no nothing'' participate in the marginal
likelihood.  When a variable is forced to be a particular value, its
observation provides no information about the relationship between the
variable and its possible direct causes.

I seem to have been the first to describe learning in these two
situations.  For details, see \cite{H95lcn}
\href{https://arxiv.org/abs/1302.4958}{[arXiv:1302.4958]}.

As mentioned, one issue with fully causal models is that the
force/set/do decision is typically not realized in practice. A notable
exception is the learning of causal models relating the expression of
various genes from a combination of observations and gene
manipulations. In this case, the gene manipulations can be made in
such a way as to directly affect only one gene and thus reflect a
force/set/do decision. In collaboration with Amgen, I used this
approach to learn causal models relating the expression of thousands
of genes under various conditions related to drug design.  Amgen has
told me that the resulting models have been very useful to them
\cite{HeckermanAmgenPC}.

\subsection{Confounders}
\label{sec:confounders}

In the 1990s, a strong negative correlation between consuming
$\beta$-carotene and lung cancer was noted. This correlation prompted
cereal companies to start adding $\beta$-carotene to their cereals and
promoting that addition right on the box. A randomized trial published
in 1996, however, produced strong evidence that $\beta$-carotene
actually caused cancer \cite{bCarotene96}. Similarly, There was a
well-known negative correlation between levels of HDL in the blood and
heart attacks. So, over the last two decades, several companies
developed drugs to raise HDL levels.  Many of the drugs successfully
increased HDL, but failed to suppress heart attacks ({\em e.g.},
\cite{Roche12}). The drug companies lost billions of dollars.

\begin{figure}
\begin{center}
\leavevmode
\includegraphics[width=5.0in]{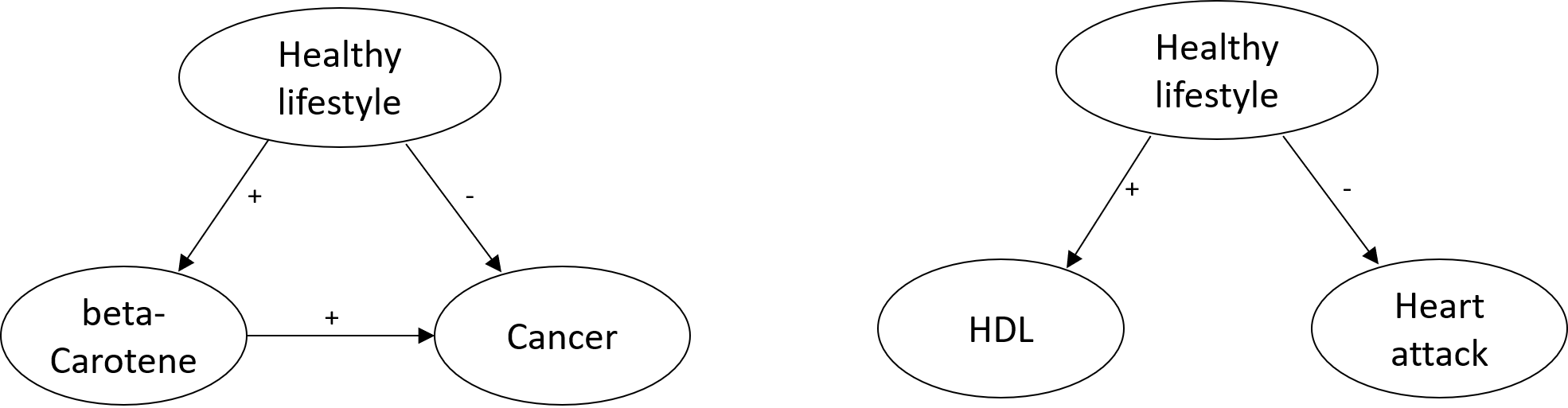}
\end{center}
\caption{Confounders create correlations that differ from direct cause.}
\label{fig:confounders}
\end{figure}

What was going on? The answer lies in the fully causal models shown in
Figure~\ref{fig:confounders}. In the case of $\beta$-carotene, an
unobserved or hidden health lifestyle was causing $\beta$-carotene in
the body to increase and also causing a reduction in cancer. In the
HDL case, a hidden health lifestyle was causing an increase in HDL and
a decrease in heart attacks. In each case, there was a hidden
variable---a {\em confounder}---that was a common cause of two
variables, thus inducing a correlation between them. When there were
interventions on the supposed cause, the effect was not as
anticipated.  As the saying goes: ``correlation is not causation.''

\begin{figure}
\begin{center}
\leavevmode
\includegraphics[width=3.5in]{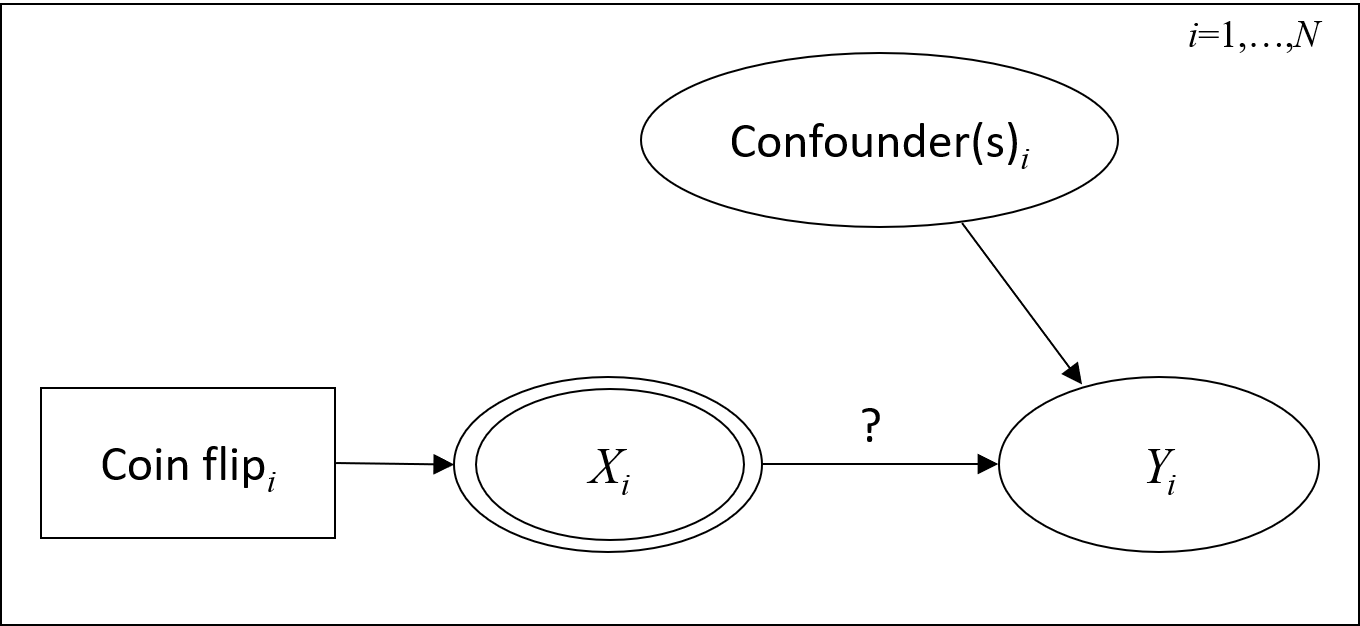}
\end{center}
\caption{A randomized trial for quantifying the causal effect of $X$ on $Y$.}
\label{fig:randomized-trial}
\end{figure}

Of course, this issue has been known for almost a century, and the
ideal solution for teasing out possible causation from correlation is
to perform a {\em randomized trial}.  An influence diagram (in plate
notation) for a generic randomized trial that measures the causal
effect of $X$ on $Y$ is shown in
Figure~\ref{fig:randomized-trial}. Here, a coin is flipped for each of
$N$ subjects. Depending on the outcome of the flip, an intervention is
performed that forces $X$ to be one of two values, and $Y$ is then
observed. The randomized trial accurately assesses the causal effect
of $X$ on $Y$, because the coin flip breaks the causal influence of
the confounders on $X$.  In effect, the coin flip is a force/set/do decision
without the alternative of ``do nothing.''  Unfortunately, the
randomized trial is often an unattainable ideal. Such trials can be
extremely expensive.  Sometimes, ethical considerations preclude a
randomized trial ({\em e.g.}, when attempting to determine the effects
of smoking on cancer).  Also, study subjects often fail to comply with
the intervention.

All attempts to solve this problem involve making
assumptions. Basically, the goal is for the analyst to construct an
argument that people will accept. The more reasonable the assumptions,
the greater the number of individuals who will accept the argument.

\begin{figure}
\begin{center}
\leavevmode
\includegraphics[width=3.5in]{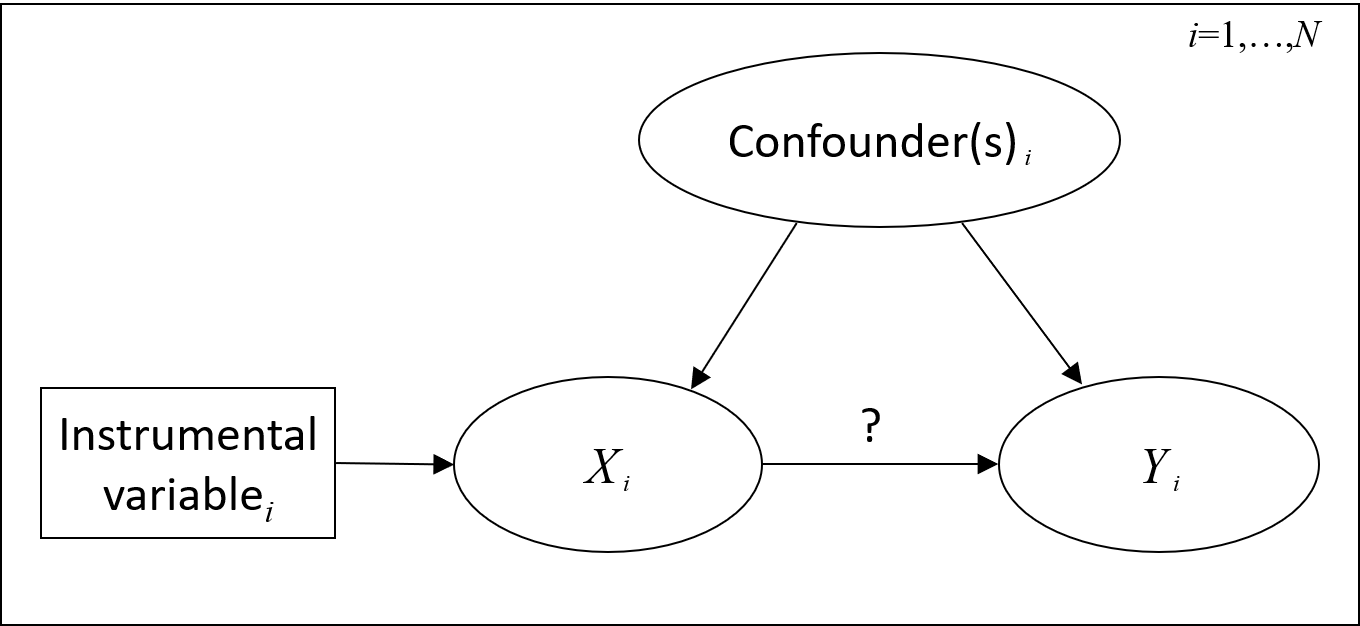}
\end{center}
\caption{An influence diagram illustrating the conditional independencies that must be satisfied by an instrumental variable.}
\label{fig:instrumental-var}
\end{figure}

One solution is to replace the coin flip with a natural intervention.
As shown in Figure~\ref{fig:instrumental-var}, this variable is called
an {\em instrumental variable.} In order for the instrumental variable
to accurately uncover the causal effect of $X$ on $Y$, it must obey
the independence assumptions shown in the figure.  In particular, the
instrumental variable must be independent of the confounder(s), and
the instrumental variable must be independent of $Y$, given $X$ and
the confounder(s). In addition, the instrumental variable must have a
direct effect on $X$.  One interesting point here is that, if there is
no causal effect, and if the instrumental variable and $Y$ are
independent, then the two independence assumptions are satisfied. So,
in this case, the assumptions don’t need to be left to debate---they
can be demonstrated with data.

Voight {\em et al.} \cite{VAK12hdl} used genetic markers as
instrumental variables to investigate the effects of HDL increasing
drugs on the incidence of heart attack.  (In general, when the
instrumental variable is a genetic marker, this process is called {\em
  Mendelian randomization.})  He found genetic markers that had a
direct effect on HDL but were independent of incidence, and thus
concluded that raising HDL had no effect on incidence.  If only this
work had been done {\em before} the drug companies spent billions on
drugs development.

Another common approach for dealing with confounders is to use
observational data only, avoiding interventions, and assume that a
complete set of relevant confounders are observed.  (\cite{SGS93}
second edition, p.209, and \cite{Pearl00} second edition, p.79,
provide an algorithm for identifying such a set.)  In this approach,
the causal effect of $X$ and $Y$ is estimated by weighting each
observation with one over the probability of $X$ given the
confounder(s).  This approach, known as propensity score weighting,
was first introduced by Robins and others in 2000 \cite{RHB00psw}.
This approach has been typically applied to problems where $X$ is
finite and has a relatively small number of values. Recently, however,
Taha Bahadori, Eric Tchetgen, and I used deep learning to create an
approach that can provide an accurate estimate when $X$ is continuous
\cite{BTH22icml}
\href{https://arxiv.org/abs/2107.13068}{[arXiv:2107.13068]}.

Another approach is to use DAG model learning that includes a
deliberate search for confounders.  Let's consider an example taken
from \cite{H98tutorial}
\href{https://arxiv.org/abs/2002.00269}{[arXiv:2002.00269]}.  Sewell
and Shah \cite{SS68} measured the following variables for 10,318
Wisconsin high school seniors: \node{Sex} ($Sex$): male, female;
\node{Socioeconomic Status} ($SES$): low, lower middle, upper middle,
high; \node{Intelligence Quotient} ($IQ$): low, lower middle, upper
middle, high; \node{Parental Encouragement} ($PE$): low, high; and
\node{College Plans} ($CP$): yes, no.  The data are described by the
sufficient statistics in Table~\ref{tab:cp}.  Each entry denotes the
number of cases in which the five variables take on some particular
configuration.  The first entry corresponds to the configuration
SEX$=$male, SES$=$low, IQ$=$low, PE$=$low, and $CP$=yes.  The
remaining entries correspond to configurations obtained by cycling
through the values of each variable such that the last variable (CP)
varies most quickly.  Thus, for example, the upper (lower) half of the
table corresponds to male (female) students.

\begin{table} \label{tab:cp}
\caption{Sufficient statistics for the Sewall and Shah (1968) study.}
\begin{tabular}{|cccccccccccccccc|}
\hline
  4&349& 13& 64&  9&207& 33& 72& 12&126& 38& 54& 10& 67& 49& 43 \\
  2&232& 27& 84&  7&201& 64& 95& 12&115& 93& 92& 17& 79&119& 59 \\
  8&166& 47& 91&  6&120& 74&110& 17& 92&148&100&  6& 42&198& 73 \\
  4& 48& 39& 57&  5& 47&123& 90&  9& 41&224& 65&  8& 17&414& 54 \\
   &   &   &   &   &   &   &   &   &   &   &   &   &   &   &    \\
  5&454&  9& 44&  5&312& 14& 47&  8&216& 20& 35& 13& 96& 28& 24 \\
 11&285& 29& 61& 19&236& 47& 88& 12&164& 62& 85& 15&113& 72& 50 \\
  7&163& 36& 72& 13&193& 75& 90& 12&174& 91&100& 20& 81&142& 77 \\
  6& 50& 36& 58&  5& 70&110& 76& 12& 48&230& 81& 13& 49&360& 98 \\
\hline
\end{tabular}
\newline \noindent Reproduced by permission from the University of Chicago
Press.
\newline \copyright 1968 by The University of Chicago.  All rights
reserved.
\end{table} 

\begin{figure}
\begin{center}
\leavevmode
\includegraphics[width=4.0in]{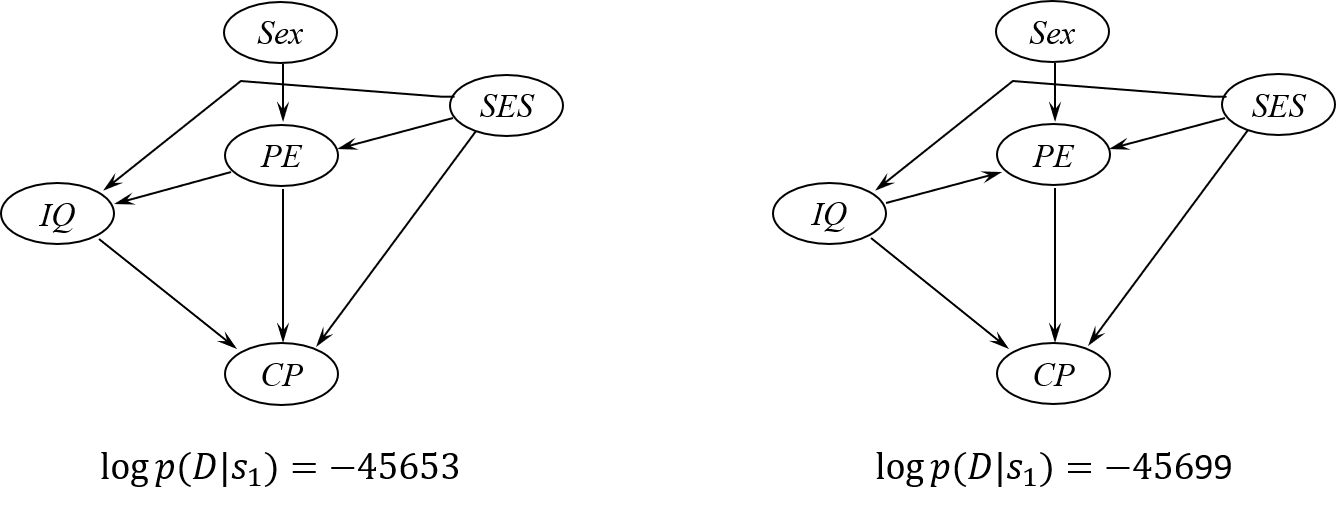}
\end{center}
\caption{The two DAG structures without confounders with the
  highest marginal likelihoods.}
\label{fig:cp1}
\end{figure}

As a first pass, I analyzed the data assuming there are no
confounders.  To generate priors for the model parameters, I used the
method described in Section~\ref{sec:ladm} with an equivalent sample
size of 5 and a prior Bayesian network where the joint distribution is
uniform.  (The results were not sensitive to the choice of parameter
priors.  For example, none of the results reported here changed
qualitatively for equivalent sample sizes ranging from 3 to 40.)  I
considered all possible structures except those where $SEX$ and/or
$SES$ had parents, and/or $CP$ had children.  The two graph structures
with the highest marginal likelihoods are shown in
Figure~\ref{fig:cp1}.

Some of the results are not surprising---for example the causal
influence of socioeconomic status and IQ on college plans.  Other
results are more interesting.  For example, from either graph we
conclude that sex influences college plans only indirectly through
parental influence.  Also, the two graphs differ only by the
orientation of the arc between PE and IQ.  Either causal relationship
is plausible.

\begin{figure}
\begin{center}
\leavevmode
\includegraphics[width=5.0in]{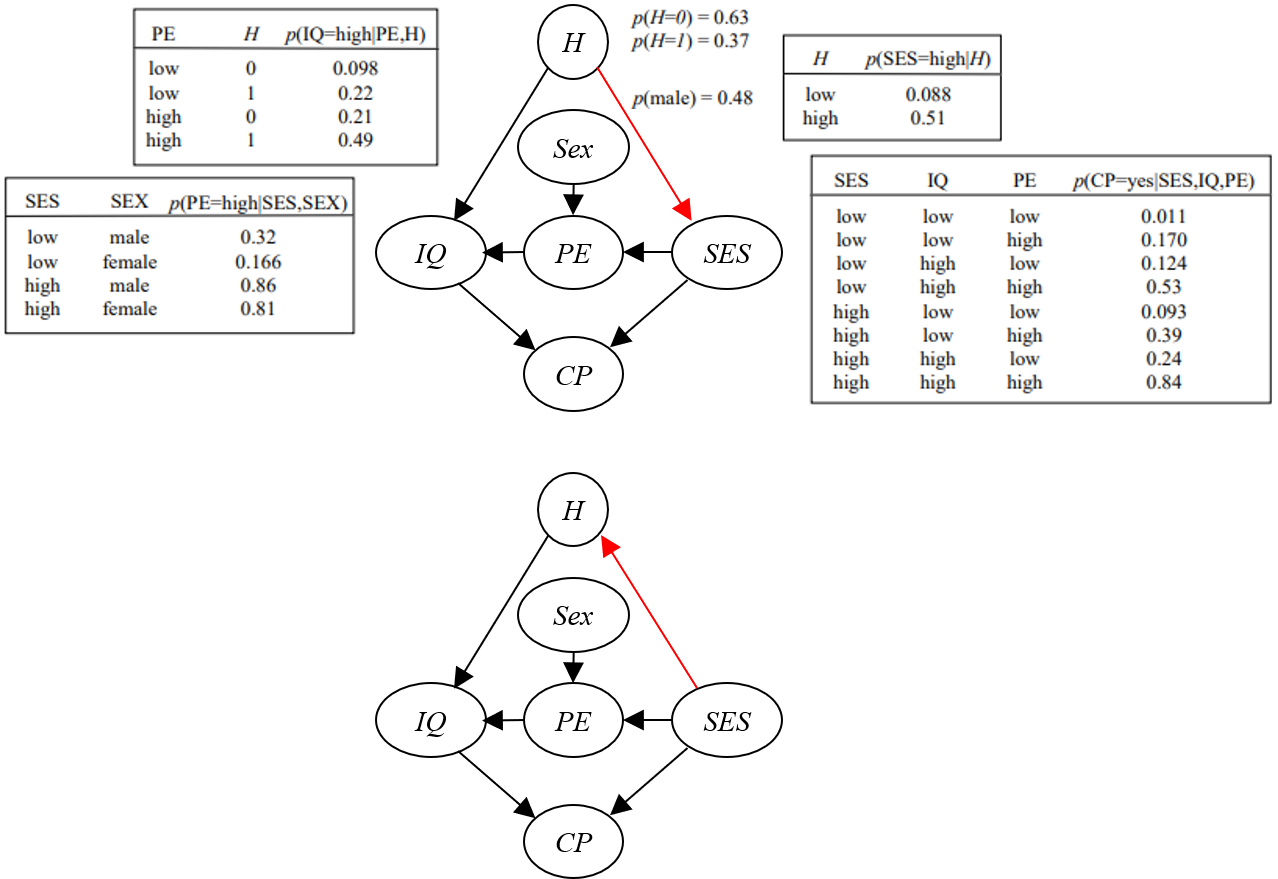}
\end{center}
\caption{The two hidden-variable graph structures with the highest
  log marginal likelihood (both -45611).  The red arcs highlight the
  difference between the two graph structures. Expectations of the
  parameters for the first model are shown.}
\label{fig:cp2}
\end{figure}

The most suspicious result is the suggestion that socioeconomic status
has a direct influence on IQ.  To question this result, I considered
additional graph structures with hidden variables.  As the approach in
Section~\ref{sec:lgm} requires complete observations on all variables,
I asked Radford Neal to apply his method known as annealed importance
sampling, a Monte-Carlo technique \cite{Neal98}, to compute the
marginal likelihoods.  Among the models considered, there were two
with equal marginal likelihoods that were much higher than the
marginal likelihoods of the best structures without hidden variables.
They are shown in Figure~\ref{fig:cp2}.  In the upper model, the
confounder $H$ is a hidden common cause of $IQ$ and $SES$ ({\em
  e.g.}, parent ``quality.''). In the lower model, $SES$ is a cause
of $IQ$ as in the best model with no hidden variables, and $H$ helps
to capture the orderings of the values in the discretized versions of
$SES$ and $IQ$.  This example illustrates that learning DAG models can
leave us uncertain.  Assuming these two models are equally likely {\em
  a priori}, they remain equally likely given the data.  A consolation
is that we can average over the graph structures to make decisions or
generate probabilities over various causal claims.

\begin{figure}
\begin{center}
\leavevmode
\includegraphics[width=1.0in]{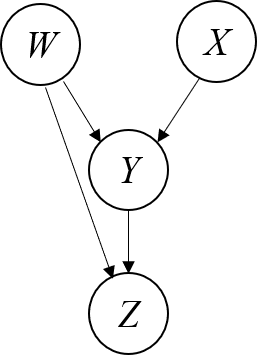}
\end{center}
\caption{The arc from $Y$ to $Z$ is guaranteed to have no confounders.}
\label{fig:y-structure}
\end{figure}

Finally, when learning fully causal DAG models, we can get lucky.  For
example, consider the graph structure for observed variables $W$, $X$,
$Y$, and $Z$, shown in Figure~\ref{fig:y-structure}.  If there were
one or more hidden confounders of $Y$ to $Z$, there would be
additional dependencies among the observed variables. Consequently,
the structure shown in the figure would have a lower posterior
probability than some other structure over the observed variables.  As
a result, if the structure shown in the figure has a high enough
posterior probability, then we can conclude that the arc from $Y$ to
$Z$ is not confounded.  A similar argument applies to a graph
structure as in the figure, but without the arc from $W$ to $Z$.

The bottom line is that, when an ideal randomized trial is not
available, assumptions are needed to argue for a causal
relationship. The good news is that the stigma in academics of talking
about causality has largely gone away, and many researchers are now
working to devise methods with assumptions that are more and more
reasonable in practice.

\subsection{Deep Learning}

At this point in reading through the manuscript, you are almost
certainly wondering: what about deep learning?  Over the last decade,
deep learning has made great advances in machine learning. Almost
everyone in the world has been touched by its accomplishments.  Yet,
with just one exception, I have so far focused only on traditional ML.
So here, I touch on several fundamental points about deep learning.

The first point is that a deep learning model is a graphical model
(often a DAG model), but with several differences in emphasis from a
traditional DAG model.  One difference is that a deep model often
represents a function from inputs to outputs rather than a joint
distribution over all variables. A related difference is that a deep
model is typically arranged in layers of nodes where only the input
and output layers correspond to definitive observations in the real
world.  As one moves from the input layer to the output layer, each
subsequent layer consists of unobserved variables that represent an
incremental transformation from the previous layer. For example, in a
convolutional deep model, the input layer consists of pixels on a
screen, and subsequent layers correspond roughly to edges and lines,
and then shapes, and then objects in the image. Interestingly, the
visual cortices of primates have a similar organization
\cite{Msail21}.

\begin{figure}
\begin{center}
\leavevmode
\includegraphics[width=3.0in]{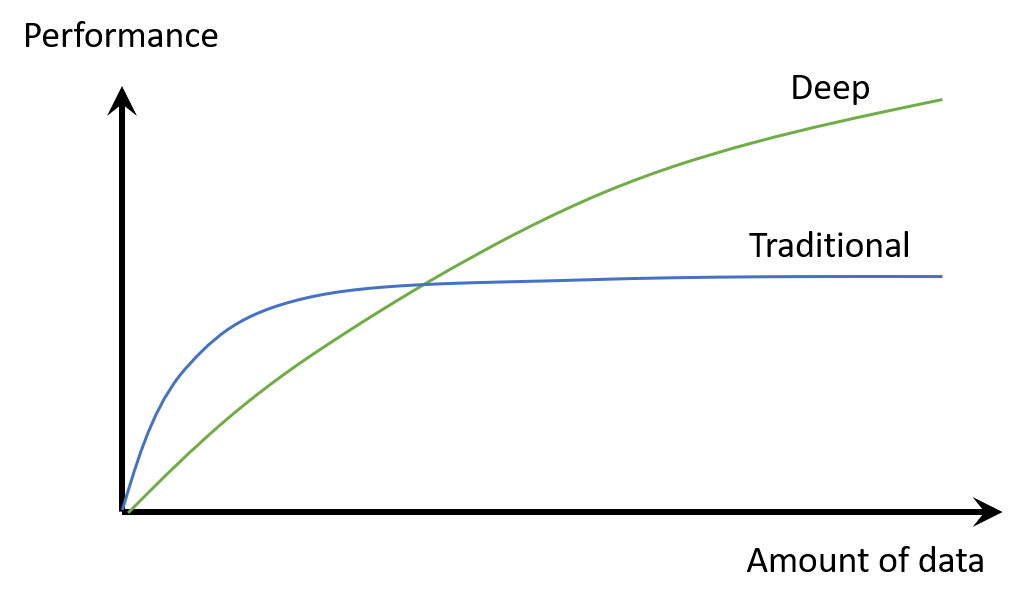}
\end{center}
\caption{Learning curves for traditional ML and deep learning.  The
  axes on these plots are deliberately not labelled, because
  performance as a function of amount of data is highly problem
  specific.}
\label{fig:deep-vs-traditional}
\end{figure}

The abundance of hidden variables in deep models leads to another set
of differences. One is that a deep model can represent very complex
functions. Another is that a deep model requires a large amount of
data for training before its performance exceeds that of a traditional
model.  These two differences are illustrated by the curves shown in
Figure~\ref{fig:deep-vs-traditional}.  These curves, known as {\em
  learning curves}, plot algorithm performance as a function of the
amount of data available to train the algorithm.  Traditional ML is
superior when there are small amounts of data, whereas deep learning
is superior there are large amounts of data.  The reason underlying
the difference is simple: traditional methods encode prior knowledge
either explicitly ({\em i.e.}, they are Bayesian) or implicitly, and
this prior knowledge gives those methods the edge when small amounts
of data are available for training.  The downside of relying on prior
knowledge is that it is not perfect and, as the data available for
training grows, these imperfections constrain the ability of the
traditional methods to excel. The large number of hidden variables in
a deep model also makes the explanation of its predictions more
difficult.

A third point is that, if you accept the von Neumann--Morgenstern
properties, you are bound to the prescriptions of decision theory for
making decisions.  This adherence holds whether you use traditional ML
or deep learning.  Of course, for low stakes decisions, it may be fine
to append some {\em ad hoc} decision making to the output of deep
learning---for example, a system that makes product recommendations
could recommend products with the highest-value outputs. For high
stakes decisions, however, following the MEU principle, which requires
probability distributions over uncertainties, is a must.  This point
is important because a deep model often lacks a probabilistic
component. Fortunately, it is always possible to augment a deep model
to output probabilities for decision making. For example, when a deep
learner outputs a number for a binary target variable, {\em
  calibration procedures} exist to map the output to a probability.
For a discussion of calibration and its relationship to proper
scoring rules, see \cite{GBR07}.

A related final point is that a deep model (actually, any
single-output neural network) can be used to represent the conditional
probability distribution associated with a parent--child relationship
in a traditional DAG model. I mentioned this point in
\cite{H98tutorial}
\href{https://arxiv.org/abs/2002.00269}{[arXiv:2002.00269]}, and such
use is becoming more and more widespread \cite{BH21}
\href{https://iclr.cc/virtual/2021/poster/2732}{[ICLR 2021]}.

\section{Applications: Fun with uncertainty, decision making, and graphical models at Microsoft Research}
\label{sec:app-msr}

This section is all about some of the applications I was lucky enough
to build while at Microsoft Research (MSR). I present them roughly in
the order in which they were conceived and don’t rehash the technical
details---you can read the original publications if you are
interested. Mostly, I tell stories about how they came to pass. As
you’ll see, my early work centered around expert systems and then
evolved into machine learning. Nathan Myhrvold hired me to work on
expert systems, but when I got to MSR, I quickly realized we had too
few experts but lots of data. So, I began to teach myself machine
learning, which---no surprise---eventually led to all sorts of
high-impact applications. I am so grateful to Bill Gates, Nathan
Myhrvold, and Rick Rashid for starting MSR and for letting those of us
who were lucky enough to be there to explore whatever paths we thought
were promising.

The first application I initiated was the Answer Wizard, an expert
system that takes short text queries such as “How do I print
sideways?” and provides an answer. The project came about during my
first few days at Microsoft when I started to use Excel for the first
time and couldn’t figure out how to make a graph. I tried the help
function, but it didn’t know what a graph was. When I found someone at
MSR to help me, they said, ``Oh, you want to make a {\em chart}.'' I
thought to myself, ``Who calls these things `charts’? Did the person
who wrote the user interface like sailing?'' It immediately occurred
to me that I could build a diagnostic Bayesian network, much like
Pathfinder, that related possible tasks the user wanted to perform to
words or phrases they might use to describe those possible tasks. Just
days later, Erich Finklestein and Sam Hobson, who were working on what
would become Office 95 met with me.  (To this day, I don’t know how
they found me.  My guess is that Nathan was funneling people in
product groups to me.)  They too were thinking about how the user
could more easily discover functionality within Office.  The Answer
Wizard was born.  You can find the technical details in \cite{HH98aw}
\href{https://arxiv.org/abs/1301.7382}{[arXiv:1301.7382]}.

\begin{figure}
\begin{center}
\leavevmode
\includegraphics[width=3.0in]{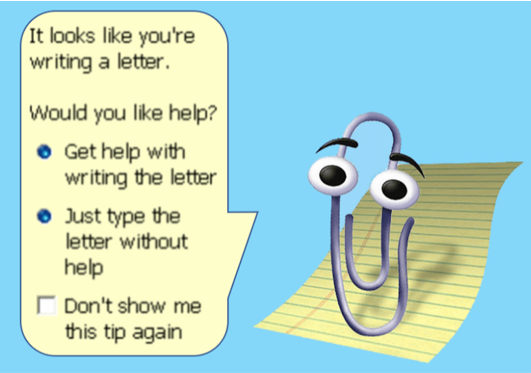}
\end{center}
\caption{Clippy: the user interface for the Answer Wizard in Microsoft
  Office from 1998 to 2003.}
\label{fig:clippy}
\end{figure}

The Answer Wizard still exists today. It’s a type-in box at the top of
Office applications that says, ``Tell me what you want to do,'' which
closely resembles its original form in Office 95. That said, for
several years starting in the late 1990s, it got a lot of attention
when an animated paper clip, named ``Clippy,'' served as its front end
(see Figure~\ref{fig:clippy}).  When the designers of Clippy first
showed it to me, they said they would have about five different
characters from which the user could select, Clippy being one of them.
I asked, ``Could you make one of the characters the invisible man [a
  famous comic-book character]?'' My intent was to suggest that they
leave the user interface alone. They understood my intent and were not
pleased with my suggestion. In some sense, however, I think I got it
right, at least at the time, as the character quickly became perceived
as too silly and too annoying. That said, the character worked its way
into the hearts of many, even becoming a popular Halloween costume. I
hear rumors that Microsoft may bring it back. With hindsight, they
should be able to make it less annoying. And with modern-day machine
learning, they should be able to make it more useful.

\begin{figure}
\begin{center}
\leavevmode
\includegraphics[width=5.0in]{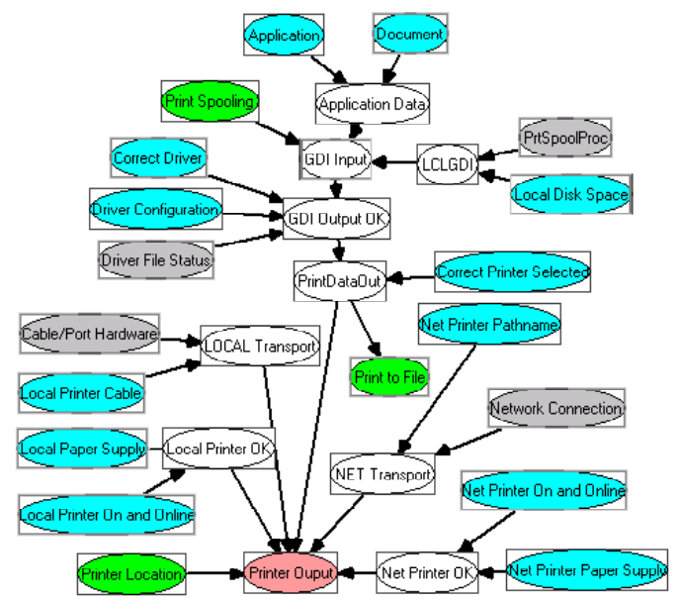}
\end{center}
\caption{A fully causal model for print troubleshooting. Colorings are
not important.}
\label{fig:print-ts}
\end{figure}

The next project, the Windows Troubleshooters, began just weeks after
the Answer Wizard. Phil Fawcett, working on the Product Support team,
sought me out. (Again, I’m guessing this came from Nathan's
direction.)  Phil was looking for ways that product support could be
automated. As an example, he mentioned that many Windows customers
were having problems with printing. To me, the problem seemed similar
to the task of medical diagnosis, so I set out to solve the problem in
a similar way.  Soon, however, I learned an important difference:
unlike medical diagnosis, where clues about the diagnosis come in the
form of pure observations, device troubleshooting requires clues that
come from a mix of interventions and observations.  For example, when
troubleshooting a print problem, one could observe a symptom, then
turn the printer off and then on again, and re-observe to see if and
how that symptom had changed. Having had lots of experience with
influence diagrams and causal reasoning by this time, it was
straightforward to adapt the technology I had built for medical
diagnosis to this new task.  You can find the details at \cite{BH96ts}
\href{https://arxiv.org/abs/1302.3563}{[arXiv:1302.3563]}.  As I was
coding the troubleshooting algorithms, Jack Breese arrived as the
second member of our group at MSR.  We called it DTG---the Decision
Theory Group.  He developed an elegant user interface for the
troubleshooters, and together we refined the fully causal model for
print toubleshooting (see Figure~\ref{fig:print-ts}).  We demoed the
system to Phil Fawcett, and the Windows Troubleshooters were born.
They were popular for at least a decade, but eventually got replaced
when sufficient problem--solution pairs were collected and could be
searched.

The next project was one of my first forays into a machine-learning
application.  It was 1994. I had just formed the Machine Learning and
Applied Statistics group at MSR and was armed with the theoretical
work on learning DAG models from data. I attended a workshop on this
new idea called collaborative filtering. The idea was explained more
as an algorithm than a task. In particular, it was explained as
recommending items ({\em e.g.}, movies, websites) to a user that they might
like, by identifying other users who liked similar items, and then
recommending items that the others liked that the original user did
not know about. Right there during the workshop, I realized that the
{\em task} was a prediction problem: given data on liked items by a
set of users, predict other items that each person would also
like. Furthermore, I thought that, while similarity-based methods
would be one way to address the task, it would be worth trying other
prediction algorithms to see if they worked better.

Over the next few months, Carl Kadie (new to the DTG team) and Jack
Breese, began exploring various algorithms on multiple datasets.  As
an example, let’s consider one of my favorites: Nielsen rating data
for TV shows. For every user and every TV show, Nielsen recorded
whether the user watched the show during February sweeps week in
1995. For every show, we built an (ML) decision tree with that show as
the classification variable, using all other TV shows as prediction
variables. The results made a lot of sense.  If you watched Seinfeld,
you were likely to watch Friends. And if you watched Seinfeld and
didn’t watch Friends, it was worth recommending Friends to you.
Another good example was that, if you liked watching Matlock, you
would likely enjoy 60 Minutes. For more details, see \cite{BHK98cf}
\href{https://arxiv.org/abs/1301.7363}{[arXiv:1301.7363]}.

The approach was a big hit at Microsoft and in the business world in
general. It became so popular that the technology shipped in Microsoft
Commerce Server 2000.  It also caught the eye of Bill Gates, who asked
me to present the work at a large conference with him. The
presentation consisted of showing a visualization of the model applied
to the Nielsen data, which I had set up on my laptop, and I remember
being quite nervous before the presentation. ``I had better not screw
this up,’’ I thought to myself. It turns out that I was so nervous
that I had forgotten to plug the charger into my laptop. Immediately
after finishing the presentation, I looked at my laptop and it went
dark. I had barely avoided a major disaster.

\begin{figure}
\begin{center}
\leavevmode
\includegraphics[width=6.0in]{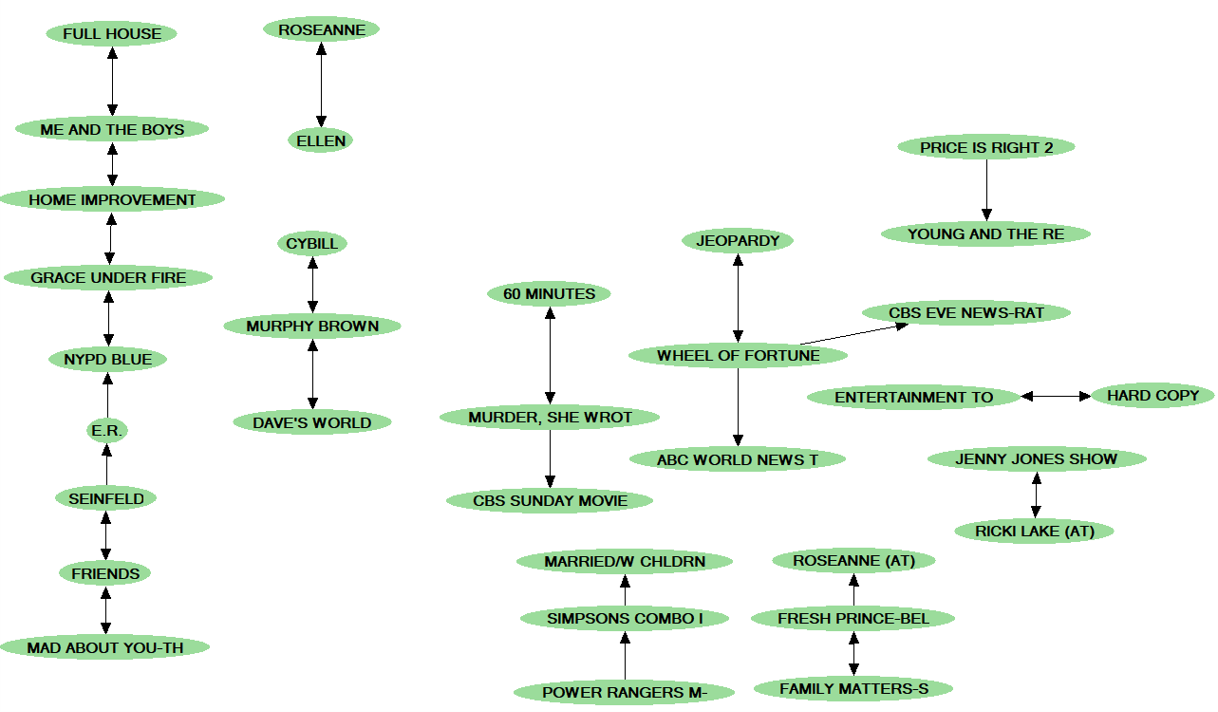}
\end{center}
\caption{The dependency network structure for the 1995 Nielsen ratings data.
  Only stronger dependencies are shown.}
\label{fig:nielsen}
\end{figure}

The work on collaborative filtering led to the creation of a new type
of graphical model, which I unimaginatively called the {\em dependency
  network.}  In this approach, rather than construct a directed
acyclic model, one constructs a typically cyclic graph, where each child
variable has all other variables as possible parents. The missing arcs
reflect conditional independence---in particular,
$p(X_i|X_1,\ldots,X_{i-1},X_{i+1},X_n)=p(X_i|\Pai)$, but now many or
all of the arcs in the dependency network are bi-directional.  A
dependency network structure for the Nielsen data is shown in
Figure~\ref{fig:nielsen}.

When predicting a target variable from observations of all other
variables, we can just look up the result from that target's local
distribution.  If more complex inference is needed, we can use Gibbs
sampling.  In this approach, we begin by assigning random values to
every variable.  Then, we cycle through each variable, sampling a new
value for it, based on the values of its parents’ variables. We keep
track of the distribution on these samples, which in the limit yields
a joint distribution over the variables.

This model and inference approach leads to several interesting
mathematical results.  To understand these results, first note that
the local distributions may be consistent or inconsistent with one
another. Inconsistency is likely when the local distributions are
learned independently. For example, consider the simple domain
consisting of two variables $X$ and $Y$. Is it possible that an
estimator of $p(X|Y)$ discards $Y$ as an input, whereas the estimator
of $p(Y|X)$ retains $X$ as an input.  The result is the structural
inconsistency that $X$ helps to predict $Y$, but $Y$ does not help to
predict $X$.  Numeric inconsistencies are also likely.
Nonetheless, in situations where the data set contains many samples,
strong inconsistencies will be rare because each local distribution
is learned from the same data set, which presumably is generated
from a single underlying joint distribution.  In other words, although
dependency networks will be inconsistent, they will be ``almost''
consistent when learned from data sets with large sample sizes.

So first, let’s consider consistent dependency networks. In this case,
every arc in the graph structure must be bi-directional, so we may as
well replace these bi-directional arcs with undirected edges, yielding
an undirected graph structure. Given this observation, a question
arises as to whether the dependence network is equivalent to another
undirected network known as a {\em Markov random field} (also known as
a {\em Markov network}) as originated by Julian Besag.  Instead of
Gibbs sampling, Julian's approach performs inference using
distributions on subsets of variables corresponding to cliques in the
undirected graph structure.  In 2000, Max Chickering, Chris Meek, and
I showed that, indeed, a consistent dependency network for a domain of
finite variables is equivalent to a Markov random field, provided all
local distributions are positive \cite{HCMRK00dn}
\href{https://jmlr.org/papers/v1/heckerman00a.html}{[jmlr/v1]}.
Here, by ``equivalent’’, I mean that, given an undirected graph
structure, any distribution represented by one approach can also be
represented by the other.  It is interesting to note that Julian
started with directed dependency networks, but instead of taking the
path of Gibbs sampling for inference, took the path involving
distributions on cliques. When I showed this work to Julian, he said,
``I wish I had thought to use Gibbs sampling back then---it would have
been so much simpler.’’

What happens with consistent dependency networks when we relax the
assumption that distributions are positive? In the case of undirected
networks, different definitions, while equivalent when distributions
are positive, are no longer equivalent \cite{Lauritzen96}.  Thomas
Richardson, Chris, and I characterized the distributions
representable by dependency networks in the non-positive case, and
showed that there are distributions representable by dependency
networks that are not representable by various definitions of
undirected network, and vice versa \cite{HMR14dn}
\href{https://doi.org/10.14736/kyb-2014-3-0363}{[10.14736/kyb-2014-3-0363]}.

What happens when dependency networks are inconsistent?  In the same
publication \cite{HCMRK00dn}
\href{https://jmlr.org/papers/v1/heckerman00a.html}{[jmlr/v1]},
Max, Chris, and I
showed that Gibbs sampling will still converge to a unique
distribution, provided the local distributions are positive. This
distribution may not be consistent with the local distributions and
can depend on the order in which the Gibbs sampler visits each
variable, but convergence is nonetheless guaranteed.

\begin{figure}
\begin{center}
\leavevmode
\includegraphics[width=6.0in]{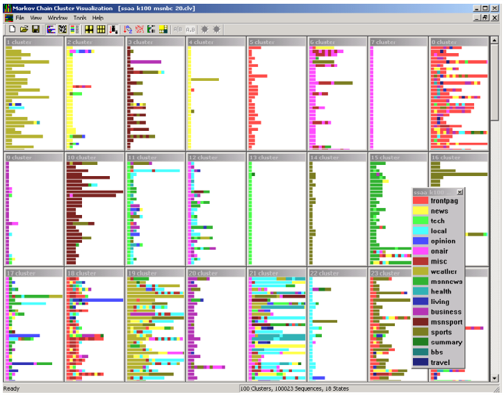}
\end{center}
\caption{Display of msnbc.com data using WebCANVAS. Each
  window corresponds to a cluster. Each row in a window corresponds to
  the path of a single user through the site. Each path is color-coded
  by category. The category legend is at the lower right of the
  screen. Clusters are listed in descending order of size (number of
  session assigned to a cluster) from left to right and then top to
  bottom.}
\label{fig:webcanvas}
\end{figure}

This next application is a bit out of time order, as I’m saving the
best for last.  In early 2000, Microsoft services were in full bloom.
Steven White from msnbc.com contacted me, hoping to gain a better
understanding of how customers were using their new site. Having
recently completed work on unsupervised learning with mixture models
({\em e.g.}, \cite{CH97ml}
\href{https://doi.org/10.1023/A:1007469629108}{[10.1023/A:1007469629108]},
\cite{TMCH99}
\href{https://arxiv.org/abs/1301.7415}{[arXiv:1301.7415]}),
I suggested that we
try clustering customers’ behavior. Steven was particularly interested
in the sequence of pages visited on the site, so I came up with the
idea of learning and then visualizing a mixture of time-series
models. I suggested we start with a very simple version---a mixture of
first-order Markov models.  This simple approach turned out to be
quite useful. Together, Igor Cadez, who joined my team as summer
intern, Padhraic Smyth, Igor’s PhD advisor, and Chris Meek built
WebCanvas \cite{CHMSW03wc}
\href{https://doi.org/10.1145/347090.347151}{[10.1145/347090.347151]}.
We took page visits from
about one million user sessions on msnbc.com, learned the
model, used the model to assign each user session to a cluster ({\em i.e.}, a
mixture component), and then displayed representative samples from
each mixture component.  The result is shown in
Figure~\ref{fig:webcanvas}.

Steven used the model on the msnbc.com data and derived a number of
unexpected, actionable insights.  For example, there were a large
number of hits to the weather pages (cluster 1), there were large
groups of people entering msnbc.com on tech (clusters 11 and 13) and
local (cluster 22) pages, there was a large group of people navigating
from on-air to local (cluster 12), and there was little navigation
between tech and business sections. I mention this project because, to
this day, I regularly get requests for the code. It seems to remain a
useful tool.

\begin{figure}
\begin{center}
\leavevmode
\includegraphics[width=5.0in]{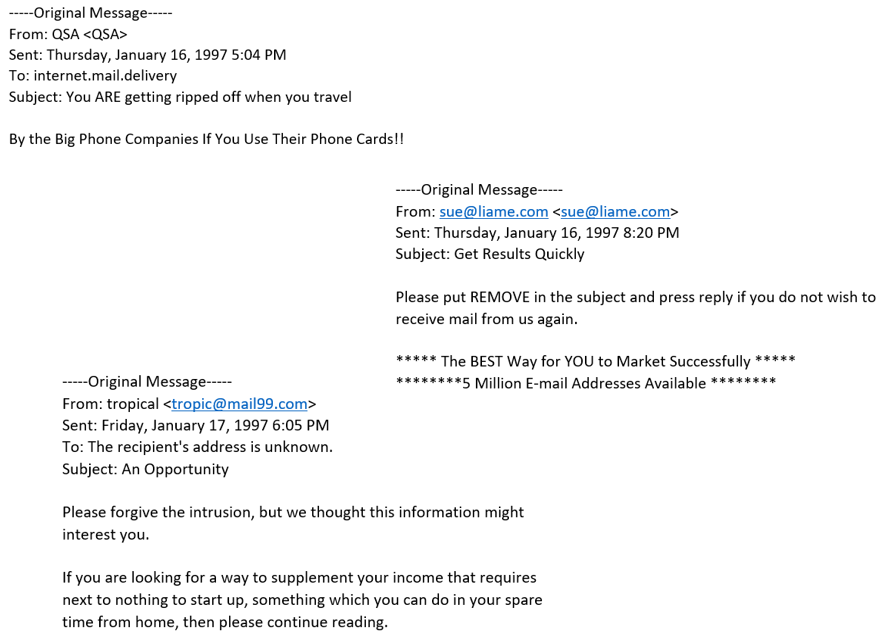}
\end{center}
\caption{The ``burst'' of emails that prompted me to invent the spam
  filter.}
\label{fig:junk-mail}
\end{figure}

\begin{figure}
\begin{center}
\leavevmode
\includegraphics[width=3.0in]{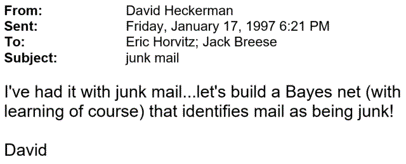}
\end{center}
\caption{The email I sent initiating the creation of a spam filter.}
\label{fig:filter-invent}
\end{figure}

Perhaps the most impactful non-medical project that I initiated at
Microsoft was, to my knowledge, the world's first ML-based spam
filter. It was late 1996, and I started to get a new type of email.
This email was coming from people I didn’t know and was usually
asking me to buy something.  In early 1997, as shown in
Figure~\ref{fig:junk-mail}, I got a ``burst'' of emails---three in
less than 24 hours!  This was the straw that broke the camel’s
back. Even though I had only received about a dozen of this new type
of message, I knew it was going to become a big problem. Sixteen
minutes after seeing the third message, I sent an email to Eric and
Jack: Figure~\ref{fig:filter-invent}.

I followed up with an email asking Eric and Jack to start saving their
email, so we could use it for training. Within a few months, Mehran
Sahami spent an internship at MSR and built our
first machine-learning spam filter. We tried several classifier
methods, including Bayes nets and (ML) decision trees, but the
approach that worked best was suggested and implemented by John Platt:
the Support Vector Machine. Because we wanted the classifier to output
probabilities for decision making (do we label the email as spam or
not?), John was motivated to calibrate the classifier output and
invented Platt Scaling.

Our spam filter was a big hit and, in 1998, it almost shipped in
Outlook Express, a free and (ironically) feature-light version of
Outlook. While still in beta release, however, Microsoft got a call
from Blue Mountain, a greeting card company, complaining that the spam
filter was blocking its greeting cards when a user-controlled
threshold knob in the filter was set to a low setting. We immediately
met with them, appreciated the problem, and said we would have it
fixed by the end of the week.  Before the end of the week came,
however, Blue Mountain sued Microsoft for deliberately trying to ruin
their business. Despite our reassurances and despite the fact that
Outlook Express was still in beta, Blue Mountain supported their claim
by pointing to a single email greeting card that Microsoft had made
available to its customers that, at the filter threshold blocking
Blue Mountain, made it through our filter.

The media went wild.  Readers accepted Blue Mountain's claim without
hesitation.  Microsoft was ``obviously guilty.''  Here’s just one blog
from boards.fool.com:
\begin{quotation}
\noindent Why would Microsoft want to prevent electronic greeting
cards from being delivered? It turns out that after an unsuccessful
attempt to purchase Blue Mountain Arts, Microsoft started its own
electronic greeting card service. The ``bug'' in Outlook Express
appeared at about the same time that Microsoft's greeting card service
began.
\end{quotation}
I got my first, brutal taste of ``fake news.’’  The spam filter was
promptly removed from Outlook Express.

\begin{figure}
\begin{center}
\leavevmode
\includegraphics[width=4.0in]{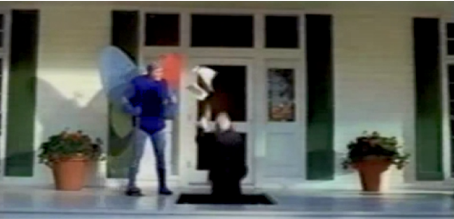}
\end{center}
\caption{A 2003 Super Bowl commercial advertising Microsoft's spam
  filter. The MSN butterfly, personifying the spam filter, pulls the
  trap door on a would-be spammer trying to get into the house,
  representing the users email inbox. As the spammer descends through
  the trap door, his ``message'' flies out of his hands.}
\label{fig:super-bowl}
\end{figure}

Years went by. Spam got more voluminous and
annoying, but Microsoft refused to ship a spam filter. Finally, Bryan
Starbuck, a programmer in MSN 8 did something about it. He stumbled on
our code and knew that shipping it was the right thing to do. His plan
was to sneak it into the release and wait until just before release
to tell others about it, when it would be at high risk to remove. His
plan worked. The spam filter shipped in MSN 8 in 2002.  This time, it
was well received. Just a few months after release, the spam filter
was the focus of Microsoft’s Super Bowl commercial. A snapshot is
shown in Figure~\ref{fig:super-bowl}.  Once the ice was broken, it
became a no-brainer to ship it in other Microsoft products including
Outlook and Exchange Server.

Besides the Blue Mountain suit, there is one other regret I have about
the path that our spam filter took. In our original versions, the
filter was personalized to the user’s email. The user would
(optionally) populate a junk-mail folder, and an ML classifier would
run in the background, constantly updating the filter. It worked
really well. For hundreds of correct calls, there was typically only
one false positive. For some reason, however, senior leadership
decided that asking the user to maintain a junk-mail folder was too
onerous and instead asked users to identify good senders versus
junk-mail senders. To this day, I spend at least an hour a month going
through my junk mail, because my one-size-fits-all filter has so many
false positives.

An immediate spinoff of the spam-filter work was text categorization,
which first shipped in Microsoft Sharepoint Portal Server 2001.
Here, personalization was central. A user would create a
category folder, drop documents into it, and the system would suggest
likely categories for new documents.  For technical details
on spam filtering and text categorization, see
\cite{SDHH98spam}
\href{https://www.aaai.org/Papers/Workshops/1998/WS-98-05/WS98-05-009.pdf}{[AAAI:WS98-05-009]}
and \cite{DPHS98text}
\href{https://doi.org/10.1145/288627.288651}{[10.1145/288627.288651]},
respectively.

\section{Applications in healthcare}
\label{sec:app-health}

I’ve already highlighted some of the healthcare applications I worked
on while at Stanford.  When I first got to MSR, despite my interest in
continuing this work, Nathan asked me to focus on non-healthcare
applications.  Nonetheless, in my first few years there, MSR began to
grow exponentially. Soon, Nathan and Rick opened things up, saying
that we could work on just about anything we wanted to. So, I began
re-exploring healthcare applications.

The first application was a no-brainer. In 1994, Steven Freedman, who
went to Medical School with Eric and me at Stanford, joined Microsoft
to build one of Microsoft’s first internet experiences, a system to
help young families manage pregnancies and the health issues of their
children. Steven, knowing about Pathfinder, Intellipath, and Knowledge
Industries, asked Eric and me to contribute to this system by
constructing an expert system to address diagnostic challenges in this
space. The work was straightforward, but rewarding. Microsoft
Pregnancy and Child Care was born. \comment{An on-computer version of the
application still runs on Windows 11.}

\begin{figure}
\begin{center}
\leavevmode
\includegraphics[width=5.0in]{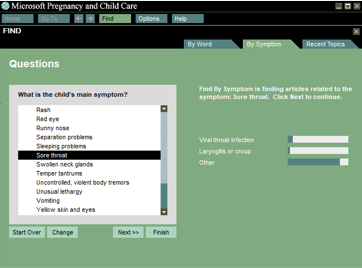}
\end{center}
\caption{A screenshot from Microsoft Pregnancy and Child Care.}
\label{fig:msn-med}
\end{figure}

Figure~\ref{fig:msn-med} shows a screenshot at the beginning of a
diagnostic case.  A parent has just told the system that her young
toddler has a sore throat. The system infers a probability
distribution, shown on the right, and is about to ask questions of the parent
to narrow the list of possible diagnoses. The project was a technical
success, but never saw release. An important customer of Microsoft
asked Steve Ballmer to kill the project, because they were working in
the same space.  Steve complied and, sadly to this day, no similar
project has been released by that customer. Another somber note: In
those days, Microsoft had lavish parties to celebrate the new release
of a product. The cancellation of the project came just days before
the release party, so the party went on as scheduled with the more
than the 100 people who worked on the project. It was a very
depressing party.

Another large project was my work to create a vaccine for HIV.  It was
2004, and I had just formed the Bioinformatics group at MSR.  Nebojsa
Jojic and I visited Jim Mullins, a prominent HIV scientist at the
University of Washington.  Both Nebojsa and I immediately got excited
about the prospect of using machine learning to advance what was
already a massive effort to construct a vaccine for HIV. The
excitement was mutual. Jim introduced us to Rick Klausner, who
introduced us to Simon Mallal, who introduced us to Bruce Walker, who
introduced us to Nicole Frahm, Zabrina Brumme, Florencia Pereyra, and
Marry Carrington, among others. At Microsoft, several folks including
Carl Kadie and Jonathan Carlson joined the effort.  We made good
progress on the goal. Here, I won’t go through the dozens of
manuscripts that resulted, but I will summarize one line of work,
which remains promising.

There is an interesting analogy between spam filtering and HIV vaccine
design that helps to explain this work. When our first spam filter
came out, spammers figured out clever ways to work around it. For
example, spammers would encode sensitive words like “Viagra” in bit
maps, so our filters wouldn’t recognize the occurrence of the word. In
response, we would change our filters to catch this case, but the
spammers would just adapt again. The battle was on.  Eventually, we
realized that we had to try to find an Achilles heel of the
spammers---something they couldn’t adapt away from. We found one:
spammers needed to make money. At the time I worked on this problem,
there were only a relatively small number of sites that could process
payments. We catalogued those sites and used any reference to such a
site as a key feature for predicting whether a message was spam. The
approach worked well.

Remarkably, we experienced something similar when working on HIV.  Our
immune system attacks HIV by killing cells that harbor the virus. In
response, the virus mutates to avoid the attack. More specifically,
just about every cell displays peptides, short fragments of proteins
manufactured inside the cell, on its surface.  The peptides are
displayed on HLA molecules, proteins coded for by three HLA genes on
chromosome six. Each of these three genes have hundreds of varieties
and, consequently, individuals rarely attack all the same HIV
peptides.  This diversity is a good thing. If there wasn’t such
diversity, then a virus (HIV or otherwise) that killed one of us would
kill all of us. (Each of us has one or two varieties of each of the
three genes---one from mom and one from dad.  It’s better when a
person has two different versions of each gene to maximize the variety
of viruses that can be attached. Interestingly, there is scientific
evidence that a women can literally smell the HLA molecules of a man
and is attracted to men that have different varieties then she
does. Perhaps the chemistry of love is indeed chemistry.)  Constantly,
T-cells come into contact with the cell to check that these peptides
are not foreign to the body. If they are, the T-cell gets excited,
which leads to the creation of T-cells that can kill any cell
presenting one or more foreign peptides. Unfortunately, when this
killing happens, HIV mutates to avoid attack. It turns out that HIV is
a very robust virus that can thrive with many mutations.  A single
individual will typically harbor thousands of different HIV varieties.
Herein lies the analogy.  The immune system is like the spam filter,
and HIV’s ability to mutate and thrive is like a spam email's ability
to exist in many forms and still get the message across (such as
including ``Viagra'' in a bit map).

\begin{figure}
\begin{center}
\leavevmode
\includegraphics[width=6.0in]{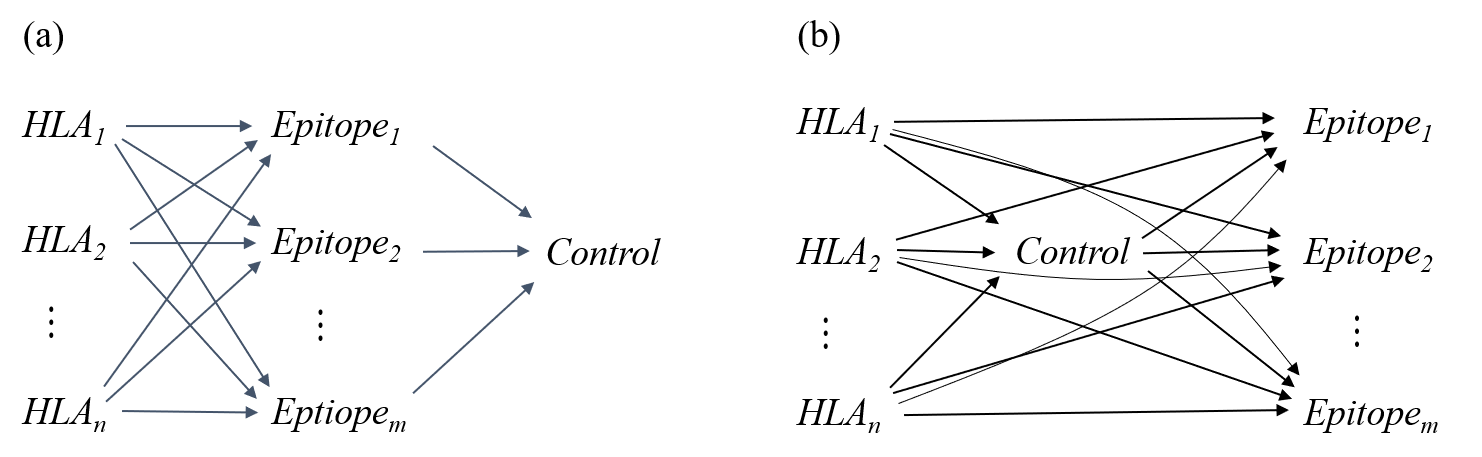}
\end{center}
\caption{Causal models depicting two hypotheses for the relationships
  among HLA molecules in a subject, which peptides are targeted by
  their immune system, and whether their immune system controls the
  virus.}
\label{fig:hiv-control}
\end{figure}

To continue the analogy, a key question becomes: does HIV have an
Achilles heel?  Evidence that it does comes from the observation that
there is a small fraction of humans, called ``controllers,'', who
naturally control HIV---that is, they get infected with HIV, but don't
go on to develop serious disease.  One causal hypothesis is that these
controllers have HLA molecules that attack the virus in vulnerable
positions along HIV proteins. To test this hypothesis Bruce Walker and
team identified the HLA molecules, targeted peptides, and the presence
or absence of control in 341 individuals infected with HIV. My team
determined the likelihood of data from the causal model in
Figure~\ref{fig:hiv-control}a and compared it to the likelihood of a
causal model for an alternative hypothesis shown in
Figure~\ref{fig:hiv-control}b.  The alternative hypothesis is that HLA
directly causes control, and that the lack of control leads to a
proliferation of HIV varieties that determines which peptides are
targeted.  The first hypothesis had a significantly higher likelihood
\cite{HIV14jv}
\href{https://doi.org/10.1128/JVI.01004-14}{[10.1128/JVI.01004-14]}.

\comment{Omitted here is the entropy method for identifying good epitopes.}

\begin{figure}
\begin{center}
\leavevmode
\includegraphics[width=5.0in]{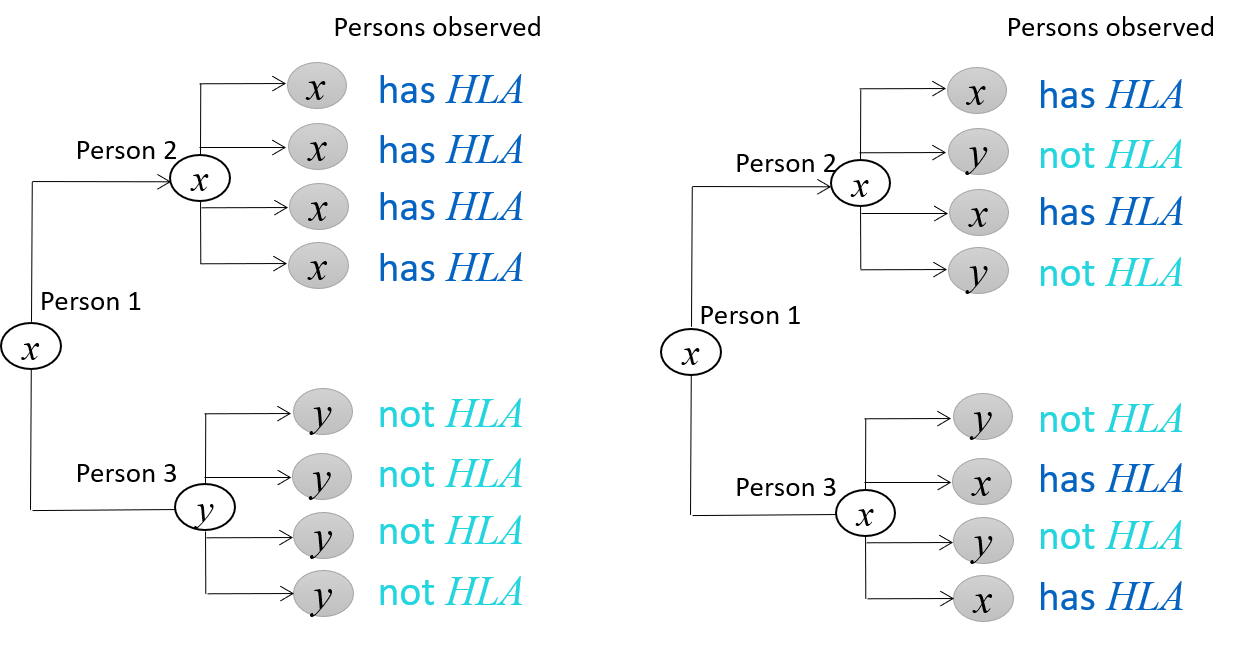}
\end{center}
\caption{Two evolution histories of HIV and their impact on mutations.}
\label{fig:phylod-why}
\end{figure}

Learning graphical models also help identify which parts of the virus
are attacked by our immune system, a key task in HIV vaccine design.
There are many locations or positions to attack. The HIV genome is
over 9,000 nucleotides long and codes for the production of nine
primary proteins with over 3,000 amino acids. Each amino acid
position is a possible target for attack. As just discussed, the task
of identifying targets is complicated because there is a lot of
variation in the human immune system from person to person, and each
variation can attack different parts of HIV.  So, the task of
identifying attack targets amounts to identifying correlations between
a person's particular HLA molecules and the particular amino acids at
each position along HIV's sequence.  A difficulty with this approach
is that HIV sequences are constantly evolving as they pass from one
individual to the next, and so history is a confounder of the
correlations. An example is shown in Figure~\ref{fig:phylod-why}.

In the figure, there are two different evolution histories---sometimes
called phylogenies---that result in observations of amino-acid $x$ or
$y$ at a given HIV sequence position. The shaded nodes are observed;
the unshaded nodes are not. Also shown are whether the individuals
with observed HIV sequences have a particular HLA variety or not. In
both possible histories, there are four individuals with a particular
HLA and an observed amino-acid $x$, and four individuals with the lack
of that HLA and amino-acid $y$ at the same position. When analyzed
naively, the correlation between the presence or absence of the HLA
and the presence of $x$ versus $y$ at this position is substantial. Now,
however, consider the two histories. In the first history, the
amino-acid $x$ is transmitted from person 1 to person 2, but person 3
gets a mutated $y$. In the second history, all three persons have
amino-acid $x$---all mutations to $y$ happen in the persons
observed. In the first history, the evolution from person 1 to persons
2 and 3 explain the correlation. In the second history, the presence
or absence of the particular HLA explains the evolution. That is, the
second history is much stronger evidence that the particular HLA
molecule is causing the mutation.

\begin{figure}
\begin{center}
\leavevmode
\includegraphics[width=3.0in]{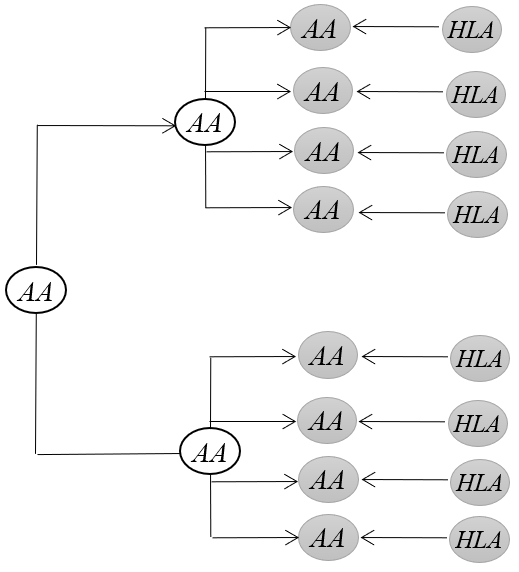}
\end{center}
\caption{The PhyloD model for eight subjects.}
\label{fig:phylod}
\end{figure}

To take phylogeny into account when identifying HLA--amino-acid
associations, Jonathan Carlson, Carl Kadie, and I developed a system
called PhyloD \cite{CH08phylod}
\href{https://doi.org/10.1371/journal.pcbi.1000225}{[10.1371/journal.pcbi.1000225]}.
At the core of the system is a DAG model, a simple example of
which is shown in Figure~\ref{fig:phylod}. Again, shaded nodes are
observed; unshaded nodes are not. The structure of the phylogeny is
learned from the amino acids observed at all sequences. Given
observations of the shaded nodes, the system uses the
Expectation--Maximization algorithm to infer distributions on the
unobserved variables. To determine the strength of the association
between $HLA$ and $AA$, the likelihood of the data for this model is
compared to a null model with only the $HLA$ observations. Of course, a
Bayesian approach could be used, but this frequentist approach was
computationally efficient and effective.

Before 2012, my work on HIV was getting modest attention by senior
leadership at Microsoft.  Then, something random happened that changed
that.  My son's elementary school asked me to speak to their
graduating class. I was happy to do it, but it took me a week or so to
create a talk for the class that was simple enough for them to
understand. The kids loved it, so I decided to give it a try at
Microsoft.  Leadership loved it and asked me to speak at the
2013 Microsoft Worldwide Partner Conference with many thousands
of people in attendance.
You can see the talk on \href{https://www.youtube.com/watch?v=Jgl6EZoJFEc}{[youtube]}.
There's lesson in here somewhere.

\comment{Include delivery mechanism story with Reid?}

The work on PhyloD led to a useful genomics tool.  In 2001, the first
human genomes were (almost completely) sequenced by teams led by Craig
Venter and Francis Collins for 300 million dollars and 3 billion
dollars, respectively. Today, sequencing the human genome costs about
500 dollars.  This is quite impressive, given that the human genome
consists of about three billion nucleotide pairs. As a result of this
accomplishment, there has been a flood of genomics data leading to
important findings and applications.

One important finding is that, remarkably, the differences between two
individuals' DNA are extremely small---about 0.1\%.  That is, if you
look at the three billion As, Cs, Ts, and Gs in the human genome
(times two, because we each have two copies of DNA, one from mom and
one from dad), you’ll only find millions of differences---called
single nucleotide polymorphisms or SNPs.  (There are also more complex
differences such as insertions, deletions, and multiple copies of
small regions of DNA, but I won’t get into that here.)  These
``small'' differences are a big plus for computer scientists and
computational biologists who are working with and analyzing genomic
data, as it drastically reduces the scale of their computational
problems.  Another interesting finding is that almost all SNPs take on
only two of the four possible nucleotide values (A, T, C or G), which
makes it possible to talk about each SNP as a binary variable with a
major and minor nucleotide or {\em allele}.

Perhaps the most compelling application of the genomics revolution is
personalized or precision medicine, where medical care is customized
to the patient based on their genome.  As an example, it turns out
that if you happen to have a particular set of SNPs in the region of
your DNA that codes for the HLA molecules, then if you take
carbamazepine (a drug that suppresses seizures), you will very likely
develop Stevens-Johnson syndrome, a horrible disease where layers of
your skin separate from one another \cite{Ferrell09}.  One idea behind
personalized medicine is to avoid such fates by checking your genome
before administering drugs.  More generally, with personalized
medicine, your genome could be used to identify drugs that would work
well for you or to warn you of diseases for which you are at high
risk, such as diabetes and heart disease, so you can take measures to
avoid them.

This application brings us to the question: How do we identify
associations between your genome and various traits such as whether
you will get a disease or whether a particular drug will harm you?
Although older techniques have been used for decades, the genomics
revolution has led to the use of a new technique known as Genome-Wide
Association Studies or GWAS.  With GWAS, you identify a set of
individuals, measure many or all of their SNPs, and then identify
correlations between those SNPs and one or more traits.

One surprising realization from these studies is that the
Stevens-Johnson-syndrome example---where only a few SNPs are
associated with the trait---is the exception rather than the rule.
Namely, it is now recognized that, typically, many SNPs (hundreds and
more) are causally related to a single trait \cite{LSFH10}.
Furthermore, it turns out that whether an association can be detected
depends on the frequency of the minor allele and the strength of the
SNP’s effect on the trait. SNPs with larger effect tend to have very
rare minor alleles due to natural selection.  In contrast, SNPs with
more common minor alleles tend to have little effect on the trait.
Thus, extremely large sample sizes---hundreds of thousands or
more---are required to create a comprehensive picture of the
relationships among SNPs and traits.  Many groups including the
governments of the US and England are working to generate such data.
Unfortunately, large data sets often lead to the introduction of
confounders.  As an example, a data set might include individuals with
different ethnicities.  In this situation, if one ethnicity is
correlated with both the trait and a SNP, then the trait and a
non-causal SNP will be correlated.  Confounders of this sort are known
as {\em population structure}.  In large data sets, we also often see
individuals who are related to each other, either closely or
distantly.  These relationships, known as {\em family relatedness},
also introduce correlations among SNPs and traits that are not
causally related.

\begin{figure}
\begin{center}
\leavevmode
\includegraphics[width=3.0in]{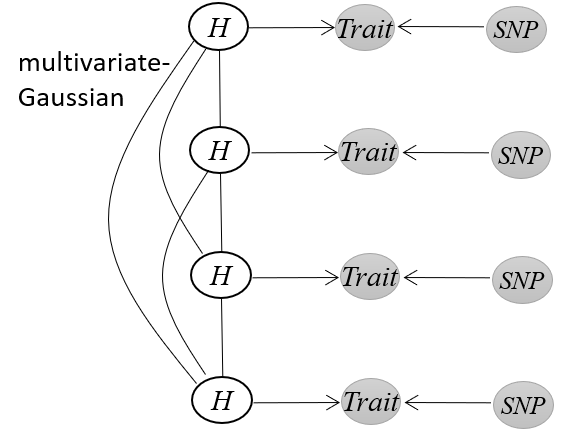}
\end{center}
\caption{A modification to PhyloD for GWAS. The model is equivalent to a linear-mixed model.}
\label{fig:lmm}
\end{figure}

This issue raises the question: How can we deal with confounders and
identify only the truly causal associations?  When I first heard about
this problem from an intern, Noah Zaitlen, I immediately thought of
adapting PhyloD to solve the problem.  I replaced amino-acid identity
with trait and HLA with SNP. It was also clear that a phylogeny would
not adequately reflect the evolutionary history of humans, as we have
two copies of each piece of DNA, whereas HIV has only one copy. So, I
replaced the phylogeny with a continuous hidden variable for each
individual and modeled these variables jointly with a
multivariate-Gaussian distribution whose covariance matrix represents
the similarity among individuals and is learned from the available
SNPs for each individual.  An example graphical model for four
individuals is shown in Figure~\ref{fig:lmm}.

I discussed this model with Elezar Eskin, Noah’s advisor. In turn,
Elezar discussed the model with Nicholas Schork, who recognized that it was
equivalent to a linear mixed model (LMM), which had been used by
animal breeders for decades. Elezar and team and I went on to
publish the model, speeding up previous methods for estimating its
parameters and demonstrating its use on several genomics datasets
\cite{EMMA08}
\href{https://doi.org/10.1534/genetics.107.080101}{[10.1534/genetics.107.080101]}.
Given a dataset with $N$ individuals and
$M$ SNPs to be tested for association, the algorithm ran with
computational complexity $O(MN^3)$, which unfortunately was still too
slow for most modern datasets.  (Animal breeders typically analyzed
datasets with dozens or hundreds of samples.)

A real advance happened when Christoph Lippert joined our 
Bioinformatics group at MSR.  Christoph identified algebraic
manipulations that yielded exact results, but with computational
complexity $O(MN\min(N,K))$, where $K$ is the number SNPs used to infer the
covariance matrix of the multivariate Gaussian \cite{FaST-LMM11}
\href{https://doi.org/doi:10.1038/nmeth.1681}{[doi:10.1038/nmeth.1681]}.
The algorithm, called FaST-LMM, is still in frequent use today.

Our group at MSR went on to develop many extensions to this work. For
example, recognizing that we could further speed up the algorithm by
reducing $K$, we developed two ideas for doing so. One worked well and
one worked horribly.  The one that worked horribly was to estimate the
covariance matrix with SNPs that were detectably correlated with the
trait. We called this approach FaST-LMM-Select. The problem with the
approach was that there was so much noise in the correlations between
SNPs and trait, that we ended up throwing out the baby with the
bathwater. Unfortunately, there was an error in one of the experiments
that suggested the approach was working well (see Figure S1 from the
original paper \cite{FaST-LMM-Select12}), and we didn't catch this
error until several additional papers were published.  The bottom
line: don’t use FaST-LMM-Select.

The approach that did work well was to take advantage of {\em linkage
  disequilibrium}, the observation that nearby SNPs are highly
correlated. Given this correlation, sampling SNPs ({\em e.g.}, using every
fifth SNP in order along the chromosome) to estimate the covariance
matrix yields results that are quite close to exact
\cite{FaST-LMM14}
\href{https://doi.org/doi: 10.1038/srep06874}{[doi: 10.1038/srep06874]}.

Regarding linkage disequilibrium, when applying LMMs to real data, I
noticed that statistical power was lost when SNPs near the SNP being
tested (and the test SNP itself) are included in the estimation of the
covariance matrix. This is not surprising: when including such SNPs in
the estimation, the correlation matrix ``steals’’ part of the
impact of the test SNP on the trait, thus diminishing the
estimate of that impact
\cite{FaST-LMM11}
\href{https://doi.org/doi:10.1038/nmeth.1681}{[doi:10.1038/nmeth.1681]}.
(The same principle applies to
PhyloD.) This observation led to what is now a standard of practice in
GWAS known as ``leave out one chromosome.’’  That is, when testing a
SNP on a particular chromosome, use a covariance matrix estimated with
SNPs from all but that chromosome.

Another extension of FaST-LMM involved testing for associations
between a set of SNPs ({\em e.g.}, those associated with a particular gene)
and a trait. The approach essentially combines very weak signals,
yielding increased statistical power \cite{FaST-LMM14set}
\href{https://doi.org/10.1093/bioinformatics/btu504}{[10.1093/bioinformatics/btu504]}.
Other extensions sped up the identification of SNP--SNP interactions that
are associated with a trait \cite{Fast-LMM13epistasis}
\href{https://doi.org/10.1038/srep01099}{[10.1038/srep01099]},
and addressed
situations where the presence of rare traits are over-represented in
the data \cite{LEAP15}
\href{https://doi.org/10.1038/nmeth.3285}{[10.1038/nmeth.3285]}.
We also created a
version of FaST-LMM that incorporated more than one covariance
matrix---for example, when there is confounding both from SNP
similarities and from environmental effects due to differences in
spatial location \cite{FaST-LMM16pnas}
\href{https://doi.org/10.1073/pnas.1510497113}{[10.1073/pnas.1510497113]}.
In addition, we optimized FaST-LMM for the cloud
\cite{KH19ludicrous}
\href{https://doi.org/10.1101/154682}{[10.1101/154682]}.

I close with two stories about my genomics work.  In 2006, after a
meeting between Microsoft and MIT in Boston, I had the fortune to fly
back to Seattle with Craig Mundie, who led Microsoft’s advanced
technology efforts. We got to talking about the huge potential for
genomics in general, and GWAS in particular, to health and
wellness. Out of that discussion came the idea to build a service that
would sequence customers’ SNPs, ask them about important traits,
identify genome-wide associations, and inform them of traits they
should look out for.  We anticipated 23andMe and Ancestry.
Unfortunately, although Microsoft filed a
\href{https://image-ppubs.uspto.gov/dirsearch-public/print/downloadPdf/20070112598}{patent}
on the idea, and I tried repeatedly to generate interest within
Microsoft to build the service, we were beaten to the punch.

Perhaps the most impactful outcome of my work in genomics came in
2010, when Bryan Traynor and team and I performed a GWAS on a cohort of Finnish
individuals, some of whom had Amyotrophic Lateral Sclerosis (ALS).  We
found a signal on chromosome 9. Bryan then did the difficult work to
find the precise mutation on the genome that was producing the signal:
an expansion of the C9ORF72 gene consisting of a repeated
six-nucleotide-long sequence \cite{ALS11}
\href{https://doi.org/10.1016/S1474-4422(10)70184-8}{[10.1016/S1474-4422(10)70184-8]}.
Later, it was found that this mutation accounts for about 30\% of inherited cases of ALS, and
treatments based on this finding are now in clinical trials
\cite{ALS22trials}.

\section*{Acknowledgments}

I sincerely thank
Carl Kadie,
Chris Meek,
Dan Geiger,
Eric Horvitz,
Jack Breese,
Jonathan Heckerman,
Max Chickering,
Ross Shachter,
and Pedro Domingos
for their comments on earlier drafts.

\bibliography{heckerthoughts}
\bibliographystyle{plain}

\section*{About the author}

David Heckerman became an ACM Fellow in 2011 for ``contributions to
reasoning and decision-making under uncertainty.'' He is known for
developing the first practical platform for constructing probabilistic
expert systems, the topic of his PhD dissertation, which won the ACM
best dissertation award in 1990.  He is also known for developing an
approach for learning Bayesian networks from a combination of expert
knowledge and data that has proven useful in causal discovery,
developing an HIV vaccine design through machine learning, and
developing state-of-the-art methods for genome associations
studies. David is currently a VP and Distinguished Scientist at
Amazon.  Before that, he worked at Microsoft Research from 1992 to
2017.  At MSR, he founded the first AI group in 1992, the first
machine-learning group in 1994, the first bioinformatics group in
2004, and the first genomics group in 2015.  Notable products he
invented include the world’s first machine-learning spam filter, the
Answer Wizard (which became the backend for Clippy), and the Windows
Troubleshooters. He received his Ph.D. in Medical Informatics in 1990
and an M.D. in 1992 from Stanford University, and is also a Fellow of
AAAI and ACMI.

\end{document}